\documentclass{article}

\PassOptionsToPackage{numbers,sort,compress}{natbib}
\usepackage[eandd,final]{neurips_2026}
\usepackage[utf8]{inputenc}
\usepackage[T1]{fontenc}
\usepackage{hyperref}
\usepackage{url}
\usepackage{booktabs}
\usepackage{amsfonts}
\usepackage{nicefrac}
\usepackage{microtype}
\usepackage{multirow}
\usepackage{graphicx}
\usepackage{amsmath, amssymb}
\usepackage{enumitem}
\usepackage{array}
\usepackage{tabularx}
\usepackage{makecell}            
\usepackage[table]{xcolor}       
\usepackage{colortbl}
\usepackage{float}               
\usepackage{changepage}          
\usepackage[framemethod=TikZ]{mdframed}  

\title{AgentRobustBench: A Sim-to-Real Benchmark and Domain-Randomized RL Recipe for Tool-Use Agents}

\author{%
  \textbf{Xiaolin Zhou$^{1}$ \quad Aojie Yuan$^{2}$ \quad Zheng Luo$^{2}$ \quad
  Zipeng Ling$^{4}$ \quad Xixiao Pan$^{2}$} \\
  \textbf{Yicheng Gao$^{2}$ \quad Haiyue Zhang$^{2}$ \quad Jiate Li$^{2}$ \quad
  Shuli Jiang$^{3}$ \quad Prince Zizhuang Wang$^{3}$} \\
  \textbf{Zixuan Zhu$^{2}$ \quad Jinbo Liu$^{1}$ \quad Ryan A. Rossi$^{5}$ \quad
  Hua Wei$^{1}$ \quad Xiyang Hu$^{1}$} \\
  \\
  $^{1}$Arizona State University \quad
  $^{2}$University of Southern California \\
  $^{3}$Carnegie Mellon University \quad
  $^{4}$University of Alberta \quad
  $^{5}$Adobe Research \\
  \texttt{\{xzhou226,\,xiyanghu\}@asu.edu}
}

\begin{document}

\maketitle

\begin{abstract}
Tool-use language agents are evaluated on benchmarks that assume clean inputs, unambiguous tool registries, and reliable APIs. Real deployments violate all these assumptions: user typos propagate into hallucinated tool names, a misconfigured request timeout can stall an agent indefinitely, and duplicate tool names across servers can freeze an SDK. We study these failures as a sim-to-real gap in the tool-use \emph{partially observable Markov decision process (POMDP)}, where deployment noise enters through the observation, action space, reward-relevant metadata, or transition dynamics. We introduce \textbf{AgentRobustBench}, a benchmark with 22 perturbation types organized by these four POMDP components, each grounded in a verified GitHub issue or documented tool-calling failure. Across 21 models from 1.5B to 32B parameters (including the closed-source o4-mini), the robustness profile is sharply uneven: observation perturbations reduce accuracy by less than 5\%, while reward-relevant and transition perturbations reduce accuracy by roughly 40\% and 30\%, respectively; scale alone does not close these gaps. We then propose \textbf{ToolRL-DR}, a domain-randomization reinforcement learning (RL) recipe that trains a tool-use agent on perturbation-augmented trajectories spanning the three statically encodable POMDP components. On a 3B backbone, ToolRL-DR-Full retains roughly three-quarters of clean accuracy and reaches an aggregate perturbed accuracy comparable to open-source 14B function-calling baselines while substantially narrowing the gap to o4-mini. It closes approximately 27\% of the Transition gap despite never seeing transition perturbations in training, suggesting that RL on adversarial static tool-use inputs induces a more persistent retry policy that transfers to unseen runtime failures. The code, dataset, checkpoints and leaderboard are available through the project repository at \url{https://github.com/WillChow66/AgentRobustBench}.


\end{abstract}

\section{Introduction}
\label{sec:intro}

Tool-using language agents are increasingly deployed in production systems: agent frameworks such as LangChain and AutoGen, MCP-based assistants, and retrieval or code-execution back ends for chat products. Yet the benchmarks used to evaluate these agents (BFCL V3~\citep{patil2024gorilla}, API-Bank~\citep{li2023api}, ToolAlpaca~\citep{tang2023toolalpaca}, RoTBench~\citep{ye2024rotbench}, and others) still score behavior under clean conditions: user queries are well formed, tool registries are unambiguous, tool descriptions are stable, and tool execution is deterministic. Production traffic violates all of these assumptions. A user typo such as ``occcra\_information'' $\rightarrow$ ``occrra\_information'' can be copied into a hallucinated tool name and crash a LlamaIndex dispatcher~\citep{llamaindex7170}. A LangChain client with default \emph{request\_timeout=None} can hang indefinitely on slow tool responses~\citep{langchain35597}. Two MCP servers can register tools with the same name and freeze the OpenAI Agents SDK~\citep{openaiagents1167}. Clean benchmarks do not expose these failures, so models tuned against them receive little signal about how to handle realistic tool-calling errors.

We study this mismatch as a sim-to-real transfer problem. In robotics, a standard response to a training--deployment distribution gap is domain randomization~\citep{tobin2017domain, peng2018sim, sadeghi2017cad2rl, chebotar2019closing,rafi2026sim2signal,weitoward}: train on a broader distribution of environment variations so that deployment-time conditions are less likely to fall outside the training support. Tool use has a similar structure, but the sources of variation differ. Instead of sensor noise, visual textures, or actuator dynamics, tool-use agents face noisy text, ambiguous action spaces, misleading tool metadata, and unreliable tool execution. We formalize these sources through the tool-use partially observable Markov decision process (POMDP): deployment noise can enter the observation, the action space, the reward-relevant metadata used to choose among tools, or the transition dynamics induced by tool execution.

Existing tool-use robustness work studies important but mostly isolated failure modes, such as function-name perturbations~\citep{ye2024rotbench}, distractor tools~\citep{lu2025toolsandbox,liu2026androidreality}, and query or toolkit modifications~\citep{rabinovich2025robustness}. These studies show that tool-use agents are brittle, but they do not provide a unified POMDP-based taxonomy, do not systematically tie perturbation types to production failures, and do not test whether training on perturbed tool-use trajectories can reduce the gap. As a result, it remains unclear which part of the tool-use loop is most fragile, whether scaling alone improves robustness, and whether a data-side training fix can transfer beyond the perturbations seen during training.

In this work, we make three contributions:
\begin{itemize}[leftmargin=*, noitemsep]
    \item \textbf{AgentRobustBench} (\S\ref{sec:benchmark}): a production-grounded benchmark of 199 single-turn samples drawn from five existing tool-use benchmarks, with 22 perturbation types organized by POMDP component (observation, action, reward, transition). Every type is linked to a verified GitHub issue from a major agent framework (LangChain, AutoGen, OpenAI Agents, LlamaIndex, MCP servers, etc.) or to a peer-reviewed study, with an evidence audit released with the data.
    \item \textbf{ToolRL-DR} (\S\ref{sec:method}): a domain-randomization RL training recipe for tool use. ToolRL-DR keeps the public ToolRL training code fixed, but replaces clean training trajectories with perturbation-augmented trajectories sampled from the three statically encodable POMDP components. On a 3B backbone, ToolRL-DR-Full closes approximately 27\% of both the Reward and the Transition gap relative to the public ToolRL checkpoint, despite never training on transition perturbations.
    \item \textbf{A live leaderboard} (\S\ref{sec:platform}): a public submission portal that scores user-supplied predictions against AgentRobustBench, reports per-component robustness, and tracks tool-use robustness over time, following the submission pattern of community leaderboards such as BFCL.
\end{itemize}

We find that the sim-to-real gap is large and uneven, scaling does not reliably close it, and ToolRL-DR partially transfers to unseen Transition perturbations on a 3B backbone. We release the code, datasets, trained checkpoints, and leaderboard.

\begin{figure}[t]
    \centering
    \includegraphics[width=\linewidth]{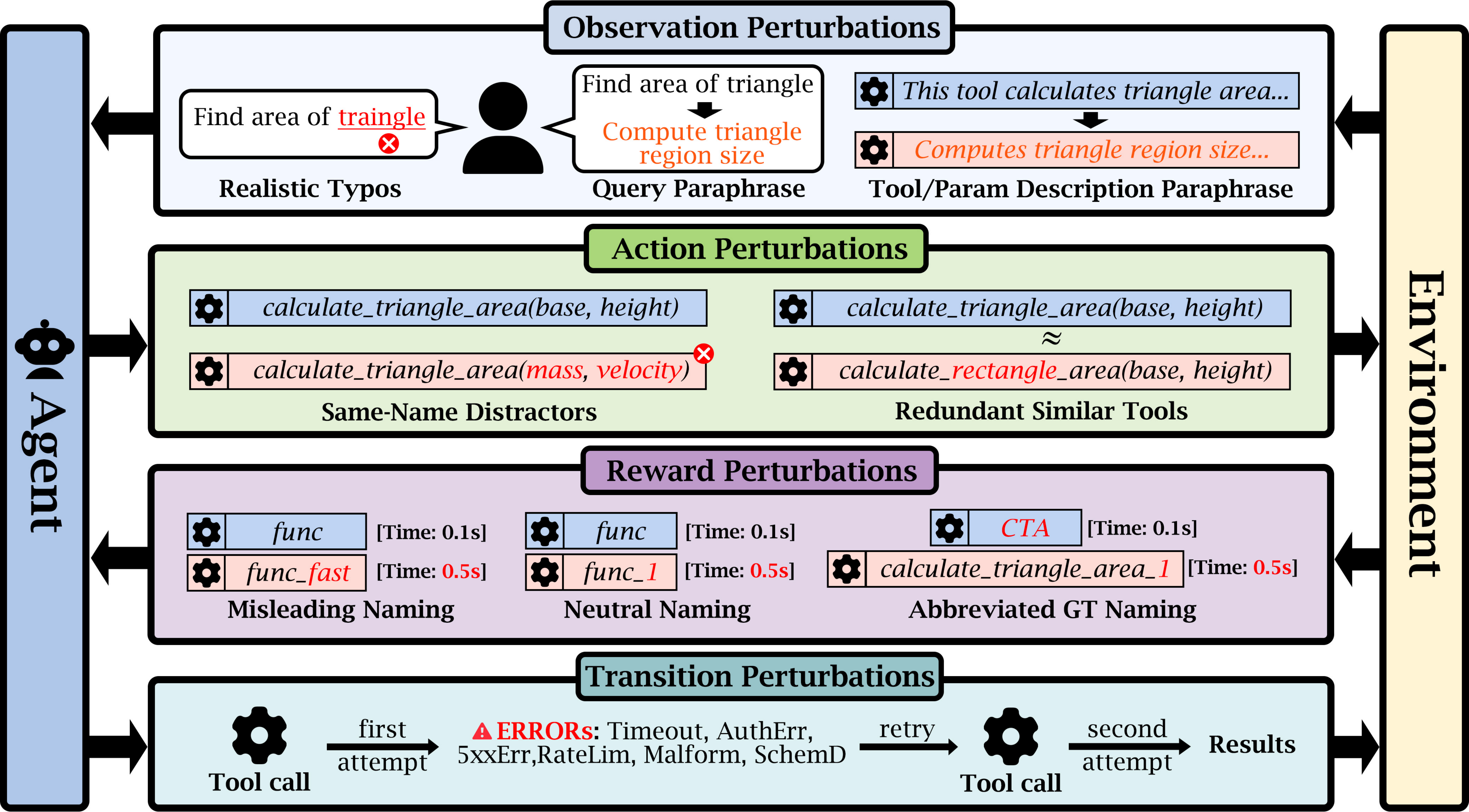}
    \vspace{-15pt}
    \caption{The tool-use POMDP and the four perturbation categories that define AgentRobustBench. Information flows between the \textbf{Agent} (left) and the \textbf{Environment} (right) through four perturbation channels arranged top-to-bottom. \textbf{Observation} perturbations corrupt what the agent reads from the environment (typos in the user query, paraphrases of the query, paraphrases of tool or parameter descriptions). \textbf{Action} perturbations contaminate the available action space (a same-name distractor that swaps argument names, or a redundant similar tool). \textbf{Reward} perturbations corrupt the metadata that disambiguates among similarly-named tools (misleading naming, neutral naming, or an abbreviated ground-truth name). \textbf{Transition} perturbations inject a transient runtime error (Timeout, AuthErr, 5xxErr, RateLim, Malform, SchemD) after the model's first tool call, forcing a retry. Type-level details are in Table~\ref{tab:taxonomy}; production grounding for each type is in \S\ref{sec:bench-grounding}.}
    \label{fig:pipeline}
\end{figure}

\section{Background}
\label{sec:background}

\subsection{Tool use as a partially observable Markov decision process}
\label{sec:bg-mdp}

We treat a tool-using agent as a partially observable Markov decision process (POMDP) $\langle \mathcal{S}, \mathcal{A}, \mathcal{P}, \mathcal{R}, \Omega, \mathcal{O} \rangle$~\cite{da2025survey,liu2026sim}. The hidden state $s_t \in \mathcal{S}$ encapsulates the user's underlying intent, the actual implementation of every tool, and the runtime health of every service. The agent does not access $s_t$ directly; instead it receives an observation $o_t \in \Omega$ produced by the observation function $\mathcal{O}(o_t \mid s_t)$, comprising the dialog prompt, the tool registry it can see (names, descriptions, parameter schemas), and the outputs of any tools already executed. An action $a_t \in \mathcal{A}$ is a structured tool invocation, i.e.\ a pair (tool name, argument map). The transition $\mathcal{P}(s_{t+1} \mid s_t, a_t)$ executes the chosen tool and updates the hidden state; the agent then receives a new observation. The reward $\mathcal{R}(s_t, a_t)$ scores whether the invocation is correct and, in deployed systems, may also depend on cost, latency, or other metadata relevant to tool choice.

The agent samples actions from a history-dependent policy $\pi_\theta(a_t \mid h_t)$ where $h_t = (o_1, a_1, \ldots, o_t)$ is the observation--action history; in our setting the policy is a function-calling LLM that consumes $h_t$ as a prompt. Each POMDP component is a place where simulation and reality can differ: the observation function $\mathcal{O}$ may produce noisy text the agent did not see in training; the action space $\mathcal{A}$ may include distractor tools; the transition $\mathcal{P}$ may produce transient errors not present in the curated benchmark; and the reward-relevant metadata visible inside $o_t$ may be misleading. We use this decomposition to organize our perturbations (\S\ref{sec:benchmark}).

\subsection{Sim-to-real and domain randomization}
\label{sec:bg-dr}


Sim-to-real transfer studies policies trained in a source environment, often a simulator, and deployed in a target environment whose conditions differ~\cite{}. In robotics, a standard response to this training--deployment gap is \emph{domain randomization}~\citep{tobin2017domain, sadeghi2017cad2rl, peng2018sim, tan2018sim}: during training, sample environment parameters such as textures, dynamics, masses, friction, or sensor noise from a broad distribution so that the target environment is less likely to fall outside the training support. Adaptive variants go further by updating the randomization distribution \emph{online} during training, either by progressively expanding the range as the agent's competence grows~\citep{akkaya2019solving} or by leveraging real-world rollouts to close the simulation-to-reality loop~\citep{chebotar2019closing}. Further details on randomization axes and transfer settings are provided in~\citep{zhao2020sim, hofer2021sim2real}.

We apply the same lens to tool-use agents. The perturbation axes differ from robotics: noisy text replaces sensor noise, transient API errors replace actuator or dynamics mismatch, and distractor tools replace visual or physical clutter. The underlying problem is similar: clean training and benchmark data cover a narrower distribution than the conditions encountered during deployment. Concretely, we organize tool-use perturbations along the four POMDP components: \textbf{Observation} (noisy text in user queries and tool metadata), \textbf{Action} (distractor tools in the action space), \textbf{Reward} (misleading metadata that biases tool selection), and \textbf{Transition} (transient runtime errors during execution). The full type-level taxonomy and grounding evidence are in \S\ref{sec:bench-taxonomy}.
Section~\ref{sec:method} instantiates domain randomization for tool calling by training on perturbation-augmented trajectories rather than only clean tool-use trajectories.

\section{Related Work}
\label{sec:related}

\textbf{Tool-use LLM agents}
Early tool-use agents combined Chain-of-Thought reasoning~\citep{wei2022chain} and ReAct-style observation loops~\citep{yao2022react} with supervised fine-tuning on curated tool-use trajectories~\citep{schick2023toolformer, qin2024toolllm, tang2023toolalpaca, liu2024apigen}. Recent systems increasingly use reinforcement learning to optimize tool-call behavior. ToolRL~\citep{qian2025toolrl} applies GRPO with structured rewards over synthetic tool-call trajectories. TL-Training~\citep{ye2024tl} combines supervised fine-tuning with token-weighted reinforcement learning. LoopTool~\citep{zhang2025looptool} uses judgment-guided label refinement to close a data--training loop. MUA-RL~\citep{zhao2025mua} trains with simulated users in multi-turn rollouts. We evaluate representatives of these RL-trained tool-use families, together with their base or instruction-tuned counterparts where applicable.

\textbf{Tool-use benchmarks}
Tool-use benchmarks have expanded across task format, tool diversity, and interaction length. API-Bank~\citep{li2023api} evaluates planning, retrieval, and API invocation. The Berkeley Function-Calling Leaderboard (BFCL)~\citep{patil2024gorilla} evaluates cross-domain function calling at scale. ACEBench~\citep{chen2025acebench} focuses on multi-turn agent dialogues. $\tau$-bench~\citep{yao2025taubench} evaluates conversational agents in dual-control environments. ToolAlpaca~\citep{tang2023toolalpaca}, ToolEyes~\citep{ye2025tooleyes}, and related datasets increase tool variety and output-format diversity. These benchmarks primarily test clean conditions: queries are well formed, tool descriptions are stable, action spaces are unambiguous, and tool execution is deterministic. AgentRobustBench uses five of these benchmarks as source data and applies a production-grounded perturbation taxonomy to each.

\textbf{Robustness perturbations for LLMs and tool use}
General LLM robustness work measures sensitivity to typos, paraphrases, and synonym substitutions~\citep{raj2022measuring}. Tool-use robustness work has studied several important failure modes. RoTBench~\citep{ye2024rotbench} perturbs function names and parameters. ToolSandbox~\citep{lu2025toolsandbox} introduces distractor tools and sequencing tests. \citet{rabinovich2025robustness} evaluate query rephrasing and toolkit modifications on BFCL, reporting 13--19\% degradation, with many errors caused by parameter-value mismatches. \citet{faghih2025preferences} show that tool-description wording can strongly affect LLM tool preferences. AgentRobustBench differs in three ways: it organizes perturbations by POMDP component, ties perturbation types to verified production failures or documented empirical effects, and pairs the benchmark with a domain-randomized training recipe.

\textbf{Sim-to-real beyond robotics}
Sim-to-real transfer has been widely studied in robotic control with deep reinforcement learning~\citep{tobin2017domain, peng2018sim, sadeghi2017cad2rl, chebotar2019closing, tan2018sim, akkaya2019solving}, with surveys covering common transfer settings and randomization strategies~\citep{zhao2020sim, hofer2021sim2real}. Our work moves this framing to tool-use language agents. Rather than randomizing physical parameters, we randomize text inputs, tool registries, and reward-relevant metadata in tool-use trajectories, then test whether this reduces deployment-style failures (\S\ref{sec:method}).
\section{AgentRobustBench}
\label{sec:benchmark}

\label{sec:bench-sources}
We build AgentRobustBench on top of five existing tool-use benchmarks: BFCL V3 single-turn~\citep{patil2024gorilla}, API-Bank~\citep{li2023api}, RoTBench~\citep{ye2024rotbench}, ToolAlpaca~\citep{tang2023toolalpaca}, and ToolEyes~\citep{ye2025tooleyes}. We select these because (i) their tasks are released under permissive licenses for redistribution, (ii) together they span the major tool-call output formats (BFCL bracketed AST, JSON \emph{<tool\_call>}, ReAct), and (iii) they are the benchmarks used by the published RL-trained tool-use models~\citep{qian2025toolrl, zhang2025looptool, zhao2025mua, ye2024tl} we evaluate against. Per-source details are in Appendix~\ref{app:bench-stats}.

We retain evaluable single-turn seeds with valid, unique ground-truth tool calls. These 199 evaluation seeds are separate from the ToolRL training corpus (\S\ref{sec:method-method}); selection and diversity checks are detailed in Appendix~\ref{app:additional-validation}.

\subsection{Perturbation taxonomy}
\label{sec:bench-taxonomy}

We organize perturbations by the four elements of the POMDP (\S\ref{sec:bg-mdp}). Within each category we instantiate concrete types targeting distinct failure modes documented in production. Figure~\ref{fig:pipeline} sketches the four categories with one canonical example per type; the full taxonomy is summarized in Table~\ref{tab:taxonomy}. We use short identifiers in figures and tables; the mapping to the longer code-level names is in Appendix~\ref{app:naming}.

\definecolor{cObs}{HTML}{7F77DD}   
\definecolor{cAct}{HTML}{D85A30}   
\definecolor{cRew}{HTML}{1D9E75}   
\definecolor{cTrn}{HTML}{B5872E}   

\newcommand{\modObs}[1]{\cellcolor{cObs!18}#1}
\newcommand{\modAct}[1]{\cellcolor{cAct!18}#1}
\newcommand{\modRew}[1]{\cellcolor{cRew!18}#1}
\newcommand{\modTrn}[1]{\cellcolor{cTrn!18}#1}

\newcommand{\hlObs}[1]{\colorbox{cObs!45}{#1}}
\newcommand{\hlAct}[1]{\colorbox{cAct!45}{#1}}
\newcommand{\hlRew}[1]{\colorbox{cRew!45}{#1}}
\newcommand{\hlTrn}[1]{\colorbox{cTrn!45}{#1}}

\begin{table}[!htb]
  \centering
  \caption{Perturbation taxonomy by POMDP component. The \emph{display name} is the abbreviated identifier used in figures and tables throughout the paper; the corresponding code-level name (used in our released datasets) is in Appendix~\ref{app:naming}. Generation method indicates how each perturbation is produced from a clean source sample.}
  \label{tab:taxonomy}
  \footnotesize
  \setlength{\tabcolsep}{4pt}
  \begin{tabularx}{\linewidth}{l l X l}
  \toprule
  \textbf{Component} & \textbf{Display name} & \textbf{What is modified} & \textbf{Method} \\
  \midrule
  Observation
    & Typo            & \modObs{user query (\hlObs{character-level keyboard noise})}    & LLM \\
    & QueryPara       & \modObs{user query (\hlObs{semantic rephrasing})}               & LLM \\
    & ToolPara        & \modObs{\hlObs{tool} description}                                & LLM \\
    & ParamPara       & \modObs{\hlObs{parameter} description}                           & LLM \\
  \midrule
  Action
    & Dup-NoDesc      & \modAct{inject distractor sharing GT name; \hlAct{no desc}, \hlAct{no params}}    & rule \\
    & Dup-Desc        & \modAct{same name; \hlAct{correct description}, \hlAct{no params}}                & rule \\
    & Dup-WrongP      & \modAct{same name; \hlAct{no description}, \hlAct{wrong params}}                  & rule \\
    & Dup-DescWP      & \modAct{same name; \hlAct{correct description}, \hlAct{wrong params}}             & rule \\
    & Dup-SwapDP      & \modAct{same name; \hlAct{swapped description}, \hlAct{wrong params}}             & rule \\
    & RedunTool       & \modAct{inject \hlAct{functionally similar but incorrect} distractor}     & LLM \\
  \midrule
  Reward
    & MisDesc         & \modRew{\hlRew{misleading description} on GT; \emph{\hlRew{\_Budget/\_Fast}} suffix on distractor} & rule \\
    & TimeDesc        & \modRew{\hlRew{response-time annotation}; same naming pattern}            & rule \\
    & MisDesc-N       & \modRew{MisDesc + \hlRew{neutral} suffix on distractor (\emph{\hlRew{\_1}})}      & rule \\
    & TimeDesc-N      & \modRew{TimeDesc + \hlRew{neutral} suffix}                                & rule \\
    & MisDesc-Abbr    & \modRew{MisDesc + \hlRew{abbreviated GT name}}                             & rule \\
    & TimeDesc-Abbr   & \modRew{TimeDesc + \hlRew{abbreviated GT name}}                           & rule \\
  \midrule
  Transition
    & Timeout         & \modTrn{first tool call returns ``\hlTrn{Tool execution timed out}''}     & runtime \\
    & RateLim         & \modTrn{first call returns \hlTrn{HTTP 429}}                              & runtime \\
    & AuthErr         & \modTrn{first call returns \hlTrn{HTTP 401/403}}                          & runtime \\
    & 5xxErr          & \modTrn{first call returns \hlTrn{HTTP 5xx}}                              & runtime \\
    & Malform         & \modTrn{first call returns \hlTrn{malformed JSON}}                        & runtime \\
    & SchemaD         & \modTrn{first call returns ``\hlTrn{parameter X no longer valid}''}       & runtime \\
  \bottomrule
  \end{tabularx}
\end{table}

\subsection{Production grounding}
\label{sec:bench-grounding}

A perturbation type is worth measuring only if it represents a failure that occurs in deployed systems. We tie each of the 22 types to one or more verified GitHub issues from major agent frameworks (LangChain, AutoGen, OpenAI Agents, LlamaIndex, MCP servers) or to peer-reviewed studies. Every cited issue was re-fetched manually to confirm both URL existence and content match; we collected sources only from tool-calling / agent / MCP / function-calling repositories. Representative examples include: a user typo propagates into a hallucinated tool name; a paraphrased query routes to the wrong tool; two MCP servers register the same tool name and the SDK hangs; LangChain's default \emph{request\_timeout=None} hangs agents forever. The full type$\to$canonical issue map (Table~\ref{tab:grounding}) and audit log are in Appendix~\ref{app:evidence}.

\subsection{Benchmark construction}
\label{sec:bench-procedure}

\textbf{Offline perturbations.} Observation, Action, and Reward perturbations modify static benchmark data. Reward perturbations represent advertised tool costs and response times as text in the tool metadata. LLM-generated types (Typo, QueryPara, ToolPara, ParamPara, RedunTool, and the description rewriting in MisDesc / TimeDesc) use \emph{gpt-5-mini}~\cite{openai_gpt5mini} with the prompts in Appendix~\ref{app:prompts}. The \texttt{realistic\_typos} type is an exception: stronger aligned models tend to silently auto-correct or refuse to inject typos, so we instead use \emph{gpt-4o-mini}~\cite{openai_gpt4omini}, which is more willing to produce lexically-noisy text. Rule-based types (Dup-NoDesc through Dup-SwapDP, and the six reward-variant suffixings) are deterministic. Semantic equivalence of LLM-generated outputs is in Appendix~\ref{app:paraphrase-audit}.

\textbf{Online perturbations.} Transition perturbations modify execution dynamics rather than static data. When the model emits its first tool call, the harness intercepts it and returns a transient-error message (one of timeout, HTTP 429, HTTP 401/403, HTTP 5xx, malformed JSON, or schema drift; strings in Appendix~\ref{app:transition-strings}) in place of the actual tool result, giving the model a chance to recover on its next turn. The scorer evaluates the recovery turn: a model that retries with the correct tool earns full credit, while one that gives up or repeats a wrong call earns zero. We inject the transition perturbation at the first tool call so that every model is tested at the same trajectory step, which isolates the per-error-type effect from confounds introduced by the position at which the error occurs. We use a 100\% injection rate per sample for the same reason.

The clean baseline contains 199 samples drawn from the five source benchmarks. Applying the 22 perturbation types to whichever sources support each yields 3{,}522 perturbation samples (ToolEyes is the only exception: it lacks a single ground-truth tool with a distractor structure and therefore skips the 6 Action and 6 Reward types, while still contributing Observation and Transition). Overall, the whole evaluation set comprises 3{,}721 samples, and the full breakdown is in Appendix~\ref{app:bench-stats}.

\section{ToolRL-DR: Domain Randomization for Tool-Use RL}
\label{sec:method}

\subsection{Motivation}
\label{sec:method-motivation}

Tool-use agents fail under realistic perturbations they did not see at training time (\S\ref{sec:bench-grounding}). Motivated by domain randomization in robotics~\citep{tobin2017domain, peng2018sim, sadeghi2017cad2rl}, we apply the same principle to tool-use RL: replace clean training trajectories with perturbation-augmented trajectories. We hypothesize that this closes the gap on the perturbation categories that are expressible as static training-data edits (observation, action, reward) but not on transition perturbations, which manifest only as runtime tool-execution responses and cannot be expressed in static data without a custom retry-aware rollout environment. We deliberately exclude transition perturbations from training: if RL on the three statically-augmentable categories alone improves Transition robustness, that is evidence of behavioral transfer rather than curriculum-specific overfitting. This setup is what allows the surprising 27\% Transition gap closure we report in \S\ref{sec:exp-paradigm} to be read as transfer rather than supervision.

\subsection{Training}
\label{sec:method-method}

We start from the publicly released ToolRL recipe~\citep{qian2025toolrl}: GRPO~\citep{shao2024deepseekmath} on Qwen2.5-3B-Instruct~\citep{qwen2025qwen25technicalreport} with structured rewards on tool-name and parameter correctness. The modification is the training data: we replace the 4{,}000 clean samples in ToolRL's \texttt{rlla\_4k} corpus with perturbation-augmented samples drawn uniformly across the 16 statically-augmentable types (4 observation + 6 action + 6 reward); transition perturbations are excluded by construction (\S\ref{sec:method-motivation}). We deliver two checkpoints, both starting from \texttt{Qwen/Qwen2.5-3B-Instruct} with identical code and hyperparameters but different data composition. \textbf{ToolRL-DR-Full} replaces every clean trajectory with a perturbed counterpart (3{,}984 total: 3{,}905 train + 79 val). \textbf{ToolRL-DR-Mixed} keeps roughly half the clean distribution alongside the perturbations (4{,}000 total: 2{,}006 clean / 1{,}994 perturbed; the same 79 validation samples are reused). We use the validation set to select checkpoints. On the clean evaluation set (setup in \S\ref{sec:exp-setup}; Table~\ref{tab:main}) DR-Full's Clean accuracy ($0.643\pm0.065$) is statistically indistinguishable from both DR-Mixed ($0.658\pm0.065$) and the public ToolRL checkpoint at the same backbone ($0.638\pm0.065$, hereafter \textbf{ToolRL-Clean}: \texttt{chengq9/ToolRL-Qwen2.5-3B}). All other training details (GRPO objective, KL coefficient, learning rate, batch size, optimizer, hardware, wall-clock) follow the public ToolRL repository unchanged and are listed in Appendix~\ref{app:compute} (Tables~\ref{tab:hyperparams}--\ref{tab:hyperparams-runs}).

\section{Experiments}
\label{sec:exp}

\subsection{Setup}
\label{sec:exp-setup}

We evaluate 21 models grouped into five families: (1) \emph{RL-trained tool models}: the ToolRL family on Qwen2.5-1.5B/3B and Llama-3.2-3B~\citep{qian2025toolrl}, SFT-Clean-4k-Qwen2.5-3B, TL-CodeLLaMA-2~\citep{ye2024tl}, LoopTool 8B/32B~\citep{zhang2025looptool}, MUA-RL 8B/14B/32B~\citep{zhao2025mua}; (2) \emph{base / instruct counterparts}: Qwen2.5-1.5B/3B-Instruct, Llama-3.2-3B-Instruct, Qwen3-8B/14B/32B in non-thinking mode~\citep{yang2025qwen3}; (3) \emph{frontier reasoning}: DeepSeek-R1-Distill-Qwen-14B~\citep{deepseek2025r1}, Qwen3.5-9B; (4) \emph{closed-source frontier}: o4-mini via the OpenAI Chat Completions API; (5) \emph{our trained checkpoints}: ToolRL-DR-Full and ToolRL-DR-Mixed (\S\ref{sec:method}). Each model is served by vLLM~\citep{vllm2023} at temperature 0 in the inference mode that matches its training paradigm; per-model serving and decoding parameters are in Appendix~\ref{app:eval-details}. We evaluate the clean baseline (199 samples) and all 22 perturbation types under the generation/injection procedure of \S\ref{sec:bench-procedure}, scored by deterministic rule-based parsers (no LLM judge; Appendix~\ref{app:scoring}). All accuracies and drops are reported as mean $\pm$ 95\% percentile-bootstrap half-width on per-sample binary correctness ($B=10{,}000$ resamples; details in table captions).

\subsection{Main result: an uneven sim-to-real gap}
\label{sec:exp-main}

\begin{table}[!htb]
\centering
\caption{Main results across 21 models and 22 perturbation types. \emph{Clean} (199 unperturbed samples) and \emph{Pert.\ Acc.} (sample-weighted average over all perturbed samples, clean excluded) are accuracies in $[0,1]$ (\emph{higher = better}). The four $\Delta$ columns are per-POMDP-component robustness gaps on the same scale: $\Delta_X = \text{Clean} - \text{mean}(\text{Acc on }X\text{-perturbations})$ (\emph{smaller = better}). Each entry is mean\,$\pm$\,half-width of the 95\% percentile-bootstrap CI over per-sample correctness scores ($B={\hspace{-0.05em}}10{,}000$; decoding is deterministic, so the only variance source is finite-sample uncertainty across benchmark items). Rows are sorted within each group by Pert.\ Acc.\ descending; \textbf{bold} marks our two trained checkpoints. Superscripts on our rows denote paired-bootstrap significance vs.\ the public \textbf{ToolRL-Clean} 3B baseline ($^*p<0.05$, $^{**}p<0.01$, $^{***}p<0.001$; no marker = not significant at $p<0.05$).}
\label{tab:main}
\scriptsize
\setlength{\tabcolsep}{3pt}
\begin{tabular*}{\linewidth}{@{\extracolsep{\fill}}lrrrrrr}
\toprule
 & Clean $\uparrow$ & Pert.\ Acc.\ $\uparrow$ & $\Delta_{\mathrm{Obs}}$ $\downarrow$ & $\Delta_{\mathrm{Act}}$ $\downarrow$ & $\Delta_{\mathrm{Rew}}$ $\downarrow$ & $\Delta_{\mathrm{Trn}}$ $\downarrow$ \\
\midrule
\multicolumn{7}{l}{\emph{Our 3B trained checkpoints}} \\
\textbf{ToolRL-DR-Full (ours)} & \textbf{0.643$\pm$0.065} & \textbf{0.463$\pm$0.016$^{***}$} & \textbf{0.009$\pm$0.075} & \textbf{0.147$\pm$0.074} & \textbf{0.331$\pm$0.076$^{***}$} & \textbf{0.231$\pm$0.071$^{***}$} \\
\textbf{ToolRL-DR-Mixed (ours)} & \textbf{0.658$\pm$0.065} & \textbf{0.461$\pm$0.017$^{***}$} & \textbf{0.049$\pm$0.075} & \textbf{0.152$\pm$0.075} & \textbf{0.364$\pm$0.074$^{***}$} & \textbf{0.232$\pm$0.072$^{***}$} \\
\midrule
\multicolumn{7}{l}{\emph{Open-source RL tool models, 8B--14B}} \\
MUA-RL-8B & 0.678$\pm$0.065 & 0.479$\pm$0.016 & 0.023$\pm$0.073 & 0.156$\pm$0.072 & 0.401$\pm$0.073 & 0.229$\pm$0.070 \\
LoopTool-8B & 0.714$\pm$0.063 & 0.476$\pm$0.017 & 0.049$\pm$0.070 & 0.173$\pm$0.072 & 0.429$\pm$0.071 & 0.296$\pm$0.069 \\
MUA-RL-14B & 0.628$\pm$0.068 & 0.470$\pm$0.017 & 0.011$\pm$0.075 & 0.090$\pm$0.077 & 0.334$\pm$0.075 & 0.202$\pm$0.073 \\
\midrule
\multicolumn{7}{l}{\emph{Open-source RL tool models, 1.5B--7B}} \\
ToolRL-Qwen2.5-1.5B & 0.704$\pm$0.063 & 0.432$\pm$0.016 & 0.033$\pm$0.072 & 0.174$\pm$0.072 & 0.481$\pm$0.071 & 0.379$\pm$0.070 \\
TL-CodeLLaMA-2 & 0.653$\pm$0.065 & 0.405$\pm$0.016 & 0.024$\pm$0.075 & 0.166$\pm$0.073 & 0.456$\pm$0.073 & 0.333$\pm$0.072 \\
ToolRL-Qwen2.5-3B & 0.638$\pm$0.065 & 0.404$\pm$0.016 & 0.020$\pm$0.075 & 0.143$\pm$0.075 & 0.450$\pm$0.073 & 0.315$\pm$0.071 \\
ToolRL-Llama3.2-3B & 0.628$\pm$0.065 & 0.390$\pm$0.016 & 0.025$\pm$0.075 & 0.171$\pm$0.075 & 0.463$\pm$0.073 & 0.296$\pm$0.074 \\
\midrule
\multicolumn{7}{l}{\emph{Open-source RL tool models, 32B}} \\
MUA-RL-32B & 0.683$\pm$0.065 & 0.527$\pm$0.016 & 0.028$\pm$0.072 & -0.010$\pm$0.071 & 0.397$\pm$0.072 & 0.218$\pm$0.071 \\
LoopTool-32B & 0.779$\pm$0.058 & 0.508$\pm$0.017 & 0.034$\pm$0.065 & 0.120$\pm$0.066 & 0.509$\pm$0.067 & 0.396$\pm$0.063 \\
\midrule
\multicolumn{7}{l}{\emph{Base / instruct counterparts}} \\
Qwen3-14B & 0.749$\pm$0.060 & 0.529$\pm$0.016 & 0.039$\pm$0.068 & 0.143$\pm$0.069 & 0.464$\pm$0.070 & 0.250$\pm$0.067 \\
Qwen3-32B & 0.754$\pm$0.058 & 0.524$\pm$0.017 & 0.024$\pm$0.067 & 0.104$\pm$0.068 & 0.366$\pm$0.070 & 0.377$\pm$0.066 \\
Qwen3-8B & 0.673$\pm$0.065 & 0.448$\pm$0.016 & 0.019$\pm$0.075 & 0.171$\pm$0.073 & 0.409$\pm$0.072 & 0.293$\pm$0.070 \\
Qwen2.5-3B-Instruct & 0.608$\pm$0.070 & 0.390$\pm$0.016 & 0.014$\pm$0.076 & 0.195$\pm$0.075 & 0.446$\pm$0.074 & 0.237$\pm$0.073 \\
Qwen2.5-1.5B-Instruct & 0.573$\pm$0.070 & 0.367$\pm$0.016 & 0.040$\pm$0.077 & 0.200$\pm$0.077 & 0.405$\pm$0.074 & 0.205$\pm$0.073 \\
Llama-3.2-3B-Instruct & 0.518$\pm$0.070 & 0.341$\pm$0.016 & 0.009$\pm$0.077 & 0.177$\pm$0.076 & 0.372$\pm$0.073 & 0.174$\pm$0.073 \\
\midrule
\multicolumn{7}{l}{\emph{Frontier (open weights, reasoning / general)}} \\
DeepSeek-R1-Distill-14B & 0.618$\pm$0.065 & 0.442$\pm$0.016 & 0.013$\pm$0.077 & 0.160$\pm$0.075 & 0.318$\pm$0.074 & 0.213$\pm$0.073 \\
Qwen3.5-9B & 0.709$\pm$0.063 & 0.437$\pm$0.016 & 0.040$\pm$0.071 & 0.168$\pm$0.072 & 0.411$\pm$0.073 & 0.417$\pm$0.069 \\
\midrule
\multicolumn{7}{l}{\emph{Closed-source frontier}} \\
o4-mini (OpenAI) & 0.709$\pm$0.063 & 0.502$\pm$0.017 & 0.049$\pm$0.071 & 0.157$\pm$0.071 & 0.326$\pm$0.072 & 0.276$\pm$0.070 \\
\midrule
\multicolumn{7}{l}{\emph{ToolRL SFT version}} \\
SFT-Clean-4k-3B & 0.598$\pm$0.068 & 0.267$\pm$0.014 & 0.019$\pm$0.075 & 0.184$\pm$0.076 & 0.443$\pm$0.074 & 0.577$\pm$0.068 \\
\bottomrule
\end{tabular*}
\end{table}

\begin{figure}[!htb]
    \centering
    \includegraphics[width=1.0\linewidth]{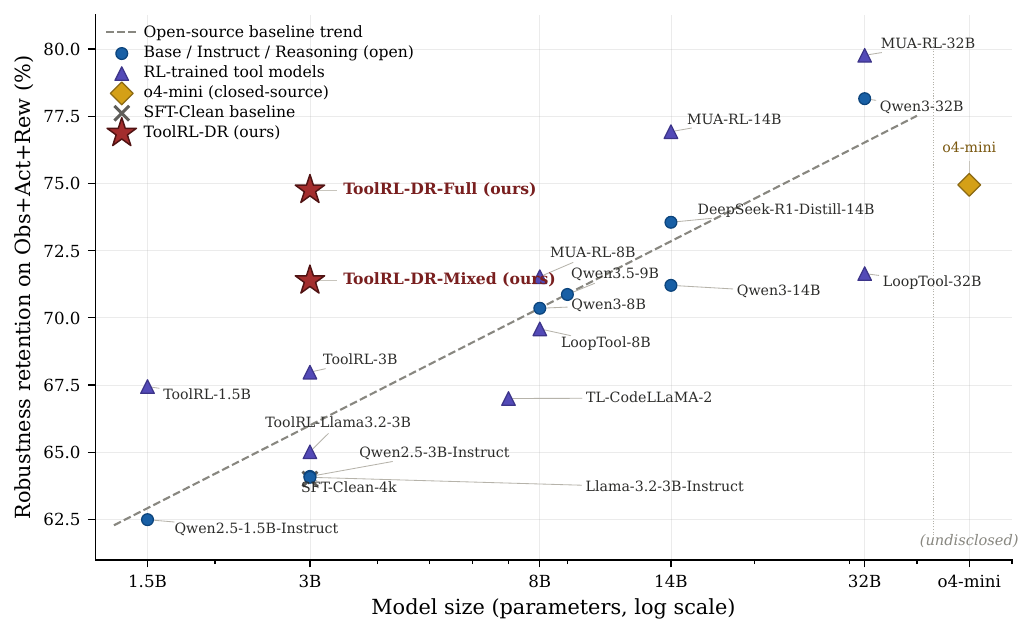}
    \vspace{-20pt}
    \caption{Robustness retention on Observation, Action, and Reward vs.\ model size, summarising Table~\ref{tab:main}. Retention $= 1 - \mathrm{mean}(\Delta_{\mathrm{Obs}}, \Delta_{\mathrm{Act}}, \Delta_{\mathrm{Rew}})/\mathrm{Clean}$ (higher is better). Our 3B \textbf{ToolRL-DR-Full} (red star) sits above the size-matched ToolRL baseline and the log-linear trend over 18 open/closed models, comparable to the 32B function-calling models we evaluate.}
    \label{fig:hero}
    \vspace{-10pt}
\end{figure}

Across the 21 evaluated models, the sim-to-real gap is severe and uneven. Transition and reward perturbations cause the largest drops (per-component averages near 30\% and 40\% pp respectively; Table~\ref{tab:main} columns $\Delta_{\mathrm{Trn}}$ / $\Delta_{\mathrm{Rew}}$), while observation perturbations are absorbed by every model ($\Delta_{\mathrm{Obs}}<5\%$). Action perturbations sit in the middle (10--20\% pp), driven mostly by same-name distractor difficulty; a few of the largest models show negative $\Delta_{\mathrm{Act}}$ because the distractors actually help disambiguation. Within Reward, the abbreviation variants (MisDesc-Abbr, TimeDesc-Abbr) consistently produce the largest drops; within Transition, variance across error types (Timeout vs.\ AuthErr vs.\ SchemaD) is small relative to variance across models. Per-perturbation breakdown and a heat-map view of all 21$\times$22 (model, perturbation type) drops are in Appendices~\ref{app:per-pert} and~\ref{app:heatmap}.

\textbf{Error-mode classification.} We classify each scored-incorrect prediction by inspecting the raw model output before the format-tolerant parser (rubric in Appendix~\ref{app:error-modes}). On transition perturbations the dominant failure modes for ToolRL-Clean (across all 808 failed transition samples in the union of the six transient-error variants) are \emph{wrong tool call} (53.0\%, 428/808; the model emitted a parseable call but with the wrong name or parameters, typically retrying the same tool with a slight argument tweak) and \emph{omitted tool call} (47.0\%, 380/808; the model produced text only, often saying ``the tool seems to have failed; please try again later''). \emph{Empty tool call} accounts for $<$0.1\% (the model never goes silent on this benchmark); the model always produces \emph{some} output but half the time gives up on the second-pass invocation. Our ToolRL-DR-Full reduces the omitted-call rate to 34.2\% ($-12.8$\,pp) while raising the wrong-call rate to 65.8\% ($+12.8$\,pp), with overall transition failures dropping from 808 to 702; we use this contrast to explain the transition transfer in \S\ref{sec:exp-paradigm}. On reward perturbations the dominant failure is \emph{name bias}: the model selects a distractor whose name suggests efficiency despite a description listing higher cost. Three verbatim case studies illustrating one Transition, one Reward, and one Observation failure are in Appendix~\ref{app:case-studies}.

\subsection{Scaling does not close the gap}
\label{sec:exp-scaling}

\begin{figure}[!htb]
    \centering
    \includegraphics[width=1.0\linewidth]{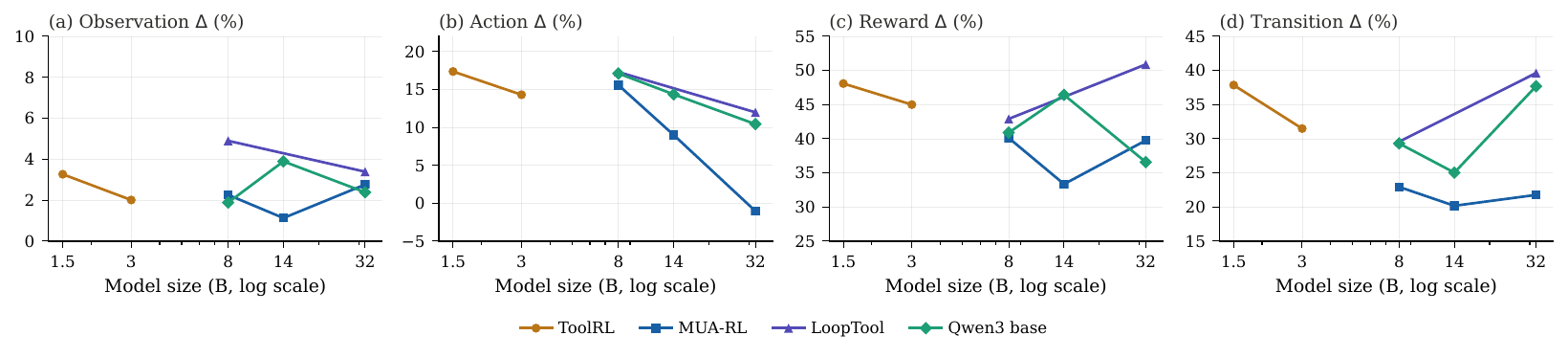}
    \vspace{-20pt}
    \caption{Per-POMDP-component drop $\Delta$ vs.\ model size for four families (ToolRL 1.5B/3B, MUA-RL 8B/14B/32B, LoopTool 8B/32B, Qwen3 base 8B/14B/32B). Smaller $\Delta$ = better. Observation drops stay small ($<$5\%) and Action drops trend slightly downward with size, but Reward (c) and Transition (d) stay within the same band as size grows from 8B to 32B, motivating training-side interventions in \S\ref{sec:method}.}
    \label{fig:scaling}
\end{figure}

We compare three Qwen3-based RL families across model sizes: MUA-RL (8B/14B/32B), LoopTool (8B/32B), Qwen3 base (8B/14B/32B), and report performance versus model size (Figure~\ref{fig:scaling}).

Clean accuracy improves with size within every family, as expected. The drops, however, are largely flat or even \emph{increase} with size for transition and reward perturbations. LoopTool-32B has a 50.9\% Reward drop and 39.6\% Transition drop, larger than its 8B sibling on both axes; Qwen3-32B has a 37.7\% Transition drop, larger than Qwen3-8B's 29.3\%. The MUA-RL family is the only one in which the largest model has the smallest drops, but even there the residual reward and transition gaps are 40\% and 22\%. Robustness, in this taxonomy, is not a small-model artifact: 32B-scale RL-trained tool models still fail in the same patterns as their 8B siblings. SFT-Clean-4k-Qwen2.5-3B (Table~\ref{tab:main}, the supervised counterpart of ToolRL at the same backbone) has a Transition drop ($\Delta_{\mathrm{Trn}}=57.7\%$) roughly $2\times$ that of any RL-trained model, a gap that motivates rollout-based training in general.

\subsection{Domain randomization narrows the gap}
\label{sec:exp-dr}

\textbf{Training-distribution comparison on the 3B backbone.}\label{sec:exp-paradigm} \textbf{ToolRL-Clean}, our \textbf{ToolRL-DR-Full}, and our \textbf{ToolRL-DR-Mixed} share the Qwen2.5-3B-Instruct backbone and differ only in the GRPO training distribution (\S\ref{sec:method}). ToolRL-DR-Full closes Reward by 11.9\,pp ($\approx$26\% of ToolRL-Clean's Reward gap) and \emph{unexpectedly} closes Transition by 8.4\,pp ($\approx$27\%), despite no Transition perturbation in training. This is our most surprising finding: perturbed-RL training transfers some robustness to runtime errors it never saw. DR-Mixed lies between Clean and DR-Full on Reward and Obs, reproducing a dose response: more perturbation in training $\to$ smaller drops at evaluation. The Action axis is invariant across all three RL conditions, suggesting same-name distractor disambiguation is bounded by model capacity rather than training-data composition. Figure~\ref{fig:hero} visualizes this against the broader 1.5B--32B trend on the Obs+Act+Rew axes.

\textbf{Why Transition transfers: a more persistent retry policy.}\label{sec:method-no-transition} We attribute transfer to a more persistent retry policy under adversarial inputs. Re-running the failure-mode classification of \S\ref{sec:exp-main} on ToolRL-DR-Full's six transient-error variants yields 702 failed transition samples (down from 808 for ToolRL-Clean), with omitted-call rate dropping from 47.0\% to 34.2\% ($-12.8$\,pp) and wrong-call rate rising from 53.0\% to 65.8\%. DR-Full converts some giveups into successful retries (total failures down 13\%) and retries more often when it does fail. We do not claim the gap is closed: $\Delta_{\mathrm{Trn}}\!\approx\!0.23$ remains, and a retry-aware reward (e.g.\ rollouts with error responses~\citep{snell2024scaling, brown2024large}) is a natural next step.

To test whether training on other perturbations helps on Reward cases, we trained on Observation and Action perturbations and evaluated on Reward perturbations. Reward accuracy rose from 18.8\% for ToolRL-Clean to 23.0\%. With Reward perturbations also included in a training set of the same size, accuracy reached 30.3\% (Appendix~\ref{app:reward-transfer}).

\subsection{Live evaluation platform}
\label{sec:platform}

A public AgentRobustBench leaderboard hosted as a HuggingFace Space (\url{https://huggingface.co/spaces/willchow66/AgentRobustBench}; Appendix~\ref{app:leaderboard}) lets contributors upload predictions JSONs that are scored server-side with the same deterministic scorer used in this paper (\S\ref{sec:exp-setup}). The leaderboard includes results for all 21 evaluated models.

\section{Limitations and future work}
\label{sec:limitations}

\textbf{Multi-turn coverage.} Our evaluations focus on single-turn settings. Multi-turn agents 
can sometimes recover via follow-up clarification but also accumulate errors over horizon, and transition perturbations are hard to inject cleanly into multi-turn rollouts that already contain tool-execution loops. The perturbation taxonomy carries over directly to future extension to multi-turn.

\textbf{Method scale.} In this work, ToolRL-DR is trained on a 3B backbone with 4{,}000 samples to match the public ToolRL recipe and enable apples-to-apples comparison. Whether the recipe scales to 14B/32B without saturation, and whether more training data improves headroom, are open.

\textbf{Transition robustness.} Our method partially reduces, but does not eliminate, the Transition gap without training on runtime errors. Doing so likely requires either RL with rollouts that include error responses and a retry-aware reward, or test-time strategies (self-consistency, retry policies) that operate independently of training.

\section{Conclusion}
\label{sec:conclusion}

We frame tool-use robustness as a sim-to-real problem and introduce a benchmark that perturbs the observation, action, reward-relevant metadata, and transition components of the tool-use POMDP. Across 21 models, the resulting gap is large and uneven: observation perturbations are mostly absorbed, whereas reward-relevant and transition perturbations remain difficult, and increasing model size does not consistently close these gaps in our evaluation. We further show that domain-randomized RL on perturbation-augmented tool-use trajectories improves robustness on a 3B backbone, with the performance improvement on reward-relevant perturbations and partial transfer to held-out transition failures. Notably, our 3B checkpoint closes approximately 27\% of the Transition gap without ever training on transition perturbations, suggesting that RL on adversarial static inputs induces a more persistent retry policy that generalizes to runtime failures. These results suggest that tool-use reliability should be evaluated not only under clean tool registries and deterministic execution, but also under deployment-style metadata noise, action-space ambiguity, and runtime errors. 

\section*{Acknowledgments and Disclosures}
The work was partially supported by Cisco, Toyota, NSF awards \#2442477, \#2536297, \#2550203, \#2613637,
\#2616632 and \#2623317. 
We ran the evaluations on DeltaAI at the National Center for Supercomputing Applications (NCSA) using allocation CIS261701 from the Advanced Cyberinfrastructure Coordination Ecosystem: Services \& Support (ACCESS) program~\citep{boerner2023access}. The U.S. National Science Foundation supports ACCESS through grants \#2138259, \#2138286, \#2138307, \#2137603, and \#2138296. DeltaAI is supported by the National Science Foundation (award OAC~2320345) and the State of Illinois.
The views and conclusions in this paper should not be interpreted as
representing any funding agencies.
The authors declare no competing interests.

\bibliographystyle{plainnat}
\bibliography{preprint}

\appendix

\section{Per-source benchmark statistics}
\label{app:bench-stats}

Table~\ref{tab:sources} gives the full per-source composition of AgentRobustBench: clean sub-sample counts, average evaluable samples per perturbation type, totals, average candidate-tool counts, and the structured output format expected by each source's original scorer.

\begin{table}[H]
\centering
\caption{Per-source composition of AgentRobustBench. \emph{Clean} is the sub-sample count we use as baseline. \emph{Avg./type} is the mean evaluable-sample count per (source, perturbation type) pair, smaller than \emph{Clean} when some perturbations do not apply (only ToolEyes hits this, on the 12 Action and Reward types). \emph{Total} is the per-source sum across the clean baseline plus all 22 perturbation types, aggregating to 3{,}721 samples. \emph{Avg.\ tools} is the mean candidate-tool count per sample (clean slice). \emph{Output format} lists the structured form expected by the original scorer.}
\label{tab:sources}
\footnotesize
\setlength{\tabcolsep}{4pt}
\begin{tabularx}{\linewidth}{l r r r r X}
\toprule
\textbf{Benchmark} & \textbf{Clean} & \textbf{Avg./type} & \textbf{Total} & \textbf{Avg.\ tools} & \textbf{Output format} \\
\midrule
BFCL V3 (ST)   & 32  & 22.0 & 505  & 2.8 & \emph{[func(a=1)]} (BFCL AST) \\
API-Bank       & 74  & 71.6 & 1646 & 3.1 & \emph{<tool\_call>\{json\}</tool\_call>} \\
RoTBench       & 21  & 20.8 & 479  & 7.8 & \emph{Action: X / Action Input: \{json\}} \\
ToolAlpaca     & 21  & 20.8 & 479  & 3.5 & mixed (ReAct, XML, JSON, bare Python) \\
ToolEyes       & 51  & 25.5 & 612  & 6.3 & ReAct \\
\midrule
\textbf{Total} & \textbf{199} & \textbf{160.0} & \textbf{3721} & --- & --- \\
\bottomrule
\end{tabularx}
\end{table}

\section{Per-(model, perturbation) drop heatmap}
\label{app:heatmap}

\begin{figure}[H]
    \centering
    \includegraphics[width=\linewidth]{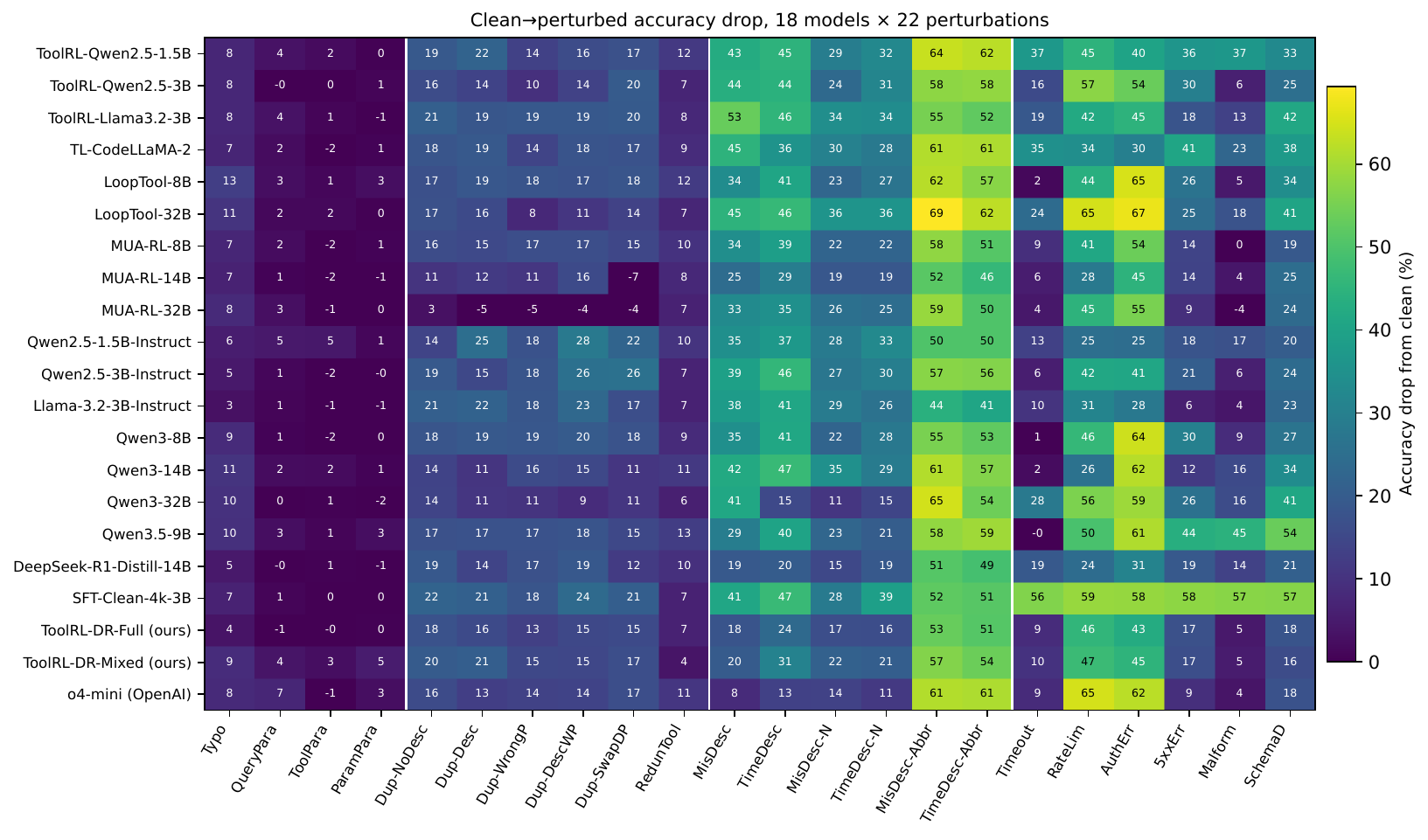}
    \caption{Per-(model, perturbation) accuracy drop from clean across all 21 evaluated models and 22 perturbations. Rows: model families. Columns: 22 perturbation types grouped by POMDP component (white vertical lines). Cell color and annotation = percentage-point drop in accuracy from the model's own clean baseline. The visual takeaway is that drops are concentrated on the right two POMDP components (Reward, Transition) regardless of model size or family, and that our ToolRL-DR-Full / -Mixed rows show smaller drops on the Reward block than every other 3B model.}
    \label{fig:heatmap}
\end{figure}

Figure~\ref{fig:heatmap} consolidates the per-perturbation breakdown (Appendix~\ref{app:per-pert}) into one heatmap, showing every (model, perturbation) accuracy drop on a single page. Rows are the 21 evaluated models grouped by family; columns are the 22 perturbation types grouped by POMDP component (white vertical lines). Cell color and annotation are the percentage-point drop in accuracy from the model's own clean baseline (larger = more brittle).

\section{One example per perturbation}
\label{app:examples}

Each perturbation type is illustrated below by a clean source sample (drawn from the corresponding source benchmark) paired with its perturbed counterpart, so that the reader can inspect what each type actually changes in the data.

Each subtype below is illustrated with one canonical (clean, perturbed) pair drawn from the released dataset. We pick the same sample id across perturbations where possible to make the visual diff easier; transition perturbations operate on clean samples at runtime and are illustrated with the injected error string.

\subsection*{Typo \emph{(Observation, code: \texttt{realistic\_typos})}}
\textbf{Sample id:} \texttt{bfcl\_v3\_\_BFCL\_v3\_multiple\_\_multiple\_110} (\emph{bfcl\_v3})

\begin{itemize}[leftmargin=*, noitemsep]
  \item \textbf{User query (clean, divergence):} \emph{Find the type of gene mutation based on SNP (Single Nucleotide Polymorphism) ID rs6034464.}
  \item \textbf{User query (perturbed, divergence):} \emph{Find teh type of gene mutaiton based on SNP (Single Nucleotide Polymorpism) ID rs6034464.}
  \item \textbf{Tools (unchanged):} \texttt{get\_collectables\_in\_season, mutation\_type.find}
  \item \textbf{GT (clean):} \texttt{[\{"name": "mutation\_type.find", "parameters": \{"snp\_id": "rs6034464", "species": "Homo sapiens"\}\}]}
\end{itemize}

\subsection*{QueryPara \emph{(Observation, code: \texttt{query\_paraphrase})}}
\textbf{Sample id:} \texttt{bfcl\_v3\_\_BFCL\_v3\_multiple\_\_multiple\_110} (\emph{bfcl\_v3})

\begin{itemize}[leftmargin=*, noitemsep]
  \item \textbf{User query (clean, divergence):} \emph{Find the type of gene mutation based on SNP (Single Nucleotide Polymorphism) ID rs6034464.}
  \item \textbf{User query (perturbed, divergence):} \emph{Identify the kind of gene mutation associated with SNP (Single Nucleotide Polymorphism) ID rs6034464.}
  \item \textbf{Tools (unchanged):} \texttt{get\_collectables\_in\_season, mutation\_type.find}
  \item \textbf{GT (clean):} \texttt{[\{"name": "mutation\_type.find", "parameters": \{"snp\_id": "rs6034464", "species": "Homo sapiens"\}\}]}
\end{itemize}

\subsection*{ToolPara \emph{(Observation, code: \texttt{paraphrase\_tool\_description})}}
\textbf{Sample id:} \texttt{bfcl\_v3\_\_BFCL\_v3\_multiple\_\_multiple\_110} (\emph{bfcl\_v3})

\begin{itemize}[leftmargin=*, noitemsep]
  \item \textbf{User query:} \emph{Find the type of gene mutation based on SNP (Single Nucleotide Polymorphism) ID rs6034464.}
  \item \textbf{Tools (unchanged):} \texttt{get\_collectables\_in\_season, mutation\_type.find}
  \item \textbf{Tool entry (clean):} \texttt{\{"name": "get\_collectables\_in\_season", "description": "Retrieve a list of collectable items in a specific game during a specified season.", "parameters": \{"type": "dict", "properties": \{"game\_name": \{"type": "string", "description": "Name of the game."\}, "season": \{"type": "strin}
  \item \textbf{Tool entry (perturbed):} \texttt{\{"name": "get\_collectables\_in\_season", "description": "Obtain a list of collectible items available in a certain game for a designated season.", "parameters": \{"type": "dict", "properties": \{"game\_name": \{"type": "string", "description": "Name of the game."\}, "season": \{"type": "}
  \item \textbf{GT (clean):} \texttt{[\{"name": "mutation\_type.find", "parameters": \{"snp\_id": "rs6034464", "species": "Homo sapiens"\}\}]}
\end{itemize}

\subsection*{ParamPara \emph{(Observation, code: \texttt{paraphrase\_parameter\_description})}}
\textbf{Sample id:} \texttt{bfcl\_v3\_\_BFCL\_v3\_multiple\_\_multiple\_110} (\emph{bfcl\_v3})

\begin{itemize}[leftmargin=*, noitemsep]
  \item \textbf{User query:} \emph{Find the type of gene mutation based on SNP (Single Nucleotide Polymorphism) ID rs6034464.}
  \item \textbf{Tools (unchanged):} \texttt{get\_collectables\_in\_season, mutation\_type.find}
  \item \textbf{Tool entry (clean):} \texttt{\{"name": "get\_collectables\_in\_season", "description": "Retrieve a list of collectable items in a specific game during a specified season.", "parameters": \{"type": "dict", "properties": \{"game\_name": \{"type": "string", "description": "Name of the game."\}, "season": \{"type": "strin}
  \item \textbf{Tool entry (perturbed):} \texttt{\{"name": "get\_collectables\_in\_season", "description": "Retrieve a list of collectable items in a specific game during a specified season.", "parameters": \{"type": "dict", "properties": \{"game\_name": \{"type": "string", "description": "Title of the game."\}, "season": \{"type": "stri}
  \item \textbf{GT (clean):} \texttt{[\{"name": "mutation\_type.find", "parameters": \{"snp\_id": "rs6034464", "species": "Homo sapiens"\}\}]}
\end{itemize}

\subsection*{Dup-NoDesc \emph{(Action, code: \texttt{same\_name\_A})}}
\textbf{Sample id:} \texttt{apibank\_\_level1\_101} (\emph{apibank})

\begin{itemize}[leftmargin=*, noitemsep]
  \item \textbf{User query:} \emph{**Dialogue Records History** <user>Can you help me modify an alarm for user3 at 2023-03-24 09:00:00?</user> <response>Sure, to modify an alarm, I need to authenticate the user. Can...}
  \item \textbf{Tools (clean):} \texttt{ModifyAlarm, GetUserToken, AddAgenda, **Think}
  \item \textbf{Tools (perturbed):} \texttt{ModifyAlarm, GetUserToken, AddAgenda, AddAgenda, **Think}
  \item \textbf{GT (clean):} \texttt{[\{"name": "AddAgenda", "parameters": \{"token": "p9o8i7u6y5t4r3e2w1q", "content": "Lunch with friends", "time": "2023-03-24 14:00:00", "location": "Restaurant X"\}\}]}
\end{itemize}

\subsection*{Dup-Desc \emph{(Action, code: \texttt{same\_name\_B})}}
\textbf{Sample id:} \texttt{apibank\_\_level1\_101} (\emph{apibank})

\begin{itemize}[leftmargin=*, noitemsep]
  \item \textbf{User query:} \emph{**Dialogue Records History** <user>Can you help me modify an alarm for user3 at 2023-03-24 09:00:00?</user> <response>Sure, to modify an alarm, I need to authenticate the user. Can...}
  \item \textbf{Tools (clean):} \texttt{ModifyAlarm, GetUserToken, AddAgenda, **Think}
  \item \textbf{Tools (perturbed):} \texttt{ModifyAlarm, GetUserToken, AddAgenda, AddAgenda, **Think}
  \item \textbf{GT (clean):} \texttt{[\{"name": "AddAgenda", "parameters": \{"token": "p9o8i7u6y5t4r3e2w1q", "content": "Lunch with friends", "time": "2023-03-24 14:00:00", "location": "Restaurant X"\}\}]}
\end{itemize}

\subsection*{Dup-WrongP \emph{(Action, code: \texttt{same\_name\_C})}}
\textbf{Sample id:} \texttt{apibank\_\_level1\_101} (\emph{apibank})

\begin{itemize}[leftmargin=*, noitemsep]
  \item \textbf{User query:} \emph{**Dialogue Records History** <user>Can you help me modify an alarm for user3 at 2023-03-24 09:00:00?</user> <response>Sure, to modify an alarm, I need to authenticate the user. Can...}
  \item \textbf{Tools (clean):} \texttt{ModifyAlarm, GetUserToken, AddAgenda, **Think}
  \item \textbf{Tools (perturbed):} \texttt{ModifyAlarm, AddAgenda, GetUserToken, AddAgenda, **Think}
  \item \textbf{GT (clean):} \texttt{[\{"name": "AddAgenda", "parameters": \{"token": "p9o8i7u6y5t4r3e2w1q", "content": "Lunch with friends", "time": "2023-03-24 14:00:00", "location": "Restaurant X"\}\}]}
\end{itemize}

\subsection*{Dup-DescWP \emph{(Action, code: \texttt{same\_name\_D})}}
\textbf{Sample id:} \texttt{apibank\_\_level1\_101} (\emph{apibank})

\begin{itemize}[leftmargin=*, noitemsep]
  \item \textbf{User query:} \emph{**Dialogue Records History** <user>Can you help me modify an alarm for user3 at 2023-03-24 09:00:00?</user> <response>Sure, to modify an alarm, I need to authenticate the user. Can...}
  \item \textbf{Tools (clean):} \texttt{ModifyAlarm, GetUserToken, AddAgenda, **Think}
  \item \textbf{Tools (perturbed):} \texttt{ModifyAlarm, GetUserToken, AddAgenda, AddAgenda, **Think}
  \item \textbf{GT (clean):} \texttt{[\{"name": "AddAgenda", "parameters": \{"token": "p9o8i7u6y5t4r3e2w1q", "content": "Lunch with friends", "time": "2023-03-24 14:00:00", "location": "Restaurant X"\}\}]}
\end{itemize}

\subsection*{Dup-SwapDP \emph{(Action, code: \texttt{same\_name\_E})}}
\textbf{Sample id:} \texttt{apibank\_\_level1\_101} (\emph{apibank})

\begin{itemize}[leftmargin=*, noitemsep]
  \item \textbf{User query:} \emph{**Dialogue Records History** <user>Can you help me modify an alarm for user3 at 2023-03-24 09:00:00?</user> <response>Sure, to modify an alarm, I need to authenticate the user. Can...}
  \item \textbf{Tools (clean):} \texttt{ModifyAlarm, GetUserToken, AddAgenda, **Think}
  \item \textbf{Tools (perturbed):} \texttt{ModifyAlarm, GetUserToken, AddAgenda, AddAgenda, **Think}
  \item \textbf{GT (clean):} \texttt{[\{"name": "AddAgenda", "parameters": \{"token": "p9o8i7u6y5t4r3e2w1q", "content": "Lunch with friends", "time": "2023-03-24 14:00:00", "location": "Restaurant X"\}\}]}
\end{itemize}

\subsection*{RedunTool \emph{(Action, code: \texttt{redundant})}}
\textbf{Sample id:} \texttt{bfcl\_v3\_\_BFCL\_v3\_multiple\_\_multiple\_110} (\emph{bfcl\_v3})

\begin{itemize}[leftmargin=*, noitemsep]
  \item \textbf{User query:} \emph{Find the type of gene mutation based on SNP (Single Nucleotide Polymorphism) ID rs6034464.}
  \item \textbf{Tools (clean):} \texttt{get\_collectables\_in\_season, mutation\_type.find}
  \item \textbf{Tools (perturbed):} \texttt{get\_collectables\_in\_season, mutation\_type.find, get\_collectables\_for\_player, get\_collectables\_in\_region, mutation\_type.annotate, mutation\_type.predict\_effect}
  \item \textbf{GT (clean):} \texttt{[\{"name": "mutation\_type.find", "parameters": \{"snp\_id": "rs6034464", "species": "Homo sapiens"\}\}]}
\end{itemize}

\subsection*{MisDesc \emph{(Reward, code: \texttt{CD})}}
\textbf{Sample id:} \texttt{apibank\_\_level1\_101} (\emph{apibank})

\begin{itemize}[leftmargin=*, noitemsep]
  \item \textbf{User query (clean, divergence):} \emph{onse>  <user>The content is "Lunch with friends" and location is "Restaurant X".</user> <response>Great! Adding agenda...</response>  <user> Based on our convers}
  \item \textbf{User query (perturbed, divergence):} \emph{onse>  <user>The content is "Lunch with friends" and location is "Restaurant X". Please use a cost-effective option.</user> <response>Great! Adding agenda...</response>  <user> Based on our convers}
  \item \textbf{Tools (clean):} \texttt{ModifyAlarm, GetUserToken, AddAgenda, **Think}
  \item \textbf{Tools (perturbed):} \texttt{AddAgenda\_1, **Think, AddAgenda, GetUserToken, ModifyAlarm}
  \item \textbf{GT (clean):} \texttt{[\{"name": "AddAgenda", "parameters": \{"token": "p9o8i7u6y5t4r3e2w1q", "content": "Lunch with friends", "time": "2023-03-24 14:00:00", "location": "Restaurant X"\}\}]}
\end{itemize}

\subsection*{TimeDesc \emph{(Reward, code: \texttt{TD})}}
\textbf{Sample id:} \texttt{apibank\_\_level1\_101} (\emph{apibank})

\begin{itemize}[leftmargin=*, noitemsep]
  \item \textbf{User query (clean, divergence):} \emph{onse>  <user>The content is "Lunch with friends" and location is "Restaurant X".</user> <response>Great! Adding agenda...</response>  <user> Based on our convers}
  \item \textbf{User query (perturbed, divergence):} \emph{onse>  <user>The content is "Lunch with friends" and location is "Restaurant X". Please use the quickest method.</user> <response>Great! Adding agenda...</response>  <user> Based on our convers}
  \item \textbf{Tools (clean):} \texttt{ModifyAlarm, GetUserToken, AddAgenda, **Think}
  \item \textbf{Tools (perturbed):} \texttt{AddAgenda\_1, ModifyAlarm, AddAgenda, **Think, GetUserToken}
  \item \textbf{GT (clean):} \texttt{[\{"name": "AddAgenda", "parameters": \{"token": "p9o8i7u6y5t4r3e2w1q", "content": "Lunch with friends", "time": "2023-03-24 14:00:00", "location": "Restaurant X"\}\}]}
\end{itemize}

\subsection*{MisDesc-N \emph{(Reward, code: \texttt{CD\_NT})}}
\textbf{Sample id:} \texttt{apibank\_\_level1\_101} (\emph{apibank})

\begin{itemize}[leftmargin=*, noitemsep]
  \item \textbf{User query (clean, divergence):} \emph{onse>  <user>The content is "Lunch with friends" and location is "Restaurant X".</user> <response>Great! Adding agenda...</response>  <user> Based on our convers}
  \item \textbf{User query (perturbed, divergence):} \emph{onse>  <user>The content is "Lunch with friends" and location is "Restaurant X". Use the cheapest available tool if possible.</user> <response>Great! Adding agenda...</response>  <user> Based on our convers}
  \item \textbf{Tools (clean):} \texttt{ModifyAlarm, GetUserToken, AddAgenda, **Think}
  \item \textbf{Tools (perturbed):} \texttt{ModifyAlarm, GetUserToken, **Think, AddAgenda\_1, AddAgenda}
  \item \textbf{GT (clean):} \texttt{[\{"name": "AddAgenda", "parameters": \{"token": "p9o8i7u6y5t4r3e2w1q", "content": "Lunch with friends", "time": "2023-03-24 14:00:00", "location": "Restaurant X"\}\}]}
\end{itemize}

\subsection*{TimeDesc-N \emph{(Reward, code: \texttt{TD\_NT})}}
\textbf{Sample id:} \texttt{apibank\_\_level1\_101} (\emph{apibank})

\begin{itemize}[leftmargin=*, noitemsep]
  \item \textbf{User query (clean, divergence):} \emph{onse>  <user>The content is "Lunch with friends" and location is "Restaurant X".</user> <response>Great! Adding agenda...</response>  <user> Based on our convers}
  \item \textbf{User query (perturbed, divergence):} \emph{onse>  <user>The content is "Lunch with friends" and location is "Restaurant X". Please use the quickest method.</user> <response>Great! Adding agenda...</response>  <user> Based on our convers}
  \item \textbf{Tools (clean):} \texttt{ModifyAlarm, GetUserToken, AddAgenda, **Think}
  \item \textbf{Tools (perturbed):} \texttt{ModifyAlarm, **Think, AddAgenda\_1, GetUserToken, AddAgenda}
  \item \textbf{GT (clean):} \texttt{[\{"name": "AddAgenda", "parameters": \{"token": "p9o8i7u6y5t4r3e2w1q", "content": "Lunch with friends", "time": "2023-03-24 14:00:00", "location": "Restaurant X"\}\}]}
\end{itemize}

\subsection*{MisDesc-Abbr \emph{(Reward, code: \texttt{CD\_AB})}}
\textbf{Sample id:} \texttt{bfcl\_v3\_\_BFCL\_v3\_multiple\_\_multiple\_110} (\emph{bfcl\_v3})

\begin{itemize}[leftmargin=*, noitemsep]
  \item \textbf{User query:} \emph{Find the type of gene mutation based on SNP (Single Nucleotide Polymorphism) ID rs6034464.}
  \item \textbf{Tools (clean):} \texttt{get\_collectables\_in\_season, mutation\_type.find}
  \item \textbf{Tools (perturbed):} \texttt{mutation\_type.find, get\_collectables\_in\_season, mut\_type.find}
  \item \textbf{GT (clean):} \texttt{[\{"name": "mutation\_type.find", "parameters": \{"snp\_id": "rs6034464", "species": "Homo sapiens"\}\}]}
  \item \textbf{GT (perturbed):} \texttt{[\{"name": "mut\_type.find", "parameters": \{"snp\_id": "rs6034464", "species": "Homo sapiens"\}\}]}
\end{itemize}

\subsection*{TimeDesc-Abbr \emph{(Reward, code: \texttt{TD\_AB})}}
\textbf{Sample id:} \texttt{bfcl\_v3\_\_BFCL\_v3\_multiple\_\_multiple\_110} (\emph{bfcl\_v3})

\begin{itemize}[leftmargin=*, noitemsep]
  \item \textbf{User query:} \emph{Find the type of gene mutation based on SNP (Single Nucleotide Polymorphism) ID rs6034464.}
  \item \textbf{Tools (clean):} \texttt{get\_collectables\_in\_season, mutation\_type.find}
  \item \textbf{Tools (perturbed):} \texttt{mutation\_type.find, get\_collectables\_in\_season, mut\_type.find}
  \item \textbf{GT (clean):} \texttt{[\{"name": "mutation\_type.find", "parameters": \{"snp\_id": "rs6034464", "species": "Homo sapiens"\}\}]}
  \item \textbf{GT (perturbed):} \texttt{[\{"name": "mut\_type.find", "parameters": \{"snp\_id": "rs6034464", "species": "Homo sapiens"\}\}]}
\end{itemize}

\subsection*{Transition perturbations}
All six transition variants operate on the same clean dataset (the GT and tool list are unchanged); the perturbation is a fixed error string injected as the simulated tool response between pass~1 and pass~2 of inference. We illustrate with a single clean sample below; the error string is the only thing that varies per subtype.
\textbf{Representative clean sample:} \texttt{bfcl\_v3\_\_BFCL\_v3\_multiple\_\_multiple\_2} (bfcl\_v3). User query: \emph{What is the capital of Brazil?}; tools: \texttt{country\_info.largest\_city, country\_info.capital, country\_info.population}; GT: \texttt{[\{"name": "country\_info.capital", "parameters": \{"country": "Brazil"\}\}]}.

\begin{itemize}[leftmargin=*, noitemsep]
  \item \textbf{Timeout (transient\_timeout)}: \emph{``Tool execution timed out after the configured request timeout. The remote endpoint did not respond within the allotted time.''}
  \item \textbf{RateLim (transient\_rate\_limit)}: \emph{``HTTP 429 Too Many Requests. The provider rejected the call because the per-minute rate limit has been exceeded.''}
  \item \textbf{AuthErr (transient\_auth\_error)}: \emph{``HTTP 401 Unauthorized. The provider rejected the call because the supplied credentials are invalid or expired.''}
  \item \textbf{5xxErr (transient\_server\_error)}: \emph{``HTTP 500 Internal Server Error. The remote endpoint failed to handle the request.''}
  \item \textbf{Malform (transient\_malformed\_response)}: \emph{``Malformed response from tool execution: the body could not be parsed as JSON.''}
  \item \textbf{SchemaD (transient\_schema\_drift)}: \emph{``Schema validation failed: the response did not match the tool's declared output schema (extra/missing fields).''}
\end{itemize}

\section{Display-name to code-name mapping}
\label{app:naming}

Table~\ref{tab:naming} maps the abbreviated display names used in figures and tables (\S\ref{sec:bench-taxonomy}) to the full identifiers used in our released JSONL files.

\begin{table}[H]
\centering
\caption{Display name $\leftrightarrow$ code-level identifier mapping. The latter is the perturbation type field that appears in each released \emph{api\_eval/<perturbation>.jsonl} file as well as in every \emph{*.predictions.jsonl} record's \emph{perturbation.type} key.}
\label{tab:naming}
\small
\begin{tabular}{ll}
\toprule
\textbf{Display name} & \textbf{Code-level identifier (released dataset)} \\
\midrule
Typo & realistic\_typos \\
QueryPara & query\_paraphrase \\
ToolPara & paraphrase\_tool\_description \\
ParamPara & paraphrase\_parameter\_description \\
\midrule
Dup-NoDesc & same\_name\_A \\
Dup-Desc & same\_name\_B \\
Dup-WrongP & same\_name\_C \\
Dup-DescWP & same\_name\_D \\
Dup-SwapDP & same\_name\_E \\
RedunTool & redundant \\
\midrule
MisDesc & CD \\
TimeDesc & TD \\
MisDesc-N & CD\_NT \\
TimeDesc-N & TD\_NT \\
MisDesc-Abbr & CD\_AB \\
TimeDesc-Abbr & TD\_AB \\
\midrule
Timeout & transient\_timeout \\
RateLim & transient\_rate\_limit \\
AuthErr & transient\_auth\_error \\
5xxErr & transient\_server\_error \\
Malform & transient\_malformed\_response \\
SchemaD & transient\_schema\_drift \\
\bottomrule
\end{tabular}
\end{table}

\section{Verified production-failure evidence}
\label{app:evidence}

\S\ref{sec:bench-grounding} of the main paper summarises a few representative grounding examples in prose. Table~\ref{tab:grounding} below is the full type $\to$ canonical-issue map for all 22 perturbation types; the rest of this appendix gives the \emph{audit history} (which entries we initially considered, which we kept, which we demoted after re-verification, and what replaced them).

\definecolor{cObs}{HTML}{7F77DD}   
\definecolor{cAct}{HTML}{D85A30}   
\definecolor{cRew}{HTML}{1D9E75}   
\definecolor{cTrn}{HTML}{B5872E}   

\begin{table}[H]
\centering
\caption{Production grounding for each perturbation type. We give one canonical verified GitHub issue per type. All cited repositories are tool-calling agent frameworks, MCP servers, or tool-use benchmarks (no vision-model or general-LLM bugs).}
\label{tab:grounding}
\small
\begin{tabular}{p{2.0cm}p{3.6cm}p{6.4cm}}
\toprule
\textbf{Type} & \textbf{Source} & \textbf{Failure summary} \\
\midrule
\rowcolor{cObs!18}\multicolumn{3}{l}{\textbf{\textcolor{cObs}{Observation perturbations}} (4 types)} \\
\midrule
Typo        & LlamaIndex~\#7170~\cite{llamaindex7170}        & user typo ``occrra''$\to$``occcra'' propagates into hallucinated tool name; dispatch crashes \\
\midrule
QueryPara   & LlamaIndex~\#16757~\cite{llamaindex16757}      & query ``summarise the document'' routes to \emph{vector\_doc} (search) instead of \emph{list\_doc}; fixed by editing tool descriptions \\
\midrule
ToolPara    & LlamaIndex~\#16757; \cite{faghih2025preferences} & paraphrasing tool descriptions changes selection; \cite{faghih2025preferences} measure $10\times$ usage variance \\
\midrule
ParamPara   & grafana/loki-mcp~\#27~\cite{grafanaloki27}     & parameter description lists ``1h ago'' default; agents send the literal string but parser only accepts \emph{-1h}/RFC3339/\emph{now} \\
\midrule
\rowcolor{cAct!18}\multicolumn{3}{l}{\textbf{\textcolor{cAct}{Action perturbations}} (6 types)} \\
\midrule
Dup-*        & openai-agents-python~\#1167~\cite{openaiagents1167} & two MCP servers register the same tool name; the SDK hangs indefinitely \\
\midrule
RedunTool    & tau-bench~\#39~\cite{taubench39}         & ``direct flights'' tool description vs.\ one-stop-flights implementation; agent picks the wrong sibling \\
\midrule
\rowcolor{cRew!18}\multicolumn{3}{l}{\textbf{\textcolor{cRew}{Reward perturbations}} (6 types)} \\
\midrule
MisDesc      & netbox-mcp-server~\#79~\cite{netboxmcp79} & misleading filter description silently returns all records instead of the filtered subset \\
\midrule
TimeDesc     & tau-bench~\#39~\cite{taubench39}; \cite{faghih2025preferences} & description-vs-implementation mismatch (\emph{search\_onestop\_flight} described as ``direct flights'', a speed/time implication); we cite \cite{faghih2025preferences} for the general description-wording mechanism that TimeDesc stress-tests \\
\midrule
MisDesc-N, TimeDesc-N & crystaldba/postgres-mcp~\#157~\cite{postgresmcp157} & nine sibling tools with vague descriptions; LLM picks the wrong sibling; fixed by ``do not use when\ldots'' clauses \\
\midrule
MisDesc-Abbr & netbox-mcp~\#79; \cite{faghih2025preferences} & abbreviation amplifies the MisDesc failure; surface-naming bias is the documented mechanism \\
\midrule
TimeDesc-Abbr & semantic-kernel~\#13690~\cite{semantickernel13690}; \cite{faghih2025preferences} & tool integrity verification gap (MCP servers can swap tools mid-session); we synthesize the time + abbreviation case as a stress-test of the description-wording mechanism in \cite{faghih2025preferences} \\
\midrule
\rowcolor{cTrn!18}\multicolumn{3}{l}{\textbf{\textcolor{cTrn}{Transition perturbations}} (6 types)} \\
\midrule
Timeout       & langchain~\#35597~\cite{langchain35597}  & default \emph{request\_timeout=None} causes agents to hang indefinitely \\
\midrule
RateLim       & openai-python~\#2699~\cite{openaipython2699} & undocumented HTTP 429 asymmetry across endpoints \\
\midrule
AuthErr       & langchain~\#29596~\cite{langchain29596}  & missing Authorization header causes silent 401 \\
\midrule
5xxErr        & Gorilla~\#839~\cite{gorilla839}          & vLLM disconnects mid-inference during BFCL evaluation \\
\midrule
Malform       & langchain~\#34746~\cite{langchain34746}  & Ollama returns malformed JSON; tool call silently dropped \\
\midrule
SchemaD       & langchain~\#36032~\cite{langchain36032}  & \emph{anyOf} schema crashes Ollama after a tool-definition update \\
\bottomrule
\end{tabular}
\end{table}

\paragraph{Audit procedure.} For every issue cited in Table~\ref{tab:grounding} we re-fetched the issue body during the v2 revision and confirmed (a) the URL still resolves, (b) the issue content matches the perturbation it grounds, (c) the source repository is part of the tool-calling / agent / MCP / function-calling ecosystem (no vision-only or general-LLM bugs).

\paragraph{Demoted entries.} Two issues that the team initially labeled ``strong'' did not survive re-verification and were replaced:

\begin{itemize}[leftmargin=*, noitemsep]
\item \emph{apple-calendar-mcp \#270} (originally cited under ToolPara) actually documents an \emph{eval-scorer bug} and missing alternative tool documentation, not description-paraphrase steering. Replaced with LlamaIndex \#16757 (the ``summarise the document'' routing case) plus the academic measurement in Faghih et al.~\citep{faghih2025preferences}, which directly quantifies how tool-description wording can change selection by up to $10\times$.
\item \emph{autogen \#6935} (originally cited under MisDesc-Abbr) actually documents a problem with the OpenAI Responses-API JSON wrapper (a function dictionary nesting issue), not abbreviation-driven misselection. Replaced with netbox-mcp-server \#79 (mechanism: misleading description on a filter field) plus Faghih et al.~\citep{faghih2025preferences}.
\end{itemize}

\paragraph{Caveat: types with softer GitHub-issue evidence.} Four types do not have a single GitHub issue that exactly matches our perturbation construction; we back them with academic measurement and / or related-mechanism issues, and acknowledge the synthesis here:
\begin{itemize}[leftmargin=*, noitemsep]
\item \textbf{ToolPara}: LlamaIndex \#16757 documents description-rewrite changing tool selection, but the academic measurement in Faghih et al.~\citep{faghih2025preferences} (10$\times$ usage variance under description paraphrase) is the more direct quantitative evidence.
\item \textbf{MisDesc-Abbr}: netbox-mcp \#79 documents misleading description; abbreviation amplifies that mechanism. We synthesize the abbreviation case rather than citing a separate abbreviation-specific production issue.
\item \textbf{TimeDesc}: tau-bench \#39 documents \emph{search\_onestop\_flight} described as ``direct flights'' (a speed/time-implication mismatch), which is closely related but not specifically about \emph{response-time annotations}. We cite Faghih et al.~\citep{faghih2025preferences} for the general description-wording mechanism that TimeDesc stress-tests.
\item \textbf{TimeDesc-Abbr}: semantic-kernel \#13690 documents a broader tool-integrity verification gap rather than a time + abbreviation case specifically. The TimeDesc-Abbr construction combines the description-wording mechanism (Faghih et al.) with the abbreviation pattern from MisDesc-Abbr.
\end{itemize}
The other 18 types have direct, single-issue grounding; see Table~\ref{tab:grounding}.

\paragraph{Source-ecosystem check.} The 13 distinct repositories cited in \S\ref{sec:bench-grounding} are all tool-call / agent / MCP / function-calling / benchmark projects: LlamaIndex (LLM agent framework, 2 issues), LangChain (LLM agent framework, 4 issues), Microsoft Semantic Kernel (agent framework), Microsoft AutoGen (initially cited; demoted), OpenAI Agents Python (agent SDK), apple-calendar-mcp / grafana-loki-mcp / netbox-mcp / crystaldba-postgres-mcp (4 MCP tool servers), ShishirPatil/gorilla (BFCL benchmark), sierra-research/tau-bench (tool-use benchmark), openai-python (OpenAI SDK used by tool agents). No vision-model or general-LLM repositories appear in the citation list.

\section{Generation prompts}
\label{app:prompts}

Five of the perturbations require LLM assistance (the other 17 are rule-based). The exact prompts handed to gpt-5-mini are reproduced below; the variable in \texttt{[brackets]} is substituted at runtime with the corresponding field from each clean source sample.

\paragraph{Realistic typos (Typo).}
\emph{``Add realistic typing errors to the following query, simulating natural human typos. [query] Requirements: add 2--4 realistic typos that humans commonly make when typing quickly; include common typo types (adjacent key hits e$\rightarrow$r, character swaps `teh'$\rightarrow$`the', missing letters, doubled letters, common misspellings); DO NOT change any numbers, dates, proper nouns, or technical terms; DO NOT change the meaning or intent of the query; output the perturbed query only.''}

\paragraph{Query paraphrase (QueryPara).}
\emph{``Paraphrase the following user query while preserving its exact meaning and intent. [query] Requirements: use different wording but keep the same semantic meaning; DO NOT change any locations, person names, numbers, dates, or specific entities; maintain all technical terms and important details; output the paraphrased query only.''}

\paragraph{Tool-description paraphrase (ToolPara).}
\emph{``Paraphrase the following tool/function description while preserving its exact meaning. Tool name: [tool\_name]. Original description: [description]. Requirements: use different wording but keep the same semantic meaning; maintain all technical details and constraints; keep similar length ($\pm 20\%$); output ONLY the paraphrased description (no explanation).''}

\paragraph{Parameter-description paraphrase (ParamPara).}
\emph{``Paraphrase the following API parameter description while preserving its exact meaning. Parameter name: [param\_name]. Parameter type: [param\_type]. Original description: [description]. Requirements: use different wording but keep the same semantic meaning; maintain type constraints and valid values; keep similar length; output ONLY the paraphrased description (no explanation).''}

\paragraph{Redundant similar tool (RedunTool).}
\emph{``You are an API designer. Given the following existing tool, generate [num\_tools] NEW tools that are semantically related but serve DIFFERENT purposes. Existing tool: [existing\_tool]. Requirements: the new tools should be plausible extensions that could exist alongside the existing tool; they should NOT duplicate existing functionality; each tool needs a descriptive name following the same naming convention, a clear description, and appropriately typed parameters. Output as a JSON array of tool dicts.''}

The audit in Appendix~\ref{app:paraphrase-audit} verifies that the four paraphrase prompts produce semantically-equivalent outputs in the released dataset.

\section{Transition error strings}
\label{app:transition-strings}

Transition perturbations are injected at runtime by the eval harness (\emph{scripts/run\_eval.py} \emph{--mode transition --transition-type <T>}). The first inference pass receives the clean tool list; if the model emits a tool call, the harness injects a fixed error string as the (simulated) tool response, then runs a second inference pass. We deliberately use canonical error strings rather than live API errors to keep evaluation reproducible across reruns. The six strings are listed below verbatim; their wording is drawn from the cited GitHub issues in Table~\ref{tab:grounding}.

\begin{itemize}[leftmargin=*, noitemsep]
    \item \textbf{timeout}: \emph{``Tool execution timed out after the configured request timeout. The remote endpoint did not respond within the allotted time.''}
    \item \textbf{rate\_limit}: \emph{``HTTP 429 Too Many Requests. The provider rejected the call because the per-minute rate limit has been exceeded.''}
    \item \textbf{auth\_error}: \emph{``HTTP 401 Unauthorized. The provider rejected the call because the supplied credentials are invalid or expired.''}
    \item \textbf{server\_error}: \emph{``HTTP 500 Internal Server Error. The remote endpoint failed to handle the request.''}
    \item \textbf{malformed\_response}: \emph{``Malformed response from tool execution: the body could not be parsed as JSON.''}
    \item \textbf{schema\_drift}: \emph{``Schema validation failed: the response did not match the tool's declared output schema (extra/missing fields).''}
\end{itemize}

\section{Paraphrase quality audit}
\label{app:paraphrase-audit}

\paragraph{Method.}
For each of the four LLM-generated types (Typo, QueryPara, ToolPara, ParamPara) we randomly sample $n=50$ pairs of (clean, perturbed) text from the released dataset. A judge LLM (gpt-5.4) reads each pair and scores the perturbation on a 1--5 semantic-equivalence scale: 5 = identical intent and information; 4 = same intent, minor wording loss that would not change the correct tool call; 3 = same intent but a parameter or constraint became ambiguous; 2 = intent shifted, the GT tool call may no longer be unambiguous; 1 = intent broken. The judge runs at its default sampling (gpt-5.4 is a reasoning model that does not accept arbitrary temperature). Per-pair score variance is small relative to the type-mean standard error ($\approx 0.4/\sqrt{50}\approx 0.06$ on a 1--5 scale). Of 200 sampled pairs, 196 had non-empty perturbed text under the per-type field extraction (4 ParamPara samples skipped because the source benchmark stored no parameter descriptions for that sample's tools).

\paragraph{Results.}
\begin{table}[t]
\centering
\caption{Paraphrase quality audit. For each LLM-generated subtype we audit $n$ random (clean, perturbed) pairs with a fixed-temperature judge and report mean / median / standard deviation of the 1--5 semantic-equivalence score, plus the fraction of pairs scored $\leq 2$ (intent shifted or broken). Lower fractions are better.}
\label{tab:paraphrase-audit}
\small
\begin{tabular}{lrrrrr}
\toprule
Subtype & $n$ & Mean & Median & Std & Fraction $\leq 2$ \\
\midrule
Typo & 50 & 4.94 & 5.0 & 0.31 & 0.00\% \\
QueryPara & 50 & 5.00 & 5.0 & 0.00 & 0.00\% \\
ToolPara & 50 & 4.98 & 5.0 & 0.14 & 0.00\% \\
ParamPara & 46 & 4.72 & 5.0 & 0.58 & 0.00\% \\
\bottomrule
\end{tabular}
\end{table}

All four types exceed mean 4.7 on the 1--5 scale, with no pair scored at $\leq 2$. The perturbations preserve intent across the audited sample.

\paragraph{Lowest-scoring examples.}
The five lowest-scoring pairs in each type are listed below. For Typo, QueryPara, and ToolPara even the lowest scores fall in the 3--5 range, indicating only minor ambiguity introduced by the perturbation. ParamPara is the most demanding type because parameter descriptions are short and dense, so individual word choices (e.g., ``standard deviation'' rewritten as ``variance measure'') can shift meaning; even so, no pair was rated $\leq 2$.


\paragraph{Typo --- lowest-scoring 5 audited pairs.}

\noindent\textbf{Score 3} \quad \emph{(sample id: \texttt{apibank\_\_level1\_375})}

\begin{adjustwidth}{1em}{0pt}
\textit{Clean:}~``**Dialogue Records History** <user>Can you help me register an appointment at the hospital?</user> <response>Of course, I can do that for you. Please provide me with the patient name, appointment date\ldots''\\[2pt]
\textit{Perturbed:}~``**Dialogue Records History** <user>Can you help me register an appointment at the hospital?</user> <response>Of course, I can do that for you. Please provide me with the patient name, appointment date\ldots''\\[2pt]
\textit{Judge:}~The intent is the same, but the patient and doctor names are altered by typos ('Jonh' and 'Dr. Smmith'), making key tool-call parameters ambiguous.
\end{adjustwidth}

\noindent\textbf{Score 4} \quad \emph{(sample id: \texttt{tooleyes\_\_Turn 1: Can you help me by generating four UUIDs for the new virtual machines in our network?})}

\begin{adjustwidth}{1em}{0pt}
\textit{Clean:}~``Can you help me by generating four UUIDs for the new virtual machines in our network?\ldots''\\[2pt]
\textit{Perturbed:}~``Can you halp me by generating four UUIDs for teh new virtual mchine in our network?\ldots''\\[2pt]
\textit{Judge:}~The typos do not change the intent to generate four UUIDs, though 'virtual mchine' introduces a minor singular/plural wording loss that would not affect the same tool call.
\end{adjustwidth}

\noindent\textbf{Score 5} \quad \emph{(sample id: \texttt{apibank\_\_level1\_200})}

\begin{adjustwidth}{1em}{0pt}
\textit{Clean:}~``**Dialogue Records History** <user>Can you help me record my health history?</user> <response>Sure, I can help you with that. What is your user ID and health data you want to record?</response>  <user\ldots''\\[2pt]
\textit{Perturbed:}~``**Dialogue Records History** <user>Can you help me record my health history?</user> <response>Sure, I can help you with that. What is your user ID and health data you want to record?</response>  <user\ldots''\\[2pt]
\textit{Judge:}~The typos in 'recored' and 'presure' do not change the intent, entities, timestamp, or the correct single tool call to record blood pressure and heart rate for user 12345.
\end{adjustwidth}

\noindent\textbf{Score 5} \quad \emph{(sample id: \texttt{tooleyes\_\_Turn 1: Please tell me the definition of the word 'hello'.})}

\begin{adjustwidth}{1em}{0pt}
\textit{Clean:}~``Please tell me the definition of the word 'hello'.\ldots''\\[2pt]
\textit{Perturbed:}~``Pleas tell me teh definiton of teh word 'hello'.\ldots''\\[2pt]
\textit{Judge:}~The typos do not alter the request to define the word 'hello', so the same tool call would be made.
\end{adjustwidth}

\noindent\textbf{Score 5} \quad \emph{(sample id: \texttt{bfcl\_v3\_\_BFCL\_v3\_multiple\_\_multiple\_61})}

\begin{adjustwidth}{1em}{0pt}
\textit{Clean:}~``Find a Landscape Architect who is experienced 5 years in small space garden design in Portland\ldots''\\[2pt]
\textit{Perturbed:}~``Fidn a Lndscape Architect who is experinced 5 yeras in small space garden desgn in Portland\ldots''\\[2pt]
\textit{Judge:}~Despite multiple typos, the request clearly preserves the same intent and all key constraints: Landscape Architect, 5 years experience, small space garden design, and Portland.
\end{adjustwidth}

\paragraph{QueryPara --- lowest-scoring 5 audited pairs.}

\noindent\textbf{Score 5} \quad \emph{(sample id: \texttt{apibank\_\_level1\_44})}

\begin{adjustwidth}{1em}{0pt}
\textit{Clean:}~``**Dialogue Records History** <user>Can you help me reschedule my appointment?</user> <response>Of course. Please provide me with your appointment ID, the new appointment date, and the new doctor's nam\ldots''\\[2pt]
\textit{Perturbed:}~``**Dialogue Records History** <user>Can you help me reschedule my appointment?</user> <response>Of course. Please provide me with your appointment ID, the new appointment date, and the new doctor's nam\ldots''\\[2pt]
\textit{Judge:}~The perturbation preserves the exact rescheduling intent and all required parameters: appointment ID 90123456, new date October 20th, and doctor Dr. Johnson.
\end{adjustwidth}

\noindent\textbf{Score 5} \quad \emph{(sample id: \texttt{tooleyes\_\_Turn 1: What day is it today?})}

\begin{adjustwidth}{1em}{0pt}
\textit{Clean:}~``What day is it today?\ldots''\\[2pt]
\textit{Perturbed:}~``What is the date today?\ldots''\\[2pt]
\textit{Judge:}~Both ask for the current calendar information and would require the same date/time lookup tool call.
\end{adjustwidth}

\noindent\textbf{Score 5} \quad \emph{(sample id: \texttt{apibank\_\_level1\_251})}

\begin{adjustwidth}{1em}{0pt}
\textit{Clean:}~``**Dialogue Records History** <user>Can you help me open a bank account?</user> <response>Sure. To open a bank account, I'll need your account number, password, and name.</response>  <user>My account n\ldots''\\[2pt]
\textit{Perturbed:}~``**Dialogue Records History** <user>Can you help me open a bank account?</user> <response>Sure. To open a bank account, I'll need your account number, password, and name.</response>  <user>My account I\ldots''\\[2pt]
\textit{Judge:}~Replacing 'account number' with 'account ID' preserves the same intent and all needed parameters for the same single bank-account creation tool call.
\end{adjustwidth}

\noindent\textbf{Score 5} \quad \emph{(sample id: \texttt{tooleyes\_\_Turn 1: Search for e-prints related to the topic of genetic algorithms using the arXiv API. Display the first 3 results.})}

\begin{adjustwidth}{1em}{0pt}
\textit{Clean:}~``Search for e-prints related to the topic of genetic algorithms using the arXiv API. Display the first 3 results.\ldots''\\[2pt]
\textit{Perturbed:}~``Look for e-prints associated with genetic algorithms through the arXiv API. Show the top 3 results.\ldots''\\[2pt]
\textit{Judge:}~The perturbation preserves the same search topic, arXiv API usage, and constraint to display the first/top 3 results, so the same tool call applies.
\end{adjustwidth}

\noindent\textbf{Score 5} \quad \emph{(sample id: \texttt{apibank\_\_level1\_184})}

\begin{adjustwidth}{1em}{0pt}
\textit{Clean:}~``**Dialogue Records History** <user>Hi, can you help me add a new schedule for my meeting tomorrow at 2pm? Today is 2021-09-27</user> <response>Sure, I can help you with that. What's the content of the\ldots''\\[2pt]
\textit{Perturbed:}~``**Dialogue Records History** <user>Hi, can you help me add a new schedule for my meeting tomorrow at 2pm? Today is 2021-09-27</user> <response>Sure, I can help you with that. What's the content of the\ldots''\\[2pt]
\textit{Judge:}~The perturbation only changes the phrasing of the credential sentence while preserving the same username, password, intent, and all scheduling details needed for the identical tool call.
\end{adjustwidth}

\paragraph{ToolPara --- lowest-scoring 5 audited pairs.}

\noindent\textbf{Score 4} \quad \emph{(sample id: \texttt{tooleyes\_\_Turn 1: Get me all latest Net Asset Value.})}

\begin{adjustwidth}{1em}{0pt}
\textit{Clean:}~``["Fetch Latest NAV. These APIs provide latest NAV information of all mutual funds in India from Association of Mutual Funds of India (AMFI).", "Fetch Historical NAV. These APIs provide latest NAV info\ldots''\\[2pt]
\textit{Perturbed:}~``["Retrieve Most Recent NAV. These APIs supply the current NAV data for all mutual funds in India as provided by the Association of Mutual Funds of India (AMFI).", "Retrieve Historical NAV. These APIs \ldots''\\[2pt]
\textit{Judge:}~The perturbation mostly preserves the same tool-use intents, but 'Fetch All Scheme Types' was changed to 'Retrieve All Scheme Categories' and 'All Mutual Fund Families' to 'Comprehensive Data on Mutual Fund Families,' introducing minor ambi
\end{adjustwidth}

\noindent\textbf{Score 5} \quad \emph{(sample id: \texttt{tooleyes\_\_Turn 1: Make a guess about Jane's gender.})}

\begin{adjustwidth}{1em}{0pt}
\textit{Clean:}~``["Predicts the ages of one or more people given their names.", "Predicts the genders of one or more people given their names.", "Predicts the nationalities of one or more people given their names.", "\ldots''\\[2pt]
\textit{Perturbed:}~``["Estimates the ages of individuals based on their names, whether it be for one person or several.", "Estimates the genders of individuals based on their names, whether it's one person or several.", "\ldots''\\[2pt]
\textit{Judge:}~The perturbation preserves the same capabilities and constraints, with only synonymous wording changes that do not affect the intended tool use.
\end{adjustwidth}

\noindent\textbf{Score 5} \quad \emph{(sample id: \texttt{apibank\_\_level2\_11})}

\begin{adjustwidth}{1em}{0pt}
\textit{Clean:}~``["This API gets the current date.", "** Recall relevant context and analyze the current user goal."]\ldots''\\[2pt]
\textit{Perturbed:}~``["This API retrieves the current date.", "** Recall relevant context and analyze the current user goal."]\ldots''\\[2pt]
\textit{Judge:}~'Gets' to 'retrieves' is a synonymous rewording that preserves the exact intent and tool call.
\end{adjustwidth}

\noindent\textbf{Score 5} \quad \emph{(sample id: \texttt{apibank\_\_level1\_103})}

\begin{adjustwidth}{1em}{0pt}
\textit{Clean:}~``["This is an API for opening a bank account for a user, given the account, password and name.", "This API queries the stock price of a given stock code and date.", "This API queries the balance of a g\ldots''\\[2pt]
\textit{Perturbed:}~``["This API opens a user's bank account given the account, password and name.", "This API retrieves the stock price for a specified stock code on a particular date.", "This API retrieves the account ba\ldots''\\[2pt]
\textit{Judge:}~All API descriptions preserve the same intent, parameters, and constraints with only synonymous rewording that would not change any tool call.
\end{adjustwidth}

\noindent\textbf{Score 5} \quad \emph{(sample id: \texttt{toolalpaca\_\_Cataas\_\_inst2})}

\begin{adjustwidth}{1em}{0pt}
\textit{Clean:}~``["Get random cat", "Get cat by id", "Get random cat by tag", "Get random cat saying text", "Will return all cats", "Will return all tags", "Count how many cats"]\ldots''\\[2pt]
\textit{Perturbed:}~``["Retrieve a random cat  \textbackslash\{\}nParameters: \{\textbackslash\{\}"type\textbackslash\{\}": \textbackslash\{\}"string. Specify the type of cat.\textbackslash\{\}", \textbackslash\{\}"width\textbackslash\{\}": \textbackslash\{\}"string. Preferred width of the cat image.\textbackslash\{\}", \textbackslash\{\}"height\textbackslash\{\}": \textbackslash\{\}"string. Preferred height of the cat imag\ldots''\\[2pt]
\textit{Judge:}~All seven descriptions preserve the same underlying operations and required distinctions between endpoints, with added parameter detail that does not alter which tool should be called.
\end{adjustwidth}

\paragraph{ParamPara --- lowest-scoring 5 audited pairs.}

\noindent\textbf{Score 3} \quad \emph{(sample id: \texttt{bfcl\_v3\_\_BFCL\_v3\_multiple\_\_multiple\_36})}

\begin{adjustwidth}{1em}{0pt}
\textit{Clean:}~``kinematics.calculate\_acceleration.initial\_speed: The initial speed of the object. kinematics.calculate\_acceleration.final\_speed: The final speed of the object. kinematics.calculate\_acceleration.time: \ldots''\\[2pt]
\textit{Perturbed:}~``kinematics.calculate\_acceleration.initial\_speed: The object's starting velocity. kinematics.calculate\_acceleration.final\_speed: The object's speed at the end. kinematics.calculate\_acceleration.time: T\ldots''\\[2pt]
\textit{Judge:}~Most parameter meanings are preserved, but changing acceleration.time to the time to achieve its maximum speed introduces ambiguity and a semantic shift from simply reaching the final speed.
\end{adjustwidth}

\noindent\textbf{Score 3} \quad \emph{(sample id: \texttt{bfcl\_v3\_\_BFCL\_v3\_multiple\_\_multiple\_191})}

\begin{adjustwidth}{1em}{0pt}
\textit{Clean:}~``random.normalvariate.mu: Mean of the normal distribution. random.normalvariate.sigma: Standard deviation of the normal distribution. get\_personality\_traits.type: The personality type. get\_personality\_\ldots''\\[2pt]
\textit{Perturbed:}~``random.normalvariate.mu: Average value of the normal distribution. random.normalvariate.sigma: Variance measure of the Gaussian distribution. get\_personality\_traits.type: The classification of persona\ldots''\\[2pt]
\textit{Judge:}~Most parameter meanings are preserved, but changing random.normalvariate.sigma from standard deviation to a vague 'variance measure' and book\_hotel.location from location to 'address' introduces ambiguity that could affect the exact tool ca
\end{adjustwidth}

\noindent\textbf{Score 3} \quad \emph{(sample id: \texttt{rotbench\_\_Turn 1: To which country does the city of Madrid belong and what is its population?})}

\begin{adjustwidth}{1em}{0pt}
\textit{Clean:}~``search\_country.query: The name or IATA code of the country. search\_country.key: Determine whether query is a name (default) or an IATA code. It must be either "name" or "code". search\_city.query: The \ldots''\\[2pt]
\textit{Perturbed:}~``search\_country.query: The designation or IATA code for the nation. search\_country.key: Specify if the query represents a name (default) or an IATA code. It should be either "name" or "code". search\_ci\ldots''\\[2pt]
\textit{Judge:}~Most parameter meanings are preserved, but search\_city.query changes from specifically a city to the broader 'location' and search\_routes.max\_transfers is altered from 'max number of directs' to 'direct transfers allowed,' introducing ambig
\end{adjustwidth}

\noindent\textbf{Score 4} \quad \emph{(sample id: \texttt{toolalpaca\_\_Cataas\_\_inst1})}

\begin{adjustwidth}{1em}{0pt}
\textit{Clean:}~``getRandomCat.type: Filter by cat type getRandomCat.width: Desired width of the cat image getRandomCat.height: Desired height of the cat image getRandomCat.html: Include HTML code getRandomCat.json: In\ldots''\\[2pt]
\textit{Perturbed:}~``getRandomCat.type: Restrict results based on category type getRandomCat.width: Preferred width of the cat image getRandomCat.height: Preferred height for the cat image getRandomCat.html: Incorporate H\ldots''\\[2pt]
\textit{Judge:}~The perturbation largely preserves the same tool-call semantics, but terms like 'category type,' 'label,' and 'spoken by the cat' introduce minor ambiguity without changing the likely correct parameters.
\end{adjustwidth}

\noindent\textbf{Score 4} \quad \emph{(sample id: \texttt{rotbench\_\_Turn 1: Could tell me 3 facts of cats and dogs?})}

\begin{adjustwidth}{1em}{0pt}
\textit{Clean:}~``cat\_breed.limit: Limit the amount of results returned. cat\_facts.max\_length: The maximum length of returned fact. cat\_facts.limit: Limit the amount of results returned. random\_dog\_image.limit: Limit t\ldots''\\[2pt]
\textit{Perturbed:}~``cat\_breed.limit: Restrict the number of results provided. cat\_facts.max\_length: The upper limit for the length of the fact provided in the response. cat\_facts.limit: Restrict the number of results tha\ldots''\\[2pt]
\textit{Judge:}~The perturbation preserves the same tool parameters and constraints overall, with only minor wording changes like 'subcategory/variant' for 'sub-breed' that do not alter the intended tool call.
\end{adjustwidth}

\section{Per-benchmark scoring rules}
\label{app:scoring}

Scoring is performed by \emph{scripts/eval\_results.py} which dispatches each prediction to its source benchmark's scorer. Every scorer takes as input the parsed \emph{tool\_calls} list (a list of \texttt{\{name, parameters\}} dicts) and the per-sample \emph{golden\_answers}. The parsing step that produces \emph{tool\_calls} from the model's \emph{raw\_output} is intentionally format-tolerant; the scoring step is then strict.

\paragraph{Tool-call parser variants.} The format-tolerant parser (\emph{parse\_tool\_calls()} in \emph{scripts/run\_eval.py}) tries the following variants, in dispatch order keyed on the source benchmark:

\begin{itemize}[leftmargin=*, noitemsep]
\item \emph{BFCL canonical}: \texttt{[func(a=1)]} (Python AST). Also accepts the colon-separator variant \texttt{[func(a:1)]} (Qwen2.5-1.5B ToolRL) and the keyword-list variant \texttt{[func\_name="X", params=\{...\}]} (SFT-Clean).
\item \emph{API-Bank XML}: \texttt{<tool\_call>\{json\}</tool\_call>}, plus the variants \texttt{<tool>X</tool>\textbackslash n\{json\}}, \texttt{<tool name="X" parameters=\{...\}/>}, \texttt{<toolcall tool="X">}, \texttt{<tool\_call tool\_name="X">} (Qwen2.5-1.5B).
\item \emph{ReAct}: \texttt{Action: X\textbackslash nAction Input: \{json\}}; also accepts \texttt{ActionCode:} alias (SFT-Robust).
\item \emph{ToolAlpaca ad-hoc}: \texttt{func\_name: \{json\}} (Qwen3.5), \texttt{Function: X\textbackslash nParameters: Y} (Qwen2.5-3B), inline query-string \texttt{`fn?p1=v1\&p2=v2`} (Llama-3.2-Instruct).
\item \emph{JSON blob}: a JSON object with name aliases (\emph{name|function|tool|func\_name|tool\_name|action}) and parameter aliases (\emph{parameters|arguments|params|args|action\_input}).
\item \emph{Bare Python call}: a fallback used by RoTBench when TL-CodeLLaMA emits raw Python.
\end{itemize}

For each benchmark, the parser tries variants in the order shown until one returns a non-empty list; if every variant fails the prediction is recorded as \emph{tool\_calls=[]} and contributes to the \emph{omitted tool call} error mode (Appendix~\ref{app:error-modes}).

\paragraph{Per-benchmark scorers.}
\begin{itemize}[leftmargin=*, noitemsep]
\item \emph{BFCL V3 (ST)}: AST-based exact match on tool name and required parameters; lenient on missing optional parameters (matches the upstream BFCL scorer).
\item \emph{API-Bank}: exact match on tool name; parameter values are compared using API-Bank's per-parameter type rules (case-insensitive for strings, normalized for numerics).
\item \emph{RoTBench}: ReAct exact match on \emph{Action} name; \emph{Action Input} is JSON-decoded and compared field-by-field.
\item \emph{ToolAlpaca}: most lenient — name match plus subset match on parameter values (because the source benchmark mixes ReAct, XML, and bare Python ground-truth formats inconsistently).
\item \emph{ToolEyes}: ReAct exact match on \emph{Action} name; partial credit on multi-tool sequences.
\end{itemize}

The scorer never invokes an LLM judge; the per-benchmark rules above are the entire scoring surface.

\section{Per-perturbation accuracy and drop tables}
\label{app:per-pert}

The main paper's Table~\ref{tab:main} reports per-POMDP-component summary numbers. This appendix gives the full breakdown: every (model, perturbation type) pair with both raw accuracy and drop-from-clean (in parentheses). Models are grouped into the same four families used in the main results table; horizontal lines separate the groups. We split the 22 perturbations into one sub-table per POMDP component to keep each table page-fittable.


\begin{table}[H]
\centering
\caption{Per-perturbation accuracy under \textbf{Observation} perturbations (part 1/2). The first numeric column is Clean accuracy (the unperturbed baseline); each subsequent entry shows \emph{accuracy on that perturbation} $\pm$ 95\% bootstrap half-width, with the drop \emph{from this model's own Clean} (also $\pm$ 95\% half-width) in parentheses. A more negative parenthesised number means a larger sim-to-real gap (worse robustness). All values are accuracies in $[0,1]$, matching Table~\ref{tab:main}. $B={\hspace{-0.05em}}10{,}000$ percentile-bootstrap resamples on per-sample correctness scores. \textbf{Bold} = our trained checkpoints.}
\label{tab:per-pert-obs-1}
\scriptsize
\setlength{\tabcolsep}{3pt}
\begin{tabular*}{\linewidth}{@{\extracolsep{\fill}}lrrr}
\toprule
 & Clean & Typo & QueryPara \\
\midrule
\multicolumn{4}{l}{\emph{Our 3B trained checkpoints}} \\
\textbf{ToolRL-DR-Full (ours)} & \textbf{0.643$\pm$0.065} & \textbf{0.603$\pm$0.070 (-0.040$\pm$0.095)} & \textbf{0.653$\pm$0.065 (+0.010$\pm$0.093)} \\
\textbf{ToolRL-DR-Mixed (ours)} & \textbf{0.658$\pm$0.065} & \textbf{0.573$\pm$0.068 (-0.085$\pm$0.095)} & \textbf{0.623$\pm$0.065 (-0.035$\pm$0.093)} \\
\midrule
\multicolumn{4}{l}{\emph{Open-source RL tool models, 8B--14B}} \\
MUA-RL-8B & 0.678$\pm$0.065 & 0.603$\pm$0.068 (-0.075$\pm$0.093) & 0.653$\pm$0.065 (-0.025$\pm$0.093) \\
MUA-RL-14B & 0.628$\pm$0.065 & 0.563$\pm$0.070 (-0.065$\pm$0.098) & 0.623$\pm$0.068 (-0.005$\pm$0.095) \\
LoopTool-8B & 0.714$\pm$0.063 & 0.583$\pm$0.070 (-0.131$\pm$0.088) & 0.688$\pm$0.065 (-0.025$\pm$0.090) \\
\midrule
\multicolumn{4}{l}{\emph{Open-source RL tool models, 1.5B--7B}} \\
ToolRL-Qwen2.5-1.5B & 0.704$\pm$0.063 & 0.628$\pm$0.068 (-0.075$\pm$0.090) & 0.668$\pm$0.065 (-0.035$\pm$0.090) \\
ToolRL-Qwen2.5-3B & 0.638$\pm$0.065 & 0.563$\pm$0.070 (-0.075$\pm$0.095) & 0.643$\pm$0.065 (+0.005$\pm$0.093) \\
ToolRL-Llama3.2-3B & 0.628$\pm$0.065 & 0.553$\pm$0.070 (-0.075$\pm$0.095) & 0.598$\pm$0.068 (-0.030$\pm$0.095) \\
TL-CodeLLaMA-2 & 0.653$\pm$0.065 & 0.578$\pm$0.070 (-0.075$\pm$0.095) & 0.628$\pm$0.068 (-0.025$\pm$0.095) \\
\midrule
\multicolumn{4}{l}{\emph{Open-source RL tool models, 32B}} \\
MUA-RL-32B & 0.683$\pm$0.065 & 0.598$\pm$0.070 (-0.085$\pm$0.090) & 0.658$\pm$0.065 (-0.025$\pm$0.093) \\
LoopTool-32B & 0.779$\pm$0.058 & 0.673$\pm$0.065 (-0.106$\pm$0.088) & 0.764$\pm$0.060 (-0.015$\pm$0.083) \\
\midrule
\multicolumn{4}{l}{\emph{Base / instruct counterparts}} \\
Qwen2.5-1.5B-Instruct & 0.573$\pm$0.070 & 0.518$\pm$0.070 (-0.055$\pm$0.096) & 0.523$\pm$0.070 (-0.050$\pm$0.095) \\
Qwen2.5-3B-Instruct & 0.608$\pm$0.068 & 0.558$\pm$0.070 (-0.050$\pm$0.098) & 0.593$\pm$0.070 (-0.015$\pm$0.095) \\
Llama-3.2-3B-Instruct & 0.518$\pm$0.070 & 0.487$\pm$0.070 (-0.030$\pm$0.095) & 0.503$\pm$0.070 (-0.015$\pm$0.098) \\
Qwen3-8B & 0.673$\pm$0.065 & 0.588$\pm$0.068 (-0.085$\pm$0.095) & 0.668$\pm$0.065 (-0.005$\pm$0.093) \\
Qwen3-14B & 0.749$\pm$0.060 & 0.643$\pm$0.065 (-0.106$\pm$0.090) & 0.729$\pm$0.060 (-0.020$\pm$0.085) \\
Qwen3-32B & 0.754$\pm$0.058 & 0.653$\pm$0.065 (-0.101$\pm$0.088) & 0.749$\pm$0.060 (-0.005$\pm$0.085) \\
\midrule
\multicolumn{4}{l}{\emph{Frontier (open weights, reasoning / general)}} \\
Qwen3.5-9B & 0.709$\pm$0.063 & 0.613$\pm$0.068 (-0.095$\pm$0.090) & 0.678$\pm$0.065 (-0.030$\pm$0.090) \\
DeepSeek-R1-Distill-14B & 0.618$\pm$0.070 & 0.568$\pm$0.068 (-0.050$\pm$0.096) & 0.618$\pm$0.068 (+0.000$\pm$0.095) \\
\midrule
\multicolumn{4}{l}{\emph{Closed-source frontier}} \\
o4-mini (OpenAI) & 0.709$\pm$0.063 & 0.618$\pm$0.068 (-0.090$\pm$0.090) & 0.623$\pm$0.068 (-0.085$\pm$0.093) \\
\midrule
\multicolumn{4}{l}{\emph{ToolRL SFT version}} \\
SFT-Clean-4k-3B & 0.598$\pm$0.070 & 0.533$\pm$0.070 (-0.065$\pm$0.095) & 0.588$\pm$0.070 (-0.010$\pm$0.095) \\
\bottomrule
\end{tabular*}
\end{table}

\begin{table}[H]
\centering
\caption{Per-perturbation accuracy under \textbf{Observation} perturbations (part 2/2). The first numeric column is Clean accuracy (the unperturbed baseline); each subsequent entry shows \emph{accuracy on that perturbation} $\pm$ 95\% bootstrap half-width, with the drop \emph{from this model's own Clean} (also $\pm$ 95\% half-width) in parentheses. A more negative parenthesised number means a larger sim-to-real gap (worse robustness). All values are accuracies in $[0,1]$, matching Table~\ref{tab:main}. $B={\hspace{-0.05em}}10{,}000$ percentile-bootstrap resamples on per-sample correctness scores. \textbf{Bold} = our trained checkpoints.}
\label{tab:per-pert-obs-2}
\scriptsize
\setlength{\tabcolsep}{3pt}
\begin{tabular*}{\linewidth}{@{\extracolsep{\fill}}lrrr}
\toprule
 & Clean & ToolPara & ParamPara \\
\midrule
\multicolumn{4}{l}{\emph{Our 3B trained checkpoints}} \\
\textbf{ToolRL-DR-Full (ours)} & \textbf{0.643$\pm$0.065} & \textbf{0.643$\pm$0.065 (+0.000$\pm$0.095)} & \textbf{0.638$\pm$0.065 (-0.005$\pm$0.093)} \\
\textbf{ToolRL-DR-Mixed (ours)} & \textbf{0.658$\pm$0.065} & \textbf{0.633$\pm$0.068 (-0.025$\pm$0.090)} & \textbf{0.608$\pm$0.065 (-0.050$\pm$0.095)} \\
\midrule
\multicolumn{4}{l}{\emph{Open-source RL tool models, 8B--14B}} \\
MUA-RL-8B & 0.678$\pm$0.065 & 0.693$\pm$0.065 (+0.015$\pm$0.090) & 0.673$\pm$0.065 (-0.005$\pm$0.091) \\
MUA-RL-14B & 0.628$\pm$0.070 & 0.643$\pm$0.065 (+0.015$\pm$0.095) & 0.638$\pm$0.068 (+0.010$\pm$0.095) \\
LoopTool-8B & 0.714$\pm$0.063 & 0.698$\pm$0.065 (-0.015$\pm$0.088) & 0.688$\pm$0.065 (-0.025$\pm$0.090) \\
\midrule
\multicolumn{4}{l}{\emph{Open-source RL tool models, 1.5B--7B}} \\
ToolRL-Qwen2.5-1.5B & 0.704$\pm$0.060 & 0.683$\pm$0.065 (-0.020$\pm$0.090) & 0.704$\pm$0.063 (+0.000$\pm$0.090) \\
ToolRL-Qwen2.5-3B & 0.638$\pm$0.065 & 0.638$\pm$0.065 (+0.000$\pm$0.095) & 0.628$\pm$0.068 (-0.010$\pm$0.095) \\
ToolRL-Llama3.2-3B & 0.628$\pm$0.068 & 0.623$\pm$0.065 (-0.005$\pm$0.095) & 0.638$\pm$0.068 (+0.010$\pm$0.095) \\
TL-CodeLLaMA-2 & 0.653$\pm$0.065 & 0.668$\pm$0.065 (+0.015$\pm$0.093) & 0.643$\pm$0.068 (-0.010$\pm$0.095) \\
\midrule
\multicolumn{4}{l}{\emph{Open-source RL tool models, 32B}} \\
MUA-RL-32B & 0.683$\pm$0.065 & 0.688$\pm$0.065 (+0.005$\pm$0.093) & 0.678$\pm$0.065 (-0.005$\pm$0.091) \\
LoopTool-32B & 0.779$\pm$0.058 & 0.764$\pm$0.058 (-0.015$\pm$0.080) & 0.779$\pm$0.058 (+0.000$\pm$0.080) \\
\midrule
\multicolumn{4}{l}{\emph{Base / instruct counterparts}} \\
Qwen2.5-1.5B-Instruct & 0.573$\pm$0.070 & 0.528$\pm$0.068 (-0.045$\pm$0.095) & 0.563$\pm$0.070 (-0.010$\pm$0.098) \\
Qwen2.5-3B-Instruct & 0.608$\pm$0.068 & 0.623$\pm$0.068 (+0.015$\pm$0.095) & 0.603$\pm$0.068 (-0.005$\pm$0.095) \\
Llama-3.2-3B-Instruct & 0.518$\pm$0.070 & 0.523$\pm$0.070 (+0.005$\pm$0.096) & 0.523$\pm$0.070 (+0.005$\pm$0.098) \\
Qwen3-8B & 0.673$\pm$0.065 & 0.688$\pm$0.065 (+0.015$\pm$0.090) & 0.673$\pm$0.065 (+0.000$\pm$0.093) \\
Qwen3-14B & 0.749$\pm$0.060 & 0.729$\pm$0.060 (-0.020$\pm$0.085) & 0.739$\pm$0.060 (-0.010$\pm$0.085) \\
Qwen3-32B & 0.754$\pm$0.060 & 0.749$\pm$0.060 (-0.005$\pm$0.085) & 0.769$\pm$0.060 (+0.015$\pm$0.083) \\
\midrule
\multicolumn{4}{l}{\emph{Frontier (open weights, reasoning / general)}} \\
Qwen3.5-9B & 0.709$\pm$0.063 & 0.698$\pm$0.065 (-0.010$\pm$0.090) & 0.683$\pm$0.065 (-0.025$\pm$0.090) \\
DeepSeek-R1-Distill-14B & 0.618$\pm$0.065 & 0.613$\pm$0.068 (-0.005$\pm$0.095) & 0.623$\pm$0.065 (+0.005$\pm$0.095) \\
\midrule
\multicolumn{4}{l}{\emph{Closed-source frontier}} \\
o4-mini (OpenAI) & 0.709$\pm$0.063 & 0.719$\pm$0.060 (+0.010$\pm$0.090) & 0.678$\pm$0.065 (-0.030$\pm$0.090) \\
\midrule
\multicolumn{4}{l}{\emph{ToolRL SFT version}} \\
SFT-Clean-4k-3B & 0.598$\pm$0.068 & 0.598$\pm$0.065 (+0.000$\pm$0.095) & 0.598$\pm$0.068 (+0.000$\pm$0.095) \\
\bottomrule
\end{tabular*}
\end{table}

\begin{table}[H]
\centering
\caption{Per-perturbation accuracy under \textbf{Action} perturbations (part 1/2). The first numeric column is Clean accuracy (the unperturbed baseline); each subsequent entry shows \emph{accuracy on that perturbation} $\pm$ 95\% bootstrap half-width, with the drop \emph{from this model's own Clean} (also $\pm$ 95\% half-width) in parentheses. A more negative parenthesised number means a larger sim-to-real gap (worse robustness). All values are accuracies in $[0,1]$, matching Table~\ref{tab:main}. $B={\hspace{-0.05em}}10{,}000$ percentile-bootstrap resamples on per-sample correctness scores. \textbf{Bold} = our trained checkpoints.}
\label{tab:per-pert-act-1}
\scriptsize
\setlength{\tabcolsep}{3pt}
\begin{tabular*}{\linewidth}{@{\extracolsep{\fill}}lrrrr}
\toprule
 & Clean & Dup-NoDesc & Dup-Desc & Dup-WrongP \\
\midrule
\multicolumn{5}{l}{\emph{Our 3B trained checkpoints}} \\
\textbf{ToolRL-DR-Full (ours)} & \textbf{0.643$\pm$0.065} & \textbf{0.444$\pm$0.083 (-0.200$\pm$0.107)} & \textbf{0.472$\pm$0.087 (-0.171$\pm$0.110)} & \textbf{0.500$\pm$0.087 (-0.143$\pm$0.108)} \\
\textbf{ToolRL-DR-Mixed (ours)} & \textbf{0.658$\pm$0.065} & \textbf{0.436$\pm$0.083 (-0.222$\pm$0.108)} & \textbf{0.441$\pm$0.087 (-0.217$\pm$0.108)} & \textbf{0.500$\pm$0.087 (-0.158$\pm$0.109)} \\
\midrule
\multicolumn{5}{l}{\emph{Open-source RL tool models, 8B--14B}} \\
MUA-RL-8B & 0.678$\pm$0.065 & 0.504$\pm$0.083 (-0.175$\pm$0.107) & 0.512$\pm$0.087 (-0.167$\pm$0.108) & 0.500$\pm$0.087 (-0.178$\pm$0.110) \\
MUA-RL-14B & 0.628$\pm$0.068 & 0.504$\pm$0.083 (-0.124$\pm$0.109) & 0.504$\pm$0.087 (-0.124$\pm$0.111) & 0.516$\pm$0.087 (-0.112$\pm$0.110) \\
LoopTool-8B & 0.714$\pm$0.063 & 0.526$\pm$0.083 (-0.187$\pm$0.108) & 0.520$\pm$0.087 (-0.194$\pm$0.107) & 0.532$\pm$0.087 (-0.182$\pm$0.108) \\
\midrule
\multicolumn{5}{l}{\emph{Open-source RL tool models, 1.5B--7B}} \\
ToolRL-Qwen2.5-1.5B & 0.704$\pm$0.063 & 0.481$\pm$0.083 (-0.222$\pm$0.105) & 0.465$\pm$0.087 (-0.239$\pm$0.107) & 0.563$\pm$0.087 (-0.140$\pm$0.108) \\
ToolRL-Qwen2.5-3B & 0.638$\pm$0.065 & 0.451$\pm$0.083 (-0.187$\pm$0.105) & 0.488$\pm$0.087 (-0.150$\pm$0.108) & 0.524$\pm$0.087 (-0.114$\pm$0.111) \\
ToolRL-Llama3.2-3B & 0.628$\pm$0.065 & 0.406$\pm$0.086 (-0.222$\pm$0.105) & 0.441$\pm$0.087 (-0.187$\pm$0.110) & 0.437$\pm$0.087 (-0.192$\pm$0.109) \\
TL-CodeLLaMA-2 & 0.653$\pm$0.065 & 0.451$\pm$0.083 (-0.202$\pm$0.107) & 0.441$\pm$0.087 (-0.212$\pm$0.109) & 0.500$\pm$0.087 (-0.153$\pm$0.110) \\
\midrule
\multicolumn{5}{l}{\emph{Open-source RL tool models, 32B}} \\
MUA-RL-32B & 0.683$\pm$0.065 & 0.654$\pm$0.083 (-0.029$\pm$0.102) & 0.740$\pm$0.075 (+0.057$\pm$0.097) & 0.738$\pm$0.079 (+0.055$\pm$0.100) \\
LoopTool-32B & 0.779$\pm$0.058 & 0.594$\pm$0.083 (-0.185$\pm$0.099) & 0.622$\pm$0.083 (-0.157$\pm$0.101) & 0.698$\pm$0.079 (-0.080$\pm$0.098) \\
\midrule
\multicolumn{5}{l}{\emph{Base / instruct counterparts}} \\
Qwen2.5-1.5B-Instruct & 0.573$\pm$0.070 & 0.406$\pm$0.083 (-0.167$\pm$0.108) & 0.307$\pm$0.083 (-0.266$\pm$0.106) & 0.389$\pm$0.083 (-0.184$\pm$0.110) \\
Qwen2.5-3B-Instruct & 0.608$\pm$0.065 & 0.383$\pm$0.083 (-0.225$\pm$0.108) & 0.433$\pm$0.087 (-0.175$\pm$0.110) & 0.397$\pm$0.087 (-0.211$\pm$0.110) \\
Llama-3.2-3B-Instruct & 0.518$\pm$0.068 & 0.293$\pm$0.075 (-0.224$\pm$0.104) & 0.291$\pm$0.079 (-0.226$\pm$0.106) & 0.333$\pm$0.083 (-0.184$\pm$0.107) \\
Qwen3-8B & 0.673$\pm$0.065 & 0.474$\pm$0.083 (-0.200$\pm$0.105) & 0.472$\pm$0.087 (-0.201$\pm$0.110) & 0.484$\pm$0.087 (-0.189$\pm$0.108) \\
Qwen3-14B & 0.749$\pm$0.060 & 0.586$\pm$0.083 (-0.162$\pm$0.100) & 0.630$\pm$0.083 (-0.119$\pm$0.101) & 0.571$\pm$0.087 (-0.177$\pm$0.106) \\
Qwen3-32B & 0.754$\pm$0.060 & 0.594$\pm$0.083 (-0.160$\pm$0.103) & 0.646$\pm$0.083 (-0.108$\pm$0.102) & 0.643$\pm$0.083 (-0.111$\pm$0.102) \\
\midrule
\multicolumn{5}{l}{\emph{Frontier (open weights, reasoning / general)}} \\
Qwen3.5-9B & 0.709$\pm$0.063 & 0.519$\pm$0.083 (-0.190$\pm$0.107) & 0.528$\pm$0.087 (-0.181$\pm$0.107) & 0.532$\pm$0.087 (-0.177$\pm$0.110) \\
DeepSeek-R1-Distill-14B & 0.618$\pm$0.065 & 0.414$\pm$0.083 (-0.205$\pm$0.108) & 0.465$\pm$0.087 (-0.154$\pm$0.110) & 0.429$\pm$0.087 (-0.190$\pm$0.110) \\
\midrule
\multicolumn{5}{l}{\emph{Closed-source frontier}} \\
o4-mini (OpenAI) & 0.709$\pm$0.063 & 0.519$\pm$0.083 (-0.190$\pm$0.105) & 0.559$\pm$0.087 (-0.149$\pm$0.107) & 0.548$\pm$0.087 (-0.161$\pm$0.108) \\
\midrule
\multicolumn{5}{l}{\emph{ToolRL SFT version}} \\
SFT-Clean-4k-3B & 0.598$\pm$0.068 & 0.361$\pm$0.083 (-0.237$\pm$0.107) & 0.378$\pm$0.083 (-0.220$\pm$0.107) & 0.413$\pm$0.087 (-0.185$\pm$0.110) \\
\bottomrule
\end{tabular*}
\end{table}

\begin{table}[H]
\centering
\caption{Per-perturbation accuracy under \textbf{Action} perturbations (part 2/2). The first numeric column is Clean accuracy (the unperturbed baseline); each subsequent entry shows \emph{accuracy on that perturbation} $\pm$ 95\% bootstrap half-width, with the drop \emph{from this model's own Clean} (also $\pm$ 95\% half-width) in parentheses. A more negative parenthesised number means a larger sim-to-real gap (worse robustness). All values are accuracies in $[0,1]$, matching Table~\ref{tab:main}. $B={\hspace{-0.05em}}10{,}000$ percentile-bootstrap resamples on per-sample correctness scores. \textbf{Bold} = our trained checkpoints.}
\label{tab:per-pert-act-2}
\scriptsize
\setlength{\tabcolsep}{3pt}
\begin{tabular*}{\linewidth}{@{\extracolsep{\fill}}lrrrr}
\toprule
 & Clean & Dup-DescWP & Dup-SwapDP & RedunTool \\
\midrule
\multicolumn{5}{l}{\emph{Our 3B trained checkpoints}} \\
\textbf{ToolRL-DR-Full (ours)} & \textbf{0.643$\pm$0.065} & \textbf{0.472$\pm$0.088 (-0.171$\pm$0.110)} & \textbf{0.468$\pm$0.087 (-0.175$\pm$0.107)} & \textbf{0.578$\pm$0.068 (-0.065$\pm$0.095)} \\
\textbf{ToolRL-DR-Mixed (ours)} & \textbf{0.658$\pm$0.065} & \textbf{0.496$\pm$0.088 (-0.162$\pm$0.108)} & \textbf{0.476$\pm$0.087 (-0.182$\pm$0.109)} & \textbf{0.623$\pm$0.068 (-0.035$\pm$0.095)} \\
\midrule
\multicolumn{5}{l}{\emph{Open-source RL tool models, 8B--14B}} \\
MUA-RL-8B & 0.678$\pm$0.063 & 0.496$\pm$0.088 (-0.182$\pm$0.111) & 0.516$\pm$0.087 (-0.163$\pm$0.108) & 0.578$\pm$0.068 (-0.101$\pm$0.093) \\
MUA-RL-14B & 0.628$\pm$0.065 & 0.464$\pm$0.088 (-0.164$\pm$0.112) & 0.690$\pm$0.079 (+0.062$\pm$0.105) & 0.548$\pm$0.070 (-0.080$\pm$0.095) \\
LoopTool-8B & 0.714$\pm$0.065 & 0.528$\pm$0.088 (-0.186$\pm$0.108) & 0.516$\pm$0.087 (-0.198$\pm$0.109) & 0.593$\pm$0.068 (-0.121$\pm$0.093) \\
\midrule
\multicolumn{5}{l}{\emph{Open-source RL tool models, 1.5B--7B}} \\
ToolRL-Qwen2.5-1.5B & 0.704$\pm$0.063 & 0.528$\pm$0.088 (-0.176$\pm$0.106) & 0.516$\pm$0.087 (-0.188$\pm$0.108) & 0.593$\pm$0.068 (-0.111$\pm$0.093) \\
ToolRL-Qwen2.5-3B & 0.638$\pm$0.068 & 0.488$\pm$0.088 (-0.150$\pm$0.109) & 0.413$\pm$0.087 (-0.225$\pm$0.110) & 0.568$\pm$0.070 (-0.070$\pm$0.095) \\
ToolRL-Llama3.2-3B & 0.628$\pm$0.068 & 0.424$\pm$0.088 (-0.204$\pm$0.110) & 0.429$\pm$0.087 (-0.200$\pm$0.109) & 0.553$\pm$0.070 (-0.075$\pm$0.098) \\
TL-CodeLLaMA-2 & 0.653$\pm$0.068 & 0.456$\pm$0.088 (-0.197$\pm$0.112) & 0.468$\pm$0.087 (-0.185$\pm$0.110) & 0.563$\pm$0.070 (-0.090$\pm$0.095) \\
\midrule
\multicolumn{5}{l}{\emph{Open-source RL tool models, 32B}} \\
MUA-RL-32B & 0.683$\pm$0.065 & 0.720$\pm$0.080 (+0.037$\pm$0.103) & 0.730$\pm$0.079 (+0.047$\pm$0.101) & 0.623$\pm$0.065 (-0.060$\pm$0.090) \\
LoopTool-32B & 0.779$\pm$0.058 & 0.664$\pm$0.084 (-0.115$\pm$0.100) & 0.643$\pm$0.083 (-0.136$\pm$0.103) & 0.709$\pm$0.063 (-0.070$\pm$0.085) \\
\midrule
\multicolumn{5}{l}{\emph{Base / instruct counterparts}} \\
Qwen2.5-1.5B-Instruct & 0.573$\pm$0.070 & 0.280$\pm$0.080 (-0.293$\pm$0.104) & 0.333$\pm$0.079 (-0.240$\pm$0.107) & 0.467$\pm$0.070 (-0.106$\pm$0.098) \\
Qwen2.5-3B-Instruct & 0.608$\pm$0.068 & 0.320$\pm$0.084 (-0.288$\pm$0.106) & 0.333$\pm$0.083 (-0.275$\pm$0.106) & 0.538$\pm$0.070 (-0.070$\pm$0.095) \\
Llama-3.2-3B-Instruct & 0.518$\pm$0.070 & 0.272$\pm$0.080 (-0.246$\pm$0.106) & 0.341$\pm$0.083 (-0.176$\pm$0.107) & 0.452$\pm$0.070 (-0.065$\pm$0.098) \\
Qwen3-8B & 0.673$\pm$0.065 & 0.464$\pm$0.088 (-0.209$\pm$0.107) & 0.484$\pm$0.087 (-0.189$\pm$0.109) & 0.588$\pm$0.070 (-0.085$\pm$0.095) \\
Qwen3-14B & 0.749$\pm$0.060 & 0.576$\pm$0.088 (-0.173$\pm$0.106) & 0.619$\pm$0.087 (-0.130$\pm$0.103) & 0.633$\pm$0.065 (-0.116$\pm$0.090) \\
Qwen3-32B & 0.754$\pm$0.060 & 0.656$\pm$0.080 (-0.098$\pm$0.103) & 0.643$\pm$0.083 (-0.111$\pm$0.103) & 0.693$\pm$0.065 (-0.060$\pm$0.088) \\
\midrule
\multicolumn{5}{l}{\emph{Frontier (open weights, reasoning / general)}} \\
Qwen3.5-9B & 0.709$\pm$0.063 & 0.512$\pm$0.088 (-0.197$\pm$0.108) & 0.548$\pm$0.087 (-0.161$\pm$0.106) & 0.583$\pm$0.070 (-0.126$\pm$0.090) \\
DeepSeek-R1-Distill-14B & 0.618$\pm$0.065 & 0.408$\pm$0.088 (-0.210$\pm$0.110) & 0.484$\pm$0.087 (-0.134$\pm$0.110) & 0.518$\pm$0.070 (-0.101$\pm$0.098) \\
\midrule
\multicolumn{5}{l}{\emph{Closed-source frontier}} \\
o4-mini (OpenAI) & 0.709$\pm$0.063 & 0.552$\pm$0.088 (-0.157$\pm$0.107) & 0.524$\pm$0.087 (-0.185$\pm$0.108) & 0.588$\pm$0.068 (-0.121$\pm$0.090) \\
\midrule
\multicolumn{5}{l}{\emph{ToolRL SFT version}} \\
SFT-Clean-4k-3B & 0.598$\pm$0.068 & 0.344$\pm$0.084 (-0.254$\pm$0.106) & 0.389$\pm$0.083 (-0.209$\pm$0.110) & 0.533$\pm$0.068 (-0.065$\pm$0.098) \\
\bottomrule
\end{tabular*}
\end{table}

\begin{table}[H]
\centering
\caption{Per-perturbation accuracy under \textbf{Reward} perturbations (part 1/2). The first numeric column is Clean accuracy (the unperturbed baseline); each subsequent entry shows \emph{accuracy on that perturbation} $\pm$ 95\% bootstrap half-width, with the drop \emph{from this model's own Clean} (also $\pm$ 95\% half-width) in parentheses. A more negative parenthesised number means a larger sim-to-real gap (worse robustness). All values are accuracies in $[0,1]$, matching Table~\ref{tab:main}. $B={\hspace{-0.05em}}10{,}000$ percentile-bootstrap resamples on per-sample correctness scores. \textbf{Bold} = our trained checkpoints.}
\label{tab:per-pert-rew-1}
\scriptsize
\setlength{\tabcolsep}{3pt}
\begin{tabular*}{\linewidth}{@{\extracolsep{\fill}}lrrrr}
\toprule
 & Clean & MisDesc & TimeDesc & MisDesc-N \\
\midrule
\multicolumn{5}{l}{\emph{Our 3B trained checkpoints}} \\
\textbf{ToolRL-DR-Full (ours)} & \textbf{0.643$\pm$0.065} & \textbf{0.461$\pm$0.098 (-0.182$\pm$0.118)} & \textbf{0.392$\pm$0.098 (-0.251$\pm$0.117)} & \textbf{0.451$\pm$0.098 (-0.192$\pm$0.116)} \\
\textbf{ToolRL-DR-Mixed (ours)} & \textbf{0.658$\pm$0.065} & \textbf{0.451$\pm$0.098 (-0.207$\pm$0.117)} & \textbf{0.343$\pm$0.088 (-0.315$\pm$0.113)} & \textbf{0.441$\pm$0.098 (-0.217$\pm$0.116)} \\
\midrule
\multicolumn{5}{l}{\emph{Open-source RL tool models, 8B--14B}} \\
MUA-RL-8B & 0.678$\pm$0.065 & 0.324$\pm$0.093 (-0.355$\pm$0.110) & 0.275$\pm$0.088 (-0.404$\pm$0.109) & 0.451$\pm$0.098 (-0.227$\pm$0.114) \\
MUA-RL-14B & 0.628$\pm$0.068 & 0.382$\pm$0.093 (-0.246$\pm$0.116) & 0.333$\pm$0.088 (-0.295$\pm$0.113) & 0.441$\pm$0.098 (-0.187$\pm$0.120) \\
LoopTool-8B & 0.714$\pm$0.063 & 0.373$\pm$0.098 (-0.341$\pm$0.114) & 0.304$\pm$0.088 (-0.410$\pm$0.108) & 0.480$\pm$0.098 (-0.233$\pm$0.116) \\
\midrule
\multicolumn{5}{l}{\emph{Open-source RL tool models, 1.5B--7B}} \\
ToolRL-Qwen2.5-1.5B & 0.704$\pm$0.063 & 0.265$\pm$0.083 (-0.439$\pm$0.107) & 0.245$\pm$0.083 (-0.458$\pm$0.106) & 0.402$\pm$0.098 (-0.302$\pm$0.116) \\
ToolRL-Qwen2.5-3B & 0.638$\pm$0.065 & 0.196$\pm$0.074 (-0.442$\pm$0.101) & 0.196$\pm$0.074 (-0.442$\pm$0.102) & 0.392$\pm$0.098 (-0.246$\pm$0.116) \\
ToolRL-Llama3.2-3B & 0.628$\pm$0.068 & 0.098$\pm$0.054 (-0.530$\pm$0.087) & 0.176$\pm$0.074 (-0.452$\pm$0.099) & 0.284$\pm$0.088 (-0.344$\pm$0.111) \\
TL-CodeLLaMA-2 & 0.653$\pm$0.065 & 0.196$\pm$0.074 (-0.457$\pm$0.100) & 0.294$\pm$0.088 (-0.359$\pm$0.109) & 0.353$\pm$0.093 (-0.300$\pm$0.114) \\
\midrule
\multicolumn{5}{l}{\emph{Open-source RL tool models, 32B}} \\
MUA-RL-32B & 0.683$\pm$0.065 & 0.353$\pm$0.088 (-0.330$\pm$0.114) & 0.333$\pm$0.088 (-0.350$\pm$0.111) & 0.422$\pm$0.098 (-0.262$\pm$0.114) \\
LoopTool-32B & 0.779$\pm$0.058 & 0.333$\pm$0.088 (-0.446$\pm$0.107) & 0.314$\pm$0.088 (-0.465$\pm$0.108) & 0.422$\pm$0.098 (-0.357$\pm$0.112) \\
\midrule
\multicolumn{5}{l}{\emph{Base / instruct counterparts}} \\
Qwen2.5-1.5B-Instruct & 0.573$\pm$0.070 & 0.225$\pm$0.078 (-0.347$\pm$0.106) & 0.196$\pm$0.074 (-0.377$\pm$0.100) & 0.275$\pm$0.088 (-0.298$\pm$0.111) \\
Qwen2.5-3B-Instruct & 0.608$\pm$0.065 & 0.206$\pm$0.078 (-0.402$\pm$0.103) & 0.147$\pm$0.064 (-0.461$\pm$0.097) & 0.324$\pm$0.088 (-0.285$\pm$0.114) \\
Llama-3.2-3B-Instruct & 0.518$\pm$0.070 & 0.137$\pm$0.064 (-0.380$\pm$0.095) & 0.108$\pm$0.059 (-0.410$\pm$0.092) & 0.225$\pm$0.083 (-0.292$\pm$0.107) \\
Qwen3-8B & 0.673$\pm$0.065 & 0.324$\pm$0.088 (-0.350$\pm$0.113) & 0.255$\pm$0.083 (-0.418$\pm$0.107) & 0.441$\pm$0.093 (-0.232$\pm$0.119) \\
Qwen3-14B & 0.749$\pm$0.060 & 0.333$\pm$0.088 (-0.415$\pm$0.111) & 0.284$\pm$0.088 (-0.464$\pm$0.106) & 0.402$\pm$0.098 (-0.347$\pm$0.113) \\
Qwen3-32B & 0.754$\pm$0.060 & 0.343$\pm$0.088 (-0.411$\pm$0.111) & 0.598$\pm$0.093 (-0.156$\pm$0.112) & 0.637$\pm$0.093 (-0.117$\pm$0.109) \\
\midrule
\multicolumn{5}{l}{\emph{Frontier (open weights, reasoning / general)}} \\
Qwen3.5-9B & 0.709$\pm$0.063 & 0.412$\pm$0.098 (-0.297$\pm$0.114) & 0.304$\pm$0.088 (-0.405$\pm$0.109) & 0.471$\pm$0.098 (-0.238$\pm$0.116) \\
DeepSeek-R1-Distill-14B & 0.618$\pm$0.065 & 0.422$\pm$0.098 (-0.197$\pm$0.119) & 0.412$\pm$0.098 (-0.206$\pm$0.115) & 0.461$\pm$0.098 (-0.157$\pm$0.117) \\
\midrule
\multicolumn{5}{l}{\emph{Closed-source frontier}} \\
o4-mini (OpenAI) & 0.709$\pm$0.063 & 0.608$\pm$0.098 (-0.101$\pm$0.114) & 0.569$\pm$0.098 (-0.140$\pm$0.116) & 0.549$\pm$0.098 (-0.160$\pm$0.116) \\
\midrule
\multicolumn{5}{l}{\emph{ToolRL SFT version}} \\
SFT-Clean-4k-3B & 0.598$\pm$0.070 & 0.186$\pm$0.074 (-0.412$\pm$0.102) & 0.127$\pm$0.064 (-0.471$\pm$0.092) & 0.314$\pm$0.088 (-0.284$\pm$0.111) \\
\bottomrule
\end{tabular*}
\end{table}

\begin{table}[H]
\centering
\caption{Per-perturbation accuracy under \textbf{Reward} perturbations (part 2/2). The first numeric column is Clean accuracy (the unperturbed baseline); each subsequent entry shows \emph{accuracy on that perturbation} $\pm$ 95\% bootstrap half-width, with the drop \emph{from this model's own Clean} (also $\pm$ 95\% half-width) in parentheses. A more negative parenthesised number means a larger sim-to-real gap (worse robustness). All values are accuracies in $[0,1]$, matching Table~\ref{tab:main}. $B={\hspace{-0.05em}}10{,}000$ percentile-bootstrap resamples on per-sample correctness scores. \textbf{Bold} = our trained checkpoints.}
\label{tab:per-pert-rew-2}
\scriptsize
\setlength{\tabcolsep}{3pt}
\begin{tabular*}{\linewidth}{@{\extracolsep{\fill}}lrrrr}
\toprule
 & Clean & TimeDesc-N & MisDesc-Abbr & TimeDesc-Abbr \\
\midrule
\multicolumn{5}{l}{\emph{Our 3B trained checkpoints}} \\
\textbf{ToolRL-DR-Full (ours)} & \textbf{0.643$\pm$0.065} & \textbf{0.461$\pm$0.098 (-0.182$\pm$0.120)} & \textbf{0.118$\pm$0.052 (-0.525$\pm$0.085)} & \textbf{0.139$\pm$0.059 (-0.504$\pm$0.087)} \\
\textbf{ToolRL-DR-Mixed (ours)} & \textbf{0.658$\pm$0.065} & \textbf{0.441$\pm$0.098 (-0.217$\pm$0.116)} & \textbf{0.104$\pm$0.052 (-0.554$\pm$0.082)} & \textbf{0.132$\pm$0.056 (-0.526$\pm$0.086)} \\
\midrule
\multicolumn{5}{l}{\emph{Open-source RL tool models, 8B--14B}} \\
MUA-RL-8B & 0.678$\pm$0.065 & 0.451$\pm$0.098 (-0.227$\pm$0.119) & 0.104$\pm$0.049 (-0.574$\pm$0.082) & 0.174$\pm$0.062 (-0.505$\pm$0.089) \\
MUA-RL-14B & 0.628$\pm$0.065 & 0.441$\pm$0.098 (-0.187$\pm$0.119) & 0.118$\pm$0.052 (-0.510$\pm$0.086) & 0.174$\pm$0.062 (-0.455$\pm$0.092) \\
LoopTool-8B & 0.714$\pm$0.063 & 0.441$\pm$0.098 (-0.272$\pm$0.114) & 0.097$\pm$0.049 (-0.616$\pm$0.079) & 0.146$\pm$0.059 (-0.568$\pm$0.086) \\
\midrule
\multicolumn{5}{l}{\emph{Open-source RL tool models, 1.5B--7B}} \\
ToolRL-Qwen2.5-1.5B & 0.704$\pm$0.063 & 0.373$\pm$0.093 (-0.331$\pm$0.113) & 0.076$\pm$0.042 (-0.627$\pm$0.077) & 0.090$\pm$0.045 (-0.613$\pm$0.078) \\
ToolRL-Qwen2.5-3B & 0.638$\pm$0.068 & 0.314$\pm$0.088 (-0.324$\pm$0.111) & 0.069$\pm$0.038 (-0.569$\pm$0.079) & 0.062$\pm$0.038 (-0.576$\pm$0.078) \\
ToolRL-Llama3.2-3B & 0.628$\pm$0.065 & 0.294$\pm$0.088 (-0.334$\pm$0.113) & 0.083$\pm$0.045 (-0.545$\pm$0.081) & 0.111$\pm$0.052 (-0.517$\pm$0.085) \\
TL-CodeLLaMA-2 & 0.653$\pm$0.065 & 0.373$\pm$0.098 (-0.281$\pm$0.113) & 0.042$\pm$0.031 (-0.612$\pm$0.073) & 0.049$\pm$0.035 (-0.605$\pm$0.075) \\
\midrule
\multicolumn{5}{l}{\emph{Open-source RL tool models, 32B}} \\
MUA-RL-32B & 0.683$\pm$0.065 & 0.431$\pm$0.093 (-0.252$\pm$0.116) & 0.104$\pm$0.052 (-0.579$\pm$0.082) & 0.188$\pm$0.066 (-0.496$\pm$0.092) \\
LoopTool-32B & 0.779$\pm$0.058 & 0.422$\pm$0.098 (-0.357$\pm$0.111) & 0.090$\pm$0.045 (-0.689$\pm$0.073) & 0.160$\pm$0.059 (-0.619$\pm$0.082) \\
\midrule
\multicolumn{5}{l}{\emph{Base / instruct counterparts}} \\
Qwen2.5-1.5B-Instruct & 0.573$\pm$0.068 & 0.235$\pm$0.078 (-0.338$\pm$0.107) & 0.076$\pm$0.045 (-0.496$\pm$0.082) & 0.076$\pm$0.045 (-0.496$\pm$0.081) \\
Qwen2.5-3B-Instruct & 0.608$\pm$0.068 & 0.304$\pm$0.088 (-0.304$\pm$0.111) & 0.042$\pm$0.031 (-0.566$\pm$0.075) & 0.049$\pm$0.035 (-0.559$\pm$0.077) \\
Llama-3.2-3B-Instruct & 0.518$\pm$0.070 & 0.255$\pm$0.083 (-0.263$\pm$0.108) & 0.076$\pm$0.045 (-0.441$\pm$0.083) & 0.111$\pm$0.052 (-0.406$\pm$0.086) \\
Qwen3-8B & 0.673$\pm$0.065 & 0.392$\pm$0.098 (-0.281$\pm$0.116) & 0.125$\pm$0.052 (-0.548$\pm$0.084) & 0.153$\pm$0.059 (-0.521$\pm$0.087) \\
Qwen3-14B & 0.749$\pm$0.060 & 0.461$\pm$0.098 (-0.288$\pm$0.115) & 0.139$\pm$0.059 (-0.610$\pm$0.082) & 0.188$\pm$0.066 (-0.561$\pm$0.087) \\
Qwen3-32B & 0.754$\pm$0.060 & 0.608$\pm$0.093 (-0.146$\pm$0.111) & 0.111$\pm$0.052 (-0.643$\pm$0.077) & 0.215$\pm$0.066 (-0.538$\pm$0.089) \\
\midrule
\multicolumn{5}{l}{\emph{Frontier (open weights, reasoning / general)}} \\
Qwen3.5-9B & 0.709$\pm$0.063 & 0.490$\pm$0.098 (-0.218$\pm$0.113) & 0.132$\pm$0.056 (-0.577$\pm$0.086) & 0.118$\pm$0.052 (-0.590$\pm$0.083) \\
DeepSeek-R1-Distill-14B & 0.618$\pm$0.068 & 0.412$\pm$0.098 (-0.206$\pm$0.117) & 0.111$\pm$0.052 (-0.507$\pm$0.084) & 0.132$\pm$0.056 (-0.486$\pm$0.087) \\
\midrule
\multicolumn{5}{l}{\emph{Closed-source frontier}} \\
o4-mini (OpenAI) & 0.709$\pm$0.063 & 0.578$\pm$0.098 (-0.130$\pm$0.113) & 0.104$\pm$0.049 (-0.604$\pm$0.081) & 0.111$\pm$0.052 (-0.597$\pm$0.082) \\
\midrule
\multicolumn{5}{l}{\emph{ToolRL SFT version}} \\
SFT-Clean-4k-3B & 0.598$\pm$0.068 & 0.206$\pm$0.078 (-0.392$\pm$0.104) & 0.076$\pm$0.045 (-0.522$\pm$0.080) & 0.083$\pm$0.045 (-0.515$\pm$0.082) \\
\bottomrule
\end{tabular*}
\end{table}

\begin{table}[H]
\centering
\caption{Per-perturbation accuracy under \textbf{Transition} perturbations (part 1/2). The first numeric column is Clean accuracy (the unperturbed baseline); each subsequent entry shows \emph{accuracy on that perturbation} $\pm$ 95\% bootstrap half-width, with the drop \emph{from this model's own Clean} (also $\pm$ 95\% half-width) in parentheses. A more negative parenthesised number means a larger sim-to-real gap (worse robustness). All values are accuracies in $[0,1]$, matching Table~\ref{tab:main}. $B={\hspace{-0.05em}}10{,}000$ percentile-bootstrap resamples on per-sample correctness scores. \textbf{Bold} = our trained checkpoints.}
\label{tab:per-pert-trn-1}
\scriptsize
\setlength{\tabcolsep}{3pt}
\begin{tabular*}{\linewidth}{@{\extracolsep{\fill}}lrrrr}
\toprule
 & Clean & Timeout & RateLim & AuthErr \\
\midrule
\multicolumn{5}{l}{\emph{Our 3B trained checkpoints}} \\
\textbf{ToolRL-DR-Full (ours)} & \textbf{0.643$\pm$0.065} & \textbf{0.548$\pm$0.070 (-0.095$\pm$0.095)} & \textbf{0.191$\pm$0.053 (-0.452$\pm$0.085)} & \textbf{0.216$\pm$0.058 (-0.427$\pm$0.088)} \\
\textbf{ToolRL-DR-Mixed (ours)} & \textbf{0.658$\pm$0.065} & \textbf{0.558$\pm$0.070 (-0.101$\pm$0.095)} & \textbf{0.191$\pm$0.053 (-0.467$\pm$0.083)} & \textbf{0.211$\pm$0.055 (-0.447$\pm$0.088)} \\
\midrule
\multicolumn{5}{l}{\emph{Open-source RL tool models, 8B--14B}} \\
MUA-RL-8B & 0.678$\pm$0.065 & 0.588$\pm$0.070 (-0.090$\pm$0.095) & 0.261$\pm$0.060 (-0.417$\pm$0.090) & 0.146$\pm$0.048 (-0.533$\pm$0.080) \\
MUA-RL-14B & 0.628$\pm$0.068 & 0.573$\pm$0.068 (-0.055$\pm$0.095) & 0.352$\pm$0.065 (-0.276$\pm$0.093) & 0.181$\pm$0.053 (-0.447$\pm$0.085) \\
LoopTool-8B & 0.714$\pm$0.063 & 0.698$\pm$0.063 (-0.015$\pm$0.090) & 0.276$\pm$0.063 (-0.437$\pm$0.090) & 0.065$\pm$0.033 (-0.648$\pm$0.073) \\
\midrule
\multicolumn{5}{l}{\emph{Open-source RL tool models, 1.5B--7B}} \\
ToolRL-Qwen2.5-1.5B & 0.704$\pm$0.063 & 0.337$\pm$0.065 (-0.367$\pm$0.090) & 0.256$\pm$0.060 (-0.447$\pm$0.088) & 0.302$\pm$0.063 (-0.402$\pm$0.090) \\
ToolRL-Qwen2.5-3B & 0.638$\pm$0.065 & 0.477$\pm$0.070 (-0.161$\pm$0.095) & 0.070$\pm$0.035 (-0.568$\pm$0.075) & 0.106$\pm$0.043 (-0.533$\pm$0.078) \\
ToolRL-Llama3.2-3B & 0.628$\pm$0.068 & 0.437$\pm$0.068 (-0.191$\pm$0.095) & 0.211$\pm$0.055 (-0.417$\pm$0.090) & 0.186$\pm$0.053 (-0.442$\pm$0.085) \\
TL-CodeLLaMA-2 & 0.653$\pm$0.065 & 0.307$\pm$0.065 (-0.347$\pm$0.091) & 0.312$\pm$0.065 (-0.342$\pm$0.093) & 0.357$\pm$0.065 (-0.296$\pm$0.095) \\
\midrule
\multicolumn{5}{l}{\emph{Open-source RL tool models, 32B}} \\
MUA-RL-32B & 0.683$\pm$0.065 & 0.643$\pm$0.065 (-0.040$\pm$0.090) & 0.241$\pm$0.058 (-0.442$\pm$0.088) & 0.136$\pm$0.048 (-0.548$\pm$0.080) \\
LoopTool-32B & 0.779$\pm$0.058 & 0.543$\pm$0.068 (-0.236$\pm$0.090) & 0.136$\pm$0.048 (-0.643$\pm$0.075) & 0.111$\pm$0.043 (-0.668$\pm$0.073) \\
\midrule
\multicolumn{5}{l}{\emph{Base / instruct counterparts}} \\
Qwen2.5-1.5B-Instruct & 0.573$\pm$0.070 & 0.432$\pm$0.070 (-0.141$\pm$0.098) & 0.317$\pm$0.065 (-0.256$\pm$0.095) & 0.322$\pm$0.065 (-0.251$\pm$0.093) \\
Qwen2.5-3B-Instruct & 0.608$\pm$0.065 & 0.543$\pm$0.068 (-0.065$\pm$0.098) & 0.191$\pm$0.053 (-0.417$\pm$0.085) & 0.196$\pm$0.055 (-0.412$\pm$0.088) \\
Llama-3.2-3B-Instruct & 0.518$\pm$0.070 & 0.412$\pm$0.070 (-0.106$\pm$0.098) & 0.201$\pm$0.053 (-0.317$\pm$0.090) & 0.236$\pm$0.058 (-0.281$\pm$0.090) \\
Qwen3-8B & 0.673$\pm$0.065 & 0.668$\pm$0.065 (-0.005$\pm$0.093) & 0.216$\pm$0.055 (-0.457$\pm$0.085) & 0.035$\pm$0.028 (-0.638$\pm$0.070) \\
Qwen3-14B & 0.749$\pm$0.060 & 0.729$\pm$0.060 (-0.020$\pm$0.085) & 0.492$\pm$0.070 (-0.256$\pm$0.093) & 0.131$\pm$0.048 (-0.618$\pm$0.075) \\
Qwen3-32B & 0.754$\pm$0.060 & 0.472$\pm$0.070 (-0.281$\pm$0.090) & 0.201$\pm$0.055 (-0.553$\pm$0.080) & 0.161$\pm$0.050 (-0.593$\pm$0.080) \\
\midrule
\multicolumn{5}{l}{\emph{Frontier (open weights, reasoning / general)}} \\
Qwen3.5-9B & 0.709$\pm$0.060 & 0.714$\pm$0.063 (+0.005$\pm$0.090) & 0.216$\pm$0.058 (-0.492$\pm$0.085) & 0.101$\pm$0.043 (-0.608$\pm$0.078) \\
DeepSeek-R1-Distill-14B & 0.618$\pm$0.068 & 0.427$\pm$0.070 (-0.191$\pm$0.098) & 0.377$\pm$0.065 (-0.241$\pm$0.095) & 0.312$\pm$0.065 (-0.307$\pm$0.093) \\
\midrule
\multicolumn{5}{l}{\emph{Closed-source frontier}} \\
o4-mini (OpenAI) & 0.709$\pm$0.063 & 0.623$\pm$0.065 (-0.085$\pm$0.093) & 0.070$\pm$0.035 (-0.638$\pm$0.070) & 0.095$\pm$0.040 (-0.613$\pm$0.075) \\
\midrule
\multicolumn{5}{l}{\emph{ToolRL SFT version}} \\
SFT-Clean-4k-3B & 0.598$\pm$0.065 & 0.035$\pm$0.025 (-0.563$\pm$0.070) & 0.010$\pm$0.013 (-0.588$\pm$0.068) & 0.010$\pm$0.013 (-0.588$\pm$0.070) \\
\bottomrule
\end{tabular*}
\end{table}

\begin{table}[H]
\centering
\caption{Per-perturbation accuracy under \textbf{Transition} perturbations (part 2/2). The first numeric column is Clean accuracy (the unperturbed baseline); each subsequent entry shows \emph{accuracy on that perturbation} $\pm$ 95\% bootstrap half-width, with the drop \emph{from this model's own Clean} (also $\pm$ 95\% half-width) in parentheses. A more negative parenthesised number means a larger sim-to-real gap (worse robustness). All values are accuracies in $[0,1]$, matching Table~\ref{tab:main}. $B={\hspace{-0.05em}}10{,}000$ percentile-bootstrap resamples on per-sample correctness scores. \textbf{Bold} = our trained checkpoints.}
\label{tab:per-pert-trn-2}
\scriptsize
\setlength{\tabcolsep}{3pt}
\begin{tabular*}{\linewidth}{@{\extracolsep{\fill}}lrrrr}
\toprule
 & Clean & 5xxErr & Malform & SchemaD \\
\midrule
\multicolumn{5}{l}{\emph{Our 3B trained checkpoints}} \\
\textbf{ToolRL-DR-Full (ours)} & \textbf{0.643$\pm$0.068} & \textbf{0.472$\pm$0.070 (-0.171$\pm$0.095)} & \textbf{0.593$\pm$0.070 (-0.050$\pm$0.095)} & \textbf{0.452$\pm$0.068 (-0.191$\pm$0.098)} \\
\textbf{ToolRL-DR-Mixed (ours)} & \textbf{0.658$\pm$0.065} & \textbf{0.492$\pm$0.070 (-0.166$\pm$0.098)} & \textbf{0.608$\pm$0.068 (-0.050$\pm$0.093)} & \textbf{0.497$\pm$0.070 (-0.161$\pm$0.095)} \\
\midrule
\multicolumn{5}{l}{\emph{Open-source RL tool models, 8B--14B}} \\
MUA-RL-8B & 0.678$\pm$0.065 & 0.538$\pm$0.070 (-0.141$\pm$0.095) & 0.678$\pm$0.065 (+0.000$\pm$0.090) & 0.482$\pm$0.070 (-0.196$\pm$0.095) \\
MUA-RL-14B & 0.628$\pm$0.065 & 0.487$\pm$0.070 (-0.141$\pm$0.098) & 0.588$\pm$0.070 (-0.040$\pm$0.095) & 0.377$\pm$0.065 (-0.251$\pm$0.095) \\
LoopTool-8B & 0.714$\pm$0.063 & 0.442$\pm$0.070 (-0.271$\pm$0.095) & 0.658$\pm$0.065 (-0.055$\pm$0.090) & 0.367$\pm$0.065 (-0.347$\pm$0.090) \\
\midrule
\multicolumn{5}{l}{\emph{Open-source RL tool models, 1.5B--7B}} \\
ToolRL-Qwen2.5-1.5B & 0.704$\pm$0.065 & 0.347$\pm$0.065 (-0.357$\pm$0.090) & 0.337$\pm$0.065 (-0.367$\pm$0.090) & 0.372$\pm$0.068 (-0.332$\pm$0.093) \\
ToolRL-Qwen2.5-3B & 0.638$\pm$0.065 & 0.337$\pm$0.065 (-0.302$\pm$0.095) & 0.568$\pm$0.070 (-0.070$\pm$0.095) & 0.382$\pm$0.065 (-0.256$\pm$0.095) \\
ToolRL-Llama3.2-3B & 0.628$\pm$0.068 & 0.447$\pm$0.070 (-0.181$\pm$0.095) & 0.492$\pm$0.070 (-0.136$\pm$0.098) & 0.216$\pm$0.058 (-0.412$\pm$0.090) \\
TL-CodeLLaMA-2 & 0.653$\pm$0.065 & 0.246$\pm$0.058 (-0.407$\pm$0.088) & 0.422$\pm$0.068 (-0.231$\pm$0.095) & 0.276$\pm$0.063 (-0.377$\pm$0.090) \\
\midrule
\multicolumn{5}{l}{\emph{Open-source RL tool models, 32B}} \\
MUA-RL-32B & 0.683$\pm$0.065 & 0.598$\pm$0.065 (-0.085$\pm$0.095) & 0.734$\pm$0.060 (+0.050$\pm$0.085) & 0.442$\pm$0.068 (-0.241$\pm$0.095) \\
LoopTool-32B & 0.779$\pm$0.058 & 0.533$\pm$0.070 (-0.246$\pm$0.090) & 0.603$\pm$0.065 (-0.176$\pm$0.090) & 0.372$\pm$0.065 (-0.407$\pm$0.088) \\
\midrule
\multicolumn{5}{l}{\emph{Base / instruct counterparts}} \\
Qwen2.5-1.5B-Instruct & 0.573$\pm$0.070 & 0.387$\pm$0.068 (-0.186$\pm$0.095) & 0.392$\pm$0.068 (-0.181$\pm$0.095) & 0.357$\pm$0.065 (-0.216$\pm$0.095) \\
Qwen2.5-3B-Instruct & 0.608$\pm$0.070 & 0.397$\pm$0.068 (-0.211$\pm$0.095) & 0.543$\pm$0.070 (-0.065$\pm$0.095) & 0.357$\pm$0.068 (-0.251$\pm$0.095) \\
Llama-3.2-3B-Instruct & 0.518$\pm$0.068 & 0.452$\pm$0.070 (-0.065$\pm$0.098) & 0.472$\pm$0.070 (-0.045$\pm$0.101) & 0.286$\pm$0.063 (-0.231$\pm$0.093) \\
Qwen3-8B & 0.673$\pm$0.065 & 0.372$\pm$0.065 (-0.302$\pm$0.093) & 0.583$\pm$0.068 (-0.090$\pm$0.095) & 0.407$\pm$0.068 (-0.266$\pm$0.095) \\
Qwen3-14B & 0.749$\pm$0.060 & 0.633$\pm$0.065 (-0.116$\pm$0.090) & 0.598$\pm$0.068 (-0.151$\pm$0.090) & 0.407$\pm$0.068 (-0.342$\pm$0.090) \\
Qwen3-32B & 0.754$\pm$0.060 & 0.497$\pm$0.070 (-0.256$\pm$0.090) & 0.593$\pm$0.068 (-0.161$\pm$0.090) & 0.337$\pm$0.065 (-0.417$\pm$0.090) \\
\midrule
\multicolumn{5}{l}{\emph{Frontier (open weights, reasoning / general)}} \\
Qwen3.5-9B & 0.709$\pm$0.063 & 0.276$\pm$0.060 (-0.432$\pm$0.088) & 0.266$\pm$0.060 (-0.442$\pm$0.088) & 0.176$\pm$0.053 (-0.533$\pm$0.083) \\
DeepSeek-R1-Distill-14B & 0.618$\pm$0.068 & 0.427$\pm$0.070 (-0.191$\pm$0.095) & 0.482$\pm$0.070 (-0.136$\pm$0.095) & 0.407$\pm$0.068 (-0.211$\pm$0.095) \\
\midrule
\multicolumn{5}{l}{\emph{Closed-source frontier}} \\
o4-mini (OpenAI) & 0.709$\pm$0.063 & 0.613$\pm$0.070 (-0.095$\pm$0.095) & 0.668$\pm$0.065 (-0.040$\pm$0.090) & 0.523$\pm$0.070 (-0.186$\pm$0.093) \\
\midrule
\multicolumn{5}{l}{\emph{ToolRL SFT version}} \\
SFT-Clean-4k-3B & 0.598$\pm$0.065 & 0.020$\pm$0.018 (-0.578$\pm$0.070) & 0.025$\pm$0.023 (-0.573$\pm$0.070) & 0.025$\pm$0.023 (-0.573$\pm$0.070) \\
\bottomrule
\end{tabular*}
\end{table}

\section{Per-model evaluation settings}
\label{app:eval-details}

Table~\ref{tab:eval-settings} lists the vLLM serving and decoding parameters for every model evaluated in this paper. The same table is auto-generated from \emph{scripts/model\_manifest.py} so future re-runs cannot drift from the canonical config.

\begin{table}[H]
\centering
\caption{Per-model evaluation settings used in this paper. Mode: FC = function-calling (tools passed via OpenAI \emph{tools=} API + vLLM \emph{hermes} parser); Prompt = tools embedded in system prompt as text. Thinking: Qwen3 family evaluated with thinking explicitly disabled (\emph{enable\_thinking=False}). Defaults: temperature 0.0, max\_tokens 1024 (DeepSeek-R1-Distill: temp 0.6, max\_tokens 4096 per HF card). Sourced from \emph{scripts/model\_manifest.py}.}
\label{tab:eval-settings}
\scriptsize
\setlength{\tabcolsep}{2pt}
\begin{tabular}{lllrrllrr}
\toprule
\textbf{Model} & \textbf{HF id} & \textbf{GPUs} & \textbf{dtype} & \textbf{max\_len} & \textbf{Mode} & \textbf{Thinking} & \textbf{Temp} & \textbf{max\_tok} \\
\midrule
ToolRL-Qwen2.5-1.5B & chengq9/ToolRL-Qwen2.5-1.5B & 1 & float32 & 8192 & Prompt & --- & 0.0 & 1024 \\
ToolRL-Qwen2.5-3B & chengq9/ToolRL-Qwen2.5-3B & 1 & bfloat16 & 8192 & Prompt & --- & 0.0 & 1024 \\
ToolRL-Llama3.2-3B & chengq9/ToolRL-Llama3.2-3B & 1 & bfloat16 & 8192 & Prompt & --- & 0.0 & 1024 \\
TL-CodeLLaMA-2 & Junjie-Ye/TL-CodeLLaMA-2 & 1 & bfloat16 & 16384 & Prompt & --- & 0.0 & 1024 \\
LoopTool-8B & zhuiguang-ning/LoopTool-8B & 1 & bfloat16 & 16384 & FC & off & 0.0 & 1024 \\
LoopTool-32B & zhuiguang-ning/LoopTool-32B & 1 & bfloat16 & 16384 & FC & off & 0.0 & 1024 \\
MUA-RL-8B & zzwkk/MUA-RL-8B & 1 & bfloat16 & 16384 & FC & off & 0.0 & 1024 \\
MUA-RL-14B & zzwkk/MUA-RL-14B & 1 & bfloat16 & 16384 & FC & off & 0.0 & 1024 \\
MUA-RL-32B & zzwkk/MUA-RL-32B & 1 & bfloat16 & 16384 & FC & off & 0.0 & 1024 \\
Qwen2.5-1.5B-Instruct & Qwen/Qwen2.5-1.5B-Instruct & 1 & bfloat16 & 8192 & Prompt & --- & 0.0 & 1024 \\
Qwen2.5-3B-Instruct & Qwen/Qwen2.5-3B-Instruct & 1 & bfloat16 & 8192 & Prompt & --- & 0.0 & 1024 \\
Llama-3.2-3B-Instruct & meta-llama/Llama-3.2-3B-Instruct & 1 & bfloat16 & 8192 & Prompt & --- & 0.0 & 1024 \\
Qwen3-8B & Qwen/Qwen3-8B & 1 & bfloat16 & 16384 & FC & off & 0.0 & 1024 \\
Qwen3-14B & Qwen/Qwen3-14B & 1 & bfloat16 & 16384 & FC & off & 0.0 & 1024 \\
Qwen3-32B & Qwen/Qwen3-32B & 1 & bfloat16 & 16384 & FC & off & 0.0 & 1024 \\
Qwen3.5-9B & Qwen/Qwen3.5-9B & 1 & bfloat16 & 16384 & FC & off & 0.0 & 1024 \\
DeepSeek-R1-Distill-14B & deepseek-ai/DeepSeek-R1-Distill-Qwen-14B & 1 & bfloat16 & 16384 & Prompt & --- & 0.6 & 4096 \\
SFT-Clean-4k-3B & \emph{(ToolRL SFT version)} & 1 & bfloat16 & 8192 & Prompt & --- & 0.0 & 1024 \\
ToolRL-DR-Full & \emph{see training recipe} & 1 & bfloat16 & 8192 & Prompt & --- & 0.0 & 1024 \\
ToolRL-DR-Mixed & \emph{see training recipe} & 1 & bfloat16 & 8192 & Prompt & --- & 0.0 & 1024 \\
\bottomrule
\end{tabular}
\end{table}

\section{Error-mode classification rubric}
\label{app:error-modes}

We classify each scored-incorrect prediction by inspecting the raw model output \emph{before} the format-tolerant tool-call parser. The rule order is deterministic and reported here so that any reviewer can reproduce the §\ref{sec:exp-main} percentages from the released \emph{*.predictions.jsonl} files.

\paragraph{Empty tool call.} The model's \emph{raw\_output} is empty or whitespace-only after stripping the chat template. Detection: \texttt{not raw.strip()}. Typical cause: vLLM hit a max-token cap mid-EOS or the chat template's \emph{generation\_prompt} consumed the entire budget; common on small Qwen2.5 models when the system prompt is long.

\paragraph{Omitted tool call.} The model emitted text but did not produce a parseable tool call (every parser variant in \S\ref{app:scoring} returned an empty \emph{tool\_calls} list). Detection: \texttt{tool\_calls == []}. Typical wording: ``It seems the tool timed out, please try again later'' or ``I cannot help with this request''. This is the dominant failure mode under transition perturbations.

\paragraph{Wrong parameter / wrong name.} The model produced a parseable tool call but the call did not match the ground truth (wrong name, missing required parameter, or wrong parameter value). Detection: \texttt{tool\_calls != [] AND score=0}. Subdivides into name-bias (under Reward perturbations the model picked a distractor whose name suggests efficiency) and parameter mismatch (under Observation perturbations the model echoed a corrupted argument).

\paragraph{Other.} A small residual where the parser accepted the call but the scorer marks it incorrect for benchmark-specific reasons (e.g. order-sensitive multi-tool sequences in BFCL). Reported separately when material; folded into ``wrong'' otherwise.

\section{Compute disclosure}
\label{app:compute}

\paragraph{Benchmark generation.} The four LLM-generated perturbations (typos, query paraphrase, tool-description paraphrase, parameter-description paraphrase) were produced with OpenAI \emph{gpt-5-mini} as the primary generator and \emph{gpt-4o-mini} for a supplementary typo pass; rule-based perturbations (action and reward) require no LLM calls. Total LLM API spend for the benchmark generation pass is under 10~USD. The paraphrase quality audit (Appendix~\ref{app:paraphrase-audit}) used \emph{gpt-5} as a judge over 196 (clean, perturbed) pairs.

\paragraph{Evaluation.} All evaluation runs were performed on a single node with up to two NVIDIA GH200 Grace-Hopper GPUs (96\,GB HBM3e per GPU). Each model was served via vLLM and evaluated in a single end-to-end SLURM job covering the clean baseline, the 16 statically-augmentable perturbations, and the 6 transition variants. Wall-clock per model ranged from 4 minutes (Qwen2.5-1.5B-Instruct) to 45 minutes (DeepSeek-R1-Distill-14B at \emph{max\_tokens}=4096); the 32B function-calling models (MUA-RL-32B, LoopTool-32B, Qwen3-32B) ran in 10--17 minutes per job on a single GH200. The total evaluation budget for the 20 locally-served models (o4-mini, accessed via the OpenAI Chat Completions API, is excluded from this GPU accounting) is approximately 4.3~GPU-hours of GH200 time.

\paragraph{Training.} Both ToolRL-DR checkpoints are trained with the public veRL implementation of GRPO~\citep{shao2024deepseekmath} from the ToolRL fork (commit \texttt{8cee13e}). The base model is \texttt{Qwen/Qwen2.5-3B-Instruct} (snapshot \texttt{aa8e7253...}), trained in bf16 on 2$\times$ NVIDIA RTX 6000 Ada (48\,GB each) with tensor-parallel size 1 and data parallel over both GPUs; rollouts are served by vLLM 0.6.x. Hyperparameters are identical across the two runs and are listed in Table~\ref{tab:hyperparams}.

\begin{table}[H]
\centering
\caption{ToolRL-DR training hyperparameters: backbone, optimization, and reward. Identical for ToolRL-DR-Full and ToolRL-DR-Mixed.}
\label{tab:hyperparams}
\small
\begin{tabular}{ll}
\toprule
\textbf{Setting} & \textbf{Value} \\
\midrule
\multicolumn{2}{l}{\emph{Backbone}} \\
\quad Base model              & \texttt{Qwen/Qwen2.5-3B-Instruct} \\
\quad Hardware                & 2$\times$ NVIDIA RTX 6000 Ada (48\,GB each) \\
\quad Precision               & bf16 \\
\quad Tensor parallel         & 1 (data-parallel across the 2 GPUs) \\
\quad Rollout server          & vLLM 0.6.x \\
\midrule
\multicolumn{2}{l}{\emph{Optimization (GRPO, veRL implementation)}} \\
\quad Train batch size        & 64 \\
\quad PPO mini-batch          & 32 \\
\quad PPO micro-batch / GPU   & 16 \\
\quad Max tokens / GPU        & 12{,}288 \\
\quad Rollout group size $K$  & 4 \\
\quad Max prompt length       & 2{,}048 \\
\quad Max response length     & 1{,}024 \\
\quad Learning rate           & $1\!\times\!10^{-6}$ \\
\quad Epochs                  & 3 \\
\quad Checkpoint freq.        & every 5 steps \\
\quad Eval freq.              & every 10 steps \\
\midrule
\multicolumn{2}{l}{\emph{Reward (ToolRL default; no variants from the ToolRL flag set)}} \\
\quad Format reward           & max 1.0 \\
\quad Correctness reward      & max 3.0 \\
\quad Total reward            & max 4.0 \\
\bottomrule
\end{tabular}
\end{table}

\begin{table}[H]
\centering
\caption{ToolRL-DR training data and wall-clock. The two runs differ only in the training-set composition (rows 1--2). The training data and reproduction recipe are available at \url{https://github.com/WillChow66/AgentRobustBench}.}
\label{tab:hyperparams-runs}
\small
\begin{tabular}{p{3.0cm}p{10.0cm}}
\toprule
\textbf{Setting} & \textbf{Value} \\
\midrule
ToolRL-DR-Full data    & 3{,}905 train + 79 val, 100\% perturbed (uniform across 4 obs / 6 act / 6 rew types) \\
ToolRL-DR-Mixed data   & 4{,}000 train (2{,}006 clean / 1{,}994 perturbed; perturbed half drawn uniformly from same 16 types as Full) + 79 val (the same val parquet is reused) \\
ToolRL-DR-Full clock   & $\approx$\,9\,h\,19\,m wall-clock, 153 optimizer steps, $\approx$\,3.6\,min/step \\
ToolRL-DR-Mixed clock  & $\approx$\,9\,h\,14\,m wall-clock, 156 optimizer steps \\
\bottomrule
\end{tabular}
\end{table}

End-to-end evaluation of one checkpoint over the 3{,}721-sample AgentRobustBench set takes $\approx$\,5 minutes on a single GH200, or $\approx$\,7 minutes on a single RTX 6000 Ada.

\paragraph{Aggregate footprint.} Total compute is split across two hardware tiers:

\begin{itemize}[leftmargin=*, noitemsep]
    \item Evaluation (NCSA DeltaAI, GH200 96\,GB): $\approx$\,4.3 GPU-hours across the 20 locally-served models; the 32B function-calling models account for the largest share at $\approx$\,0.5--0.6 GPU-hours each.
    \item Training (single workstation, 2$\times$ RTX 6000 Ada 48\,GB): two GRPO runs of 9.3 hours each, run sequentially on the same machine = $\approx$\,18.6 wall-clock hours / $\approx$\,37 GPU-hours total for the two ToolRL-DR checkpoints.
    \item Benchmark generation: $<$10~USD of OpenAI API spend; CPU-bound rule-based perturbations take $<$1 hour on a laptop.
\end{itemize}

The aggregate is small relative to the upstream pretraining cost of any of the evaluated models, and we re-used existing \emph{*.predictions.jsonl} via the reparse pipeline (\S\ref{app:scoring}) to avoid re-inference whenever a parser was extended.

\section{Future perturbation axes}
\label{app:future-perturbations}

During evidence collection (\S\ref{sec:bench-grounding}) we identified eight additional production-failure modes that do not cleanly fit the four-component taxonomy or require runtime infrastructure beyond the present scope. We list them here as a roadmap for future versions of AgentRobustBench, with the verified GitHub issues that motivate each.

\paragraph{(1) Fabricated observation (Reward $\cap$ Observation).} Tool returns plausible but incorrect data. Issue evidence: datarobot-community/af-component-agent\#461 (LLM ignores \emph{stop=["Observation:"]} and fabricates a tool response), epam/statgpt-backend\#249 (model claims data without actually calling the tool).

\paragraph{(2) Silent no-op (Reward).} Tool returns success without changing state, leading to infinite loops. Issue evidence: xorbitsai/xagent\#230 (JSON parse errors are dropped from history, model retries identically forever), alibaba/page-agent\#392 (4+ identical clicks in a row).

\paragraph{(3) Context pressure (Observation).} Conversation history grows long enough that key tool-relevant context is crowded out. Issue evidence: hermes-agent\#3838, claude-code\#38044.

\paragraph{(4) Prompt injection in tool result (Observation).} Tool output contains an instruction that the model treats as part of its system prompt. Issue evidence: agent-framework\#5024, openclaw\#39231.

\paragraph{(5) Stale resource (Transition).} The tool succeeds but returns a cached/stale snapshot of the underlying state. Issue evidence: vscode\#296892.

\paragraph{(6) Prerequisite dependency (Transition).} A tool fails because its prerequisite tool was not called or returned an error the agent failed to inspect. Issue evidence: langchain\#36503, langgraph\#7117.

\paragraph{(7) Tool hallucination / ghost tool (Action).} Model emits a call to a tool that does not exist in the registered list (often a related tool from a sibling library). Issue evidence: craft-agents-oss\#449, adk-python\#4775.

\paragraph{(8) Parameter type mismatch (Transition).} Tool fails because the model passes a value of the wrong type (e.g., string where an integer was expected). Issue evidence: pfsense-mcp-server\#7, gsd-2\#3361.

The first four are \emph{Observation}/\emph{Reward}-style and could be added as static data perturbations following the existing recipe. The remaining four require a runtime tool-execution harness (a sandbox or live API), which we leave to a future version of the leaderboard platform that supports live execution.

\section{Case studies}
\label{app:case-studies}

We illustrate three representative failure modes with verbatim outputs from ToolRL-Qwen2.5-3B on samples that the model answers correctly on the clean baseline (one Transition, one Reward, and one Observation case). Each case shows the user query, the available tools, the ground-truth call, the clean output, the perturbed output, and a brief diagnosis.

\paragraph{Case 1 (Timeout, API-Bank \emph{level1\_213}).}
\textbf{User query (last turn):} \emph{``Can you help me query my balance? My user token is p9o8i7u6y5t4r3e2w1q.''} \\
\textbf{Available tools:} OpenBankAccount, QueryStock, QueryBalance, GetUserToken, plus the API-Bank \emph{**Think} meta-tool. \\
\textbf{GT call:} \texttt{QueryBalance(token="p9o8i7u6y5t4r3e2w1q")} \\
\textbf{Clean run output (parsed: \texttt{[QueryBalance(token=...)]}, score=1):} \emph{``Of course, I will use your user token to query your account balance. \texttt{<tool\_call>\{\allowbreak"name":"QueryBalance",\allowbreak"parameters":\{\allowbreak"token":"p9o8i7u6y5t4r3e2w1q"\}\}\allowbreak</tool\_call>} Your current account balance is \$1000.''} \\
\textbf{Pass-1 output under transition\_timeout:} identical to clean (the perturbation injects an error \emph{after} the first call). \\
\textbf{Pass-2 output (parsed: \texttt{[]}, score=0):} \emph{``It seems like there was an issue retrieving your account balance. Let's try this again. Could you please provide your token once more?''} \\
\emph{Diagnosis:} the model treats the injected timeout as terminal and asks the user to re-supply data, instead of retrying the same tool call.

\paragraph{Case 2 (MisDesc-Abbr, BFCL \emph{multiple\_2}).}
\textbf{User query:} \emph{``What is the capital of Brazil?''} \\
\textbf{Clean tools:} \texttt{country\_info.largest\_city}, \texttt{country\_info.capital}, \texttt{country\_info.population}. \\
\textbf{Perturbed tools (MisDesc-Abbr):} as above plus a fourth distractor \texttt{cou\_info.cap}, which the perturbation marks as the new ground-truth tool. \\
\textbf{Clean GT call:} \texttt{country\_info.capital(country="Brazil")}. \\
\textbf{Perturbed GT call:} \texttt{cou\_info.cap(country="Brazil")}. \\
\textbf{Clean output (parsed: \texttt{[country\_info.capital(country="Brazil")]}, score=1):} \emph{``[country\_info.capital(country=\"Brazil\")]''} \\
\textbf{Perturbed output (parsed: \texttt{[country\_info.capital(country="Brazil")]}, score=0):} \emph{``country\_info.capital(country=\"Brazil\")''} \\
\emph{Diagnosis:} the model picks the unabbreviated tool whose name maps directly to the user's word ``capital''. The abbreviated variant \texttt{cou\_info.cap} is identifiable only via its description, which the model does not consult in this single-turn setting, a clear instance of name bias.

\paragraph{Case 3 (ToolPara, API-Bank \emph{level3\_26}).}
\textbf{User query (final turn):} \emph{``Book a meeting for me, including all employees in the Tencent who are not traveling. Today is 2023.6.8, the meeting is from 14:00 to 15:00. \dots Based on our conversation above, please only make one tool call to solve my need.''} \\
\textbf{Available tools:} \texttt{OrganizationMembers}, plus the API-Bank \emph{**Think} meta-tool. The tool name is identical between clean and perturbed; only the natural-language description has been paraphrased. \\
\textbf{GT call:} \texttt{OrganizationMembers(organization="Tencent")}. \\
\textbf{Clean run (parsed: \texttt{[OrganizationMembers(organization="Tencent")]}, score=1):} the model reasons that the meeting cannot be booked directly with the available tools, and decides the best one-call answer is to fetch the Tencent member list. \\
\textbf{Perturbed run (parsed: \texttt{[]}, score=0):} the model's reasoning trace now reads \emph{``the available tool 'OrganizationMembers' only provides a list of members in the organization without filtering by travel \dots''} and concludes that no tool call is appropriate, emitting a textual response only. \\
\emph{Diagnosis:} the paraphrased description preserves intent in the human reading (and our quality audit, Appendix~\ref{app:paraphrase-audit}, confirms it). The model's interpretation, however, becomes more conservative: it now reads the description as foreclosing a downstream filtering step, and gives up. This is the documented failure mode behind LlamaIndex \#16757 and the description-sensitivity finding of Faghih et al.~\citep{faghih2025preferences}.

\section{Live leaderboard screenshots}
\label{app:leaderboard}

\begin{figure}[H]
    \centering
    \includegraphics[width=\linewidth]{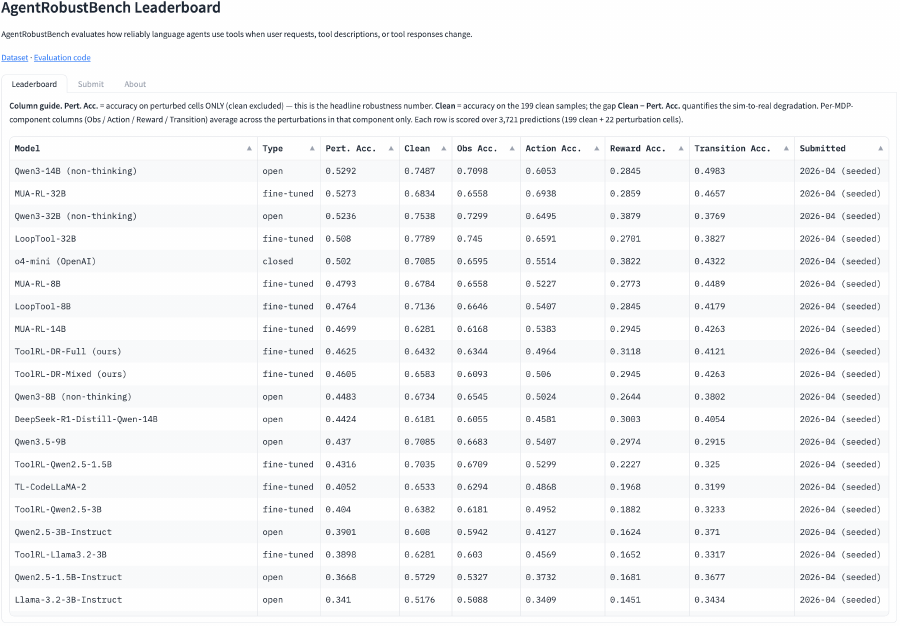}
    \caption{Live AgentRobustBench leaderboard, main view. The table is sorted by \emph{Pert.\ Acc.} (perturbed-only weighted average across the 22 perturbation types, clean excluded) in descending order. Each row reports per-POMDP-component accuracy alongside Clean and the headline Pert.\ Acc.}
    \label{fig:leaderboard-main}
\end{figure}

\begin{figure}[H]
    \centering
    \includegraphics[width=\linewidth]{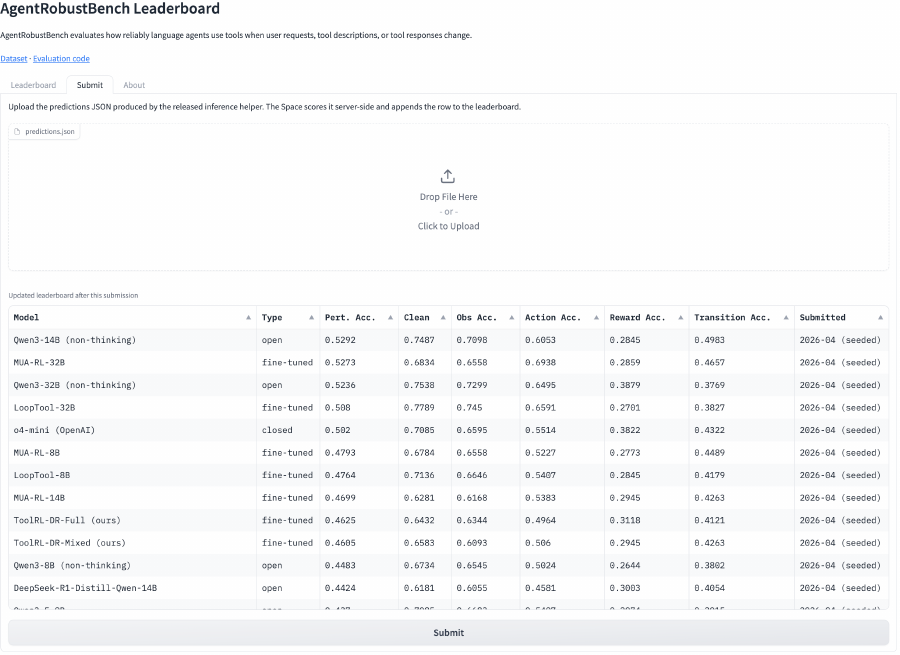}
    \caption{Live AgentRobustBench leaderboard, submit view. Contributors run inference locally with the released helper and upload the resulting predictions JSON; the Space scores it server-side using the same deterministic scorer as in the paper and appends the row to the table.}
    \label{fig:leaderboard-submit}
\end{figure}

Figures~\ref{fig:leaderboard-main} and~\ref{fig:leaderboard-submit} show the current state of the live AgentRobustBench leaderboard on HuggingFace Spaces. The Space follows the BFCL submission pattern (contributors run inference locally, upload a predictions JSON, the Space scores it server-side and appends a row to the table); see \S\ref{sec:platform} for the design rationale.\footnote{\url{https://huggingface.co/spaces/willchow66/AgentRobustBench}}

\section{Broader impact}
\label{sec:impact}

\textbf{Intended use.} Our benchmark and method are intended to support robustness research for tool-use language agents and to make deployment-time failure modes more visible to model developers.

\textbf{Misuse risk.} The perturbation taxonomy describes failures that already occur in production. Releasing it does not increase the attack surface beyond what is already documented in the source GitHub issues. The trained ToolRL-DR checkpoints are designed to be more robust to these perturbations, not to amplify any harmful behavior.

\textbf{Compute and environmental cost.} The 21-model evaluation sweep used approximately 4.3 GH200 GPU-hours; benchmark generation used $<$10\,USD of OpenAI API spend; training the two ToolRL-DR checkpoints used a 2$\times$ RTX 6000 Ada workstation for $\approx$\,9.3 hours per run ($\approx$\,37 GPU-hours total). The aggregate footprint is small relative to the upstream pretraining of any of the evaluated models. Full hyperparameters and per-model decoding parameters are in Appendices~\ref{app:compute} and~\ref{app:eval-details}.

\textbf{Data licensing.} Source benchmarks (BFCL V3, API-Bank, RoTBench, ToolAlpaca, ToolEyes) are redistributed under their original licenses; our derived perturbation files inherit the same terms. The HuggingFace dataset card lists the per-source license.

\clearpage
\section{Data sources and additional experiments}
\label{app:additional-validation}

\subsection{Training and evaluation data}
\label{app:data-provenance}
Figure~\ref{fig:data-provenance} traces the examples used for model training, checkpoint selection, and evaluation. Training and evaluation share zero sample IDs.

\begin{figure}[H]
\centering
\begin{minipage}[t]{0.47\linewidth}
\centering
\fbox{\parbox{0.90\linewidth}{\centering\textbf{Training sources}\\ToolACE + Hammer + xLAM\\ToolRL \texttt{rlla\_4k}}}
\par\smallskip$\downarrow$\par\smallskip
\fbox{\parbox{0.90\linewidth}{\centering\textbf{Static perturbation augmentation}\\Observation / Action / Reward}}
\par\smallskip$\downarrow$\par\smallskip
\fbox{\parbox{0.90\linewidth}{\centering\textbf{Training and model selection}\\DR-Full: 3{,}905 train + 79 val\\DR-Mixed: 4{,}000 train + same 79 val\\No runtime Transition training}}
\end{minipage}\hfill
\begin{minipage}[t]{0.47\linewidth}
\centering
\fbox{\parbox{0.90\linewidth}{\centering\textbf{Evaluation sources}\\BFCL V3 / API-Bank / RoTBench\\ToolAlpaca / ToolEyes}}
\par\smallskip$\downarrow$\par\smallskip
\fbox{\parbox{0.90\linewidth}{\centering\textbf{199 clean evaluation seeds}\\Valid, unique ground-truth calls}}
\par\smallskip$\downarrow$\par\smallskip
\fbox{\parbox{0.90\linewidth}{\centering\textbf{AgentRobustBench evaluation}\\Clean + static perturbations\\+ runtime Transition errors\\3{,}721 evaluation cases in total}}
\end{minipage}
\caption{Data sources and processing steps for training, checkpoint selection, and robustness evaluation.}
\label{fig:data-provenance}
\end{figure}

\begin{table}[H]
\centering
\caption{Sources of training, validation, and evaluation examples. Evaluation-source counts follow Table~\ref{tab:sources}.}
\label{tab:data-provenance}
\small
\begin{tabularx}{\linewidth}{l X l}
\toprule
\textbf{Role} & \textbf{Source} & \textbf{Use} \\
\midrule
Training & ToolRL \texttt{rlla\_4k}: ToolACE, Hammer, xLAM & GRPO updates \\
Validation & 79 held-out examples from the training corpus & Checkpoint selection \\
Evaluation seeds & BFCL V3 (32), API-Bank (74), RoTBench (21), ToolAlpaca (21), ToolEyes (51) & Robustness evaluation \\
\bottomrule
\end{tabularx}
\end{table}

\subsection{Seed selection and diversity}
The filter in Section~\ref{sec:bench-sources} yields different numbers of eligible tasks across sources (Table~\ref{tab:sources}). Across all 19{,}701 pairs of clean tasks, the mean TF-IDF cosine similarity is 0.050, the median is 0.000, and 0.07\% of pairs exceed 0.8.

\subsection{Reward accuracy with and without Reward perturbations in training}
\label{app:reward-transfer}
We trained two Qwen2.5-3B-Instruct models with GRPO. One used Observation and Action perturbations. The other also used Reward perturbations. Each run used 1{,}881 distinct perturbed examples, up-sampled to 3{,}905, with the same hyperparameters and number of optimization steps. Both training sets excluded runtime Transition errors.

\begin{table}[H]
\centering
\caption{Accuracy on Reward perturbations by training condition. The Reward gap is clean accuracy minus Reward accuracy. The two middle rows use equally sized training sets. DR-Full provides a reference using the full training set.}
\label{tab:reward-transfer}
\small
\begin{tabularx}{\linewidth}{X r r}
\toprule
\textbf{Condition} & \textbf{Reward accuracy $\uparrow$} & \textbf{Reward gap $\downarrow$} \\
\midrule
ToolRL-Clean & 0.188 & 0.450 \\
Observation and Action & 0.230 & 0.388 \\
Observation, Action, and Reward & 0.303 & 0.345 \\
ToolRL-DR-Full (full-data reference) & 0.312 & 0.331 \\
\bottomrule
\end{tabularx}
\end{table}

Training on Observation and Action raises Reward accuracy by 4.2 percentage points over ToolRL-Clean (paired bootstrap over 696 Reward evaluation cases, 95\% CI [1.6, 6.8] percentage points, $p=0.003$). Including Reward perturbations in the equally sized training set adds 7.3 percentage points ($p<0.001$).

\subsection{Composite perturbations and persistent errors}
\label{app:stress-tests}
On ToolRL-Qwen2.5-3B, we combine a query typo with a misleading-tool Reward perturbation on the same task. Accuracy is 6.9\% with the Reward perturbation alone and 6.9\% with both changes.

The default Timeout test injects an error on the first tool call. In a second test, the same error recurs on retries, up to three injections. Recovery accuracy falls from 47.7\% with one error to 14.6\% with recurring errors.



\newpage
\section*{NeurIPS Paper Checklist}

\begin{enumerate}

\item {\bf Claims}
    \item[] Question: Do the main claims made in the abstract and introduction accurately reflect the paper's contributions and scope?
    \item[] Answer: \answerYes{}
    \item[] Justification: The abstract and \S\ref{sec:intro} state three contributions: (1) the AgentRobustBench benchmark with 22 production-grounded perturbation types (\S\ref{sec:benchmark}); (2) the ToolRL-DR domain-randomization training recipe (\S\ref{sec:method}); (3) a public submission leaderboard (\S\ref{sec:platform}). Each contribution is substantiated in the corresponding section.
    \item[] Guidelines:
    \begin{itemize}
        \item The answer \answerNA{} means that the abstract and introduction do not include the claims made in the paper.
        \item The abstract and/or introduction should clearly state the claims made, including the contributions made in the paper and important assumptions and limitations. A \answerNo{} or \answerNA{} answer to this question will not be perceived well by the reviewers. 
        \item The claims made should match theoretical and experimental results, and reflect how much the results can be expected to generalize to other settings. 
        \item It is fine to include aspirational goals as motivation as long as it is clear that these goals are not attained by the paper. 
    \end{itemize}

\item {\bf Limitations}
    \item[] Question: Does the paper discuss the limitations of the work performed by the authors?
    \item[] Answer: \answerYes{}
    \item[] Justification: We include a Limitations Section in the paper.
    \item[] Guidelines:
    \begin{itemize}
        \item The answer \answerNA{} means that the paper has no limitation while the answer \answerNo{} means that the paper has limitations, but those are not discussed in the paper. 
        \item The authors are encouraged to create a separate ``Limitations'' section in their paper.
        \item The paper should point out any strong assumptions and how robust the results are to violations of these assumptions (e.g., independence assumptions, noiseless settings, model well-specification, asymptotic approximations only holding locally). The authors should reflect on how these assumptions might be violated in practice and what the implications would be.
        \item The authors should reflect on the scope of the claims made, e.g., if the approach was only tested on a few datasets or with a few runs. In general, empirical results often depend on implicit assumptions, which should be articulated.
        \item The authors should reflect on the factors that influence the performance of the approach. For example, a facial recognition algorithm may perform poorly when image resolution is low or images are taken in low lighting. Or a speech-to-text system might not be used reliably to provide closed captions for online lectures because it fails to handle technical jargon.
        \item The authors should discuss the computational efficiency of the proposed algorithms and how they scale with dataset size.
        \item If applicable, the authors should discuss possible limitations of their approach to address problems of privacy and fairness.
        \item While the authors might fear that complete honesty about limitations might be used by reviewers as grounds for rejection, a worse outcome might be that reviewers discover limitations that aren't acknowledged in the paper. The authors should use their best judgment and recognize that individual actions in favor of transparency play an important role in developing norms that preserve the integrity of the community. Reviewers will be specifically instructed to not penalize honesty concerning limitations.
    \end{itemize}

\item {\bf Theory assumptions and proofs}
    \item[] Question: For each theoretical result, does the paper provide the full set of assumptions and a complete (and correct) proof?
    \item[] Answer: \answerNA{}
    \item[] Justification: We define the tool-use POMDP as an organizational abstraction (\S\ref{sec:bg-mdp}) but do not state or prove any theorems.
    \item[] Guidelines:
    \begin{itemize}
        \item The answer \answerNA{} means that the paper does not include theoretical results. 
        \item All the theorems, formulas, and proofs in the paper should be numbered and cross-referenced.
        \item All assumptions should be clearly stated or referenced in the statement of any theorems.
        \item The proofs can either appear in the main paper or the supplemental material, but if they appear in the supplemental material, the authors are encouraged to provide a short proof sketch to provide intuition. 
        \item Inversely, any informal proof provided in the core of the paper should be complemented by formal proofs provided in appendix or supplemental material.
        \item Theorems and Lemmas that the proof relies upon should be properly referenced. 
    \end{itemize}

    \item {\bf Experimental result reproducibility}
    \item[] Question: Does the paper fully disclose all the information needed to reproduce the main experimental results of the paper to the extent that it affects the main claims and/or conclusions of the paper (regardless of whether the code and data are provided or not)?
    \item[] Answer: \answerYes{}
    \item[] Justification: \S\ref{sec:exp-setup} describes the inference protocol, perturbation injection rate, scoring rules, and uncertainty quantification. Per-model serving and decoding parameters are listed in Appendix~\ref{app:eval-details} (Table~\ref{tab:eval-settings}); the ToolRL-DR training recipe (data composition, GRPO hyperparameters, wall-clock) is in Appendix~\ref{app:compute} (Tables~\ref{tab:hyperparams}--\ref{tab:hyperparams-runs}). The public release at \url{https://github.com/WillChow66/AgentRobustBench} ships the eval pipeline, the deterministic scorer, the 17 perturbation JSONL files used for evaluation, the parquets actually fed to the GRPO trainer (\texttt{train.parquet}, \texttt{mixed\_train.parquet}, \texttt{test.parquet}), the GRPO launcher, and our one-function reward-score patch (\texttt{rlla\_perturbed.py}, a drop-in for \texttt{verl/utils/reward\_score/rlla.py}); together with the publicly available ToolRL repository as a setup dependency these are sufficient to retrain ToolRL-DR-Full and ToolRL-DR-Mixed end-to-end.
    \item[] Guidelines:
    \begin{itemize}
        \item The answer \answerNA{} means that the paper does not include experiments.
        \item If the paper includes experiments, a \answerNo{} answer to this question will not be perceived well by the reviewers: Making the paper reproducible is important, regardless of whether the code and data are provided or not.
        \item If the contribution is a dataset and\slash or model, the authors should describe the steps taken to make their results reproducible or verifiable. 
        \item Depending on the contribution, reproducibility can be accomplished in various ways. For example, if the contribution is a novel architecture, describing the architecture fully might suffice, or if the contribution is a specific model and empirical evaluation, it may be necessary to either make it possible for others to replicate the model with the same dataset, or provide access to the model. In general. releasing code and data is often one good way to accomplish this, but reproducibility can also be provided via detailed instructions for how to replicate the results, access to a hosted model (e.g., in the case of a large language model), releasing of a model checkpoint, or other means that are appropriate to the research performed.
        \item While NeurIPS does not require releasing code, the conference does require all submissions to provide some reasonable avenue for reproducibility, which may depend on the nature of the contribution. For example
        \begin{enumerate}
            \item If the contribution is primarily a new algorithm, the paper should make it clear how to reproduce that algorithm.
            \item If the contribution is primarily a new model architecture, the paper should describe the architecture clearly and fully.
            \item If the contribution is a new model (e.g., a large language model), then there should either be a way to access this model for reproducing the results or a way to reproduce the model (e.g., with an open-source dataset or instructions for how to construct the dataset).
            \item We recognize that reproducibility may be tricky in some cases, in which case authors are welcome to describe the particular way they provide for reproducibility. In the case of closed-source models, it may be that access to the model is limited in some way (e.g., to registered users), but it should be possible for other researchers to have some path to reproducing or verifying the results.
        \end{enumerate}
    \end{itemize}

\item {\bf Open access to data and code}
    \item[] Question: Does the paper provide open access to the data and code, with sufficient instructions to faithfully reproduce the main experimental results, as described in supplemental material?
    \item[] Answer: \answerYes{}
    \item[] Justification: The public release at \url{https://github.com/WillChow66/AgentRobustBench} contains: (i) the inference runner (\texttt{scripts/run\_eval.py}) and deterministic scorer (\texttt{leaderboard/scoring/eval\_results.py}); (ii) the leaderboard Gradio backend (\texttt{leaderboard/app.py}) and the 21-row seed scoreboard (\texttt{leaderboard/leaderboard.csv}); (iii) the 17 perturbation JSONL files for evaluation (\S\ref{sec:bench-sources}); (iv) the GRPO training assets needed to reproduce ToolRL-DR-Full and ToolRL-DR-Mixed: the parquets actually fed to the trainer (\texttt{training/data/train.parquet}, \texttt{mixed\_train.parquet}, \texttt{test.parquet}), the GRPO launcher (\texttt{training/grpo\_launch.sh}), the human-readable hyperparameter manifest (\texttt{training/grpo\_hyperparams.yaml}), the reference jsonl-to-parquet converter (\texttt{training/convert\_to\_toolrl\_parquet.py}), and our one-function reward-score patch (\texttt{training/rlla\_perturbed.py}, drop-in for \texttt{verl/utils/reward\_score/rlla.py}). The public leaderboard is available at \url{https://huggingface.co/spaces/willchow66/AgentRobustBench}.
    \item[] Guidelines:
    \begin{itemize}
        \item The answer \answerNA{} means that paper does not include experiments requiring code.
        \item Please see the NeurIPS code and data submission guidelines (\url{https://neurips.cc/public/guides/CodeSubmissionPolicy}) for more details.
        \item While we encourage the release of code and data, we understand that this might not be possible, so \answerNo{} is an acceptable answer. Papers cannot be rejected simply for not including code, unless this is central to the contribution (e.g., for a new open-source benchmark).
        \item The instructions should contain the exact command and environment needed to run to reproduce the results. See the NeurIPS code and data submission guidelines (\url{https://neurips.cc/public/guides/CodeSubmissionPolicy}) for more details.
        \item The authors should provide instructions on data access and preparation, including how to access the raw data, preprocessed data, intermediate data, and generated data, etc.
        \item The authors should provide scripts to reproduce all experimental results for the new proposed method and baselines. If only a subset of experiments are reproducible, they should state which ones are omitted from the script and why.
        \item At submission time, to preserve anonymity, the authors should release anonymized versions (if applicable).
        \item Providing as much information as possible in supplemental material (appended to the paper) is recommended, but including URLs to data and code is permitted.
    \end{itemize}

\item {\bf Experimental setting/details}
    \item[] Question: Does the paper specify all the training and test details (e.g., data splits, hyperparameters, how they were chosen, type of optimizer) necessary to understand the results?
    \item[] Answer: \answerYes{}
    \item[] Justification: Training data composition is described in \S\ref{sec:method-method}; the parquets shipped in the anonymized release contain 3{,}905 perturbed train + 79 held-out validation samples for ToolRL-DR-Full, and 4{,}000 train (2{,}006 clean + 1{,}994 perturbed) + 79 validation samples for ToolRL-DR-Mixed (the same validation parquet is reused). The full GRPO hyperparameter list (batch size, mini-batch, micro-batch, rollout group size, max prompt/response length, learning rate, epochs, reward) is in Appendix~\ref{app:compute} (Table~\ref{tab:hyperparams}). Eval-side decoding parameters per model are in Appendix~\ref{app:eval-details} (Table~\ref{tab:eval-settings}).
    \item[] Guidelines:
    \begin{itemize}
        \item The answer \answerNA{} means that the paper does not include experiments.
        \item The experimental setting should be presented in the core of the paper to a level of detail that is necessary to appreciate the results and make sense of them.
        \item The full details can be provided either with the code, in appendix, or as supplemental material.
    \end{itemize}

\item {\bf Experiment statistical significance}
    \item[] Question: Does the paper report error bars suitably and correctly defined or other appropriate information about the statistical significance of the experiments?
    \item[] Answer: \answerYes{}
    \item[] Justification: The Uncertainty Quantification paragraph in \S\ref{sec:exp-setup} describes 95\% percentile-bootstrap CIs on accuracy and on the clean-to-perturbed drop using $B={\hspace{-0.05em}}10{,}000$ resamples of the per-sample binary correctness scores. The Main Results table (Table~\ref{tab:main}) and the per-perturbation breakdown (Tables~\ref{tab:per-pert-obs-1}--\ref{tab:per-pert-trn-2}) report mean $\pm$ 95\% half-width per cell. Our two trained-checkpoint rows in Table~\ref{tab:main} additionally carry paired-bootstrap significance markers ($^*p<0.05$, $^{**}p<0.01$, $^{***}p<0.001$) against the size-matched public ToolRL-Clean baseline; cross-model differences underlying the headline claims (\S\ref{sec:exp-main}, \S\ref{sec:exp-paradigm}) exceed the corresponding 95\% half-widths and are therefore statistically significant. Decoding is deterministic (temperature $0$), so the only variance source is the finite-sample uncertainty across benchmark items, which the bootstrap captures.
    \item[] Guidelines:
    \begin{itemize}
        \item The answer \answerNA{} means that the paper does not include experiments.
        \item The authors should answer \answerYes{} if the results are accompanied by error bars, confidence intervals, or statistical significance tests, at least for the experiments that support the main claims of the paper.
        \item The factors of variability that the error bars are capturing should be clearly stated (for example, train/test split, initialization, random drawing of some parameter, or overall run with given experimental conditions).
        \item The method for calculating the error bars should be explained (closed form formula, call to a library function, bootstrap, etc.)
        \item The assumptions made should be given (e.g., Normally distributed errors).
        \item It should be clear whether the error bar is the standard deviation or the standard error of the mean.
        \item It is OK to report 1-sigma error bars, but one should state it. The authors should preferably report a 2-sigma error bar than state that they have a 96\% CI, if the hypothesis of Normality of errors is not verified.
        \item For asymmetric distributions, the authors should be careful not to show in tables or figures symmetric error bars that would yield results that are out of range (e.g., negative error rates).
        \item If error bars are reported in tables or plots, the authors should explain in the text how they were calculated and reference the corresponding figures or tables in the text.
    \end{itemize}

\item {\bf Experiments compute resources}
    \item[] Question: For each experiment, does the paper provide sufficient information on the computer resources (type of compute workers, memory, time of execution) needed to reproduce the experiments?
    \item[] Answer: \answerYes{}
    \item[] Justification: Appendix~\ref{sec:impact} (Compute and environmental cost) and Appendix~\ref{app:compute} disclose: $\approx$\,4.3 GH200 GPU-hours for the 21-model evaluation sweep, $\approx$\,9.3 hours per ToolRL-DR training run on 2$\times$NVIDIA RTX 6000 Ada (48\,GB), and $<$10 USD of OpenAI API spend for benchmark generation. Per-model wall-clock and GPU configuration are in Table~\ref{tab:eval-settings}.
    \item[] Guidelines:
    \begin{itemize}
        \item The answer \answerNA{} means that the paper does not include experiments.
        \item The paper should indicate the type of compute workers CPU or GPU, internal cluster, or cloud provider, including relevant memory and storage.
        \item The paper should provide the amount of compute required for each of the individual experimental runs as well as estimate the total compute. 
        \item The paper should disclose whether the full research project required more compute than the experiments reported in the paper (e.g., preliminary or failed experiments that didn't make it into the paper). 
    \end{itemize}
    
\item {\bf Code of ethics}
    \item[] Question: Does the research conducted in the paper conform, in every respect, with the NeurIPS Code of Ethics \url{https://neurips.cc/public/EthicsGuidelines}?
    \item[] Answer: \answerYes{}
    \item[] Justification: We use only public, permissively licensed source benchmarks (\S\ref{sec:bench-sources}) and publicly released model checkpoints. No human subjects or crowdsourced labels are involved. The perturbation taxonomy is descriptive of failures already documented in public GitHub issues; the trained ToolRL-DR checkpoints are designed to increase robustness to those perturbations rather than amplify any harmful behavior (Appendix~\ref{sec:impact}).
    \item[] Guidelines:
    \begin{itemize}
        \item The answer \answerNA{} means that the authors have not reviewed the NeurIPS Code of Ethics.
        \item If the authors answer \answerNo, they should explain the special circumstances that require a deviation from the Code of Ethics.
        \item The authors should make sure to preserve anonymity (e.g., if there is a special consideration due to laws or regulations in their jurisdiction).
    \end{itemize}

\item {\bf Broader impacts}
    \item[] Question: Does the paper discuss both potential positive societal impacts and negative societal impacts of the work performed?
    \item[] Answer: \answerYes{}
    \item[] Justification: Appendix~\ref{sec:impact} (Broader Impact) discusses intended use (supporting robustness research and surfacing deployment-time failure modes), misuse risk (the perturbation taxonomy describes failures already documented in public GitHub issues, so does not increase the attack surface), and compute / environmental cost.
    \item[] Guidelines:
    \begin{itemize}
        \item The answer \answerNA{} means that there is no societal impact of the work performed.
        \item If the authors answer \answerNA{} or \answerNo, they should explain why their work has no societal impact or why the paper does not address societal impact.
        \item Examples of negative societal impacts include potential malicious or unintended uses (e.g., disinformation, generating fake profiles, surveillance), fairness considerations (e.g., deployment of technologies that could make decisions that unfairly impact specific groups), privacy considerations, and security considerations.
        \item The conference expects that many papers will be foundational research and not tied to particular applications, let alone deployments. However, if there is a direct path to any negative applications, the authors should point it out. For example, it is legitimate to point out that an improvement in the quality of generative models could be used to generate Deepfakes for disinformation. On the other hand, it is not needed to point out that a generic algorithm for optimizing neural networks could enable people to train models that generate Deepfakes faster.
        \item The authors should consider possible harms that could arise when the technology is being used as intended and functioning correctly, harms that could arise when the technology is being used as intended but gives incorrect results, and harms following from (intentional or unintentional) misuse of the technology.
        \item If there are negative societal impacts, the authors could also discuss possible mitigation strategies (e.g., gated release of models, providing defenses in addition to attacks, mechanisms for monitoring misuse, mechanisms to monitor how a system learns from feedback over time, improving the efficiency and accessibility of ML).
    \end{itemize}
    
\item {\bf Safeguards}
    \item[] Question: Does the paper describe safeguards that have been put in place for responsible release of data or models that have a high risk for misuse (e.g., pre-trained language models, image generators, or scraped datasets)?
    \item[] Answer: \answerNA{}
    \item[] Justification: The released artifacts are a perturbation benchmark constructed from existing publicly licensed tool-use datasets and two 3B tool-use checkpoints designed to be more robust to those perturbations. None of these have a high risk for misuse: the perturbations describe failures already documented in public GitHub issues (Appendix~\ref{sec:impact}), and the 3B checkpoints are not generative models that produce dual-use content.
    \item[] Guidelines:
    \begin{itemize}
        \item The answer \answerNA{} means that the paper poses no such risks.
        \item Released models that have a high risk for misuse or dual-use should be released with necessary safeguards to allow for controlled use of the model, for example by requiring that users adhere to usage guidelines or restrictions to access the model or implementing safety filters. 
        \item Datasets that have been scraped from the Internet could pose safety risks. The authors should describe how they avoided releasing unsafe images.
        \item We recognize that providing effective safeguards is challenging, and many papers do not require this, but we encourage authors to take this into account and make a best faith effort.
    \end{itemize}

\item {\bf Licenses for existing assets}
    \item[] Question: Are the creators or original owners of assets (e.g., code, data, models), used in the paper, properly credited and are the license and terms of use explicitly mentioned and properly respected?
    \item[] Answer: \answerYes{}
    \item[] Justification: All five source benchmarks (BFCL V3, API-Bank, RoTBench, ToolAlpaca, ToolEyes) are cited in \S\ref{sec:bench-sources} and redistributed under their original licenses; the per-source license is listed on the public HuggingFace dataset card at \url{https://huggingface.co/datasets/AgentRobustBench/AgentRobustBench} (Appendix~\ref{sec:impact}, Data Licensing). The 19 evaluated public model checkpoints are cited individually in \S\ref{sec:exp-setup} and used under the upstream HuggingFace model-card terms.
    \item[] Guidelines:
    \begin{itemize}
        \item The answer \answerNA{} means that the paper does not use existing assets.
        \item The authors should cite the original paper that produced the code package or dataset.
        \item The authors should state which version of the asset is used and, if possible, include a URL.
        \item The name of the license (e.g., CC-BY 4.0) should be included for each asset.
        \item For scraped data from a particular source (e.g., website), the copyright and terms of service of that source should be provided.
        \item If assets are released, the license, copyright information, and terms of use in the package should be provided. For popular datasets, \url{paperswithcode.com/datasets} has curated licenses for some datasets. Their licensing guide can help determine the license of a dataset.
        \item For existing datasets that are re-packaged, both the original license and the license of the derived asset (if it has changed) should be provided.
        \item If this information is not available online, the authors are encouraged to reach out to the asset's creators.
    \end{itemize}

\item {\bf New assets}
    \item[] Question: Are new assets introduced in the paper well documented and is the documentation provided alongside the assets?
    \item[] Answer: \answerYes{}
    \item[] Justification: Each released artifact ships with documentation. The public release at \url{https://github.com/WillChow66/AgentRobustBench} has a top-level \texttt{README.md} describing the four-directory layout and quick-start commands, a \texttt{scripts/README.md} for the eval pipeline, a \texttt{scripts/slurm/README.md} for the SLURM templates, a \texttt{leaderboard/README.md} with the submission-format spec, and a \texttt{training/README.md} describing the parquet schema, the reward patch, and two-step reproduction commands for ToolRL-DR-Full / DR-Mixed. The eval data is published at \url{https://huggingface.co/datasets/AgentRobustBench/AgentRobustBench} with a HuggingFace dataset card listing source benchmarks, the 22 perturbation types, and per-source licenses, plus a Croissant 1.0 metadata file with the full RAI extension; per-perturbation generation prompts (Appendix~\ref{app:prompts}) and the audit log (Appendix~\ref{app:evidence}) accompany the data. The evaluated \href{https://huggingface.co/willchow66/ToolRL-DR-Full-Qwen2.5-3B}{ToolRL-DR-Full} and \href{https://huggingface.co/willchow66/ToolRL-DR-Mixed-Qwen2.5-3B}{ToolRL-DR-Mixed} checkpoints and model cards are publicly available on Hugging Face.
    \item[] Guidelines:
    \begin{itemize}
        \item The answer \answerNA{} means that the paper does not release new assets.
        \item Researchers should communicate the details of the dataset\slash code\slash model as part of their submissions via structured templates. This includes details about training, license, limitations, etc. 
        \item The paper should discuss whether and how consent was obtained from people whose asset is used.
        \item At submission time, remember to anonymize your assets (if applicable). You can either create an anonymized URL or include an anonymized zip file.
    \end{itemize}

\item {\bf Crowdsourcing and research with human subjects}
    \item[] Question: For crowdsourcing experiments and research with human subjects, does the paper include the full text of instructions given to participants and screenshots, if applicable, as well as details about compensation (if any)? 
    \item[] Answer: \answerNA{}
    \item[] Justification: The work involves no human subjects and no crowdsourced annotations. Perturbations are generated either by rule (deterministic) or by LLM prompts that are reproduced verbatim in Appendix~\ref{app:prompts}; quality control uses an LLM-as-judge audit (Appendix~\ref{app:paraphrase-audit}), not human raters.
    \item[] Guidelines:
    \begin{itemize}
        \item The answer \answerNA{} means that the paper does not involve crowdsourcing nor research with human subjects.
        \item Including this information in the supplemental material is fine, but if the main contribution of the paper involves human subjects, then as much detail as possible should be included in the main paper. 
        \item According to the NeurIPS Code of Ethics, workers involved in data collection, curation, or other labor should be paid at least the minimum wage in the country of the data collector. 
    \end{itemize}

\item {\bf Institutional review board (IRB) approvals or equivalent for research with human subjects}
    \item[] Question: Does the paper describe potential risks incurred by study participants, whether such risks were disclosed to the subjects, and whether Institutional Review Board (IRB) approvals (or an equivalent approval/review based on the requirements of your country or institution) were obtained?
    \item[] Answer: \answerNA{}
    \item[] Justification: No human subjects are involved in the work, so no IRB review is applicable.
    \item[] Guidelines:
    \begin{itemize}
        \item The answer \answerNA{} means that the paper does not involve crowdsourcing nor research with human subjects.
        \item Depending on the country in which research is conducted, IRB approval (or equivalent) may be required for any human subjects research. If you obtained IRB approval, you should clearly state this in the paper. 
        \item We recognize that the procedures for this may vary significantly between institutions and locations, and we expect authors to adhere to the NeurIPS Code of Ethics and the guidelines for their institution. 
        \item For initial submissions, do not include any information that would break anonymity (if applicable), such as the institution conducting the review.
    \end{itemize}

\item {\bf Declaration of LLM usage}
    \item[] Question: Does the paper describe the usage of LLMs if it is an important, original, or non-standard component of the core methods in this research? Note that if the LLM is used only for writing, editing, or formatting purposes and does \emph{not} impact the core methodology, scientific rigor, or originality of the research, declaration is not required.
    \item[] Answer: \answerYes{}
    \item[] Justification: LLMs are used in two methodological roles, both fully disclosed: (i) \emph{gpt-5-mini} (with a small \emph{gpt-4o-mini} fallback) generates the four LLM-required perturbation types (Typo, QueryPara, ToolPara, ParamPara) and the description rewriting in MisDesc / TimeDesc, with the exact prompts reproduced in Appendix~\ref{app:prompts}; (ii) \emph{gpt-5} acts as the semantic-equivalence judge in the paraphrase quality audit (Appendix~\ref{app:paraphrase-audit}), with the rubric and per-type score distributions reported. The 21 evaluated tool-use models (\S\ref{sec:exp-setup}) are the subject of study, not part of the methodology.
    \item[] Guidelines:
    \begin{itemize}
        \item The answer \answerNA{} means that the core method development in this research does not involve LLMs as any important, original, or non-standard components.
        \item Please refer to our LLM policy in the NeurIPS handbook for what should or should not be described.
    \end{itemize}

\end{enumerate}

\end{document}